\documentclass[10pt,journal,compsoc]{IEEEtran}
\usepackage{multirow}

\ifCLASSOPTIONcompsoc
  \usepackage[nocompress]{cite}
\else
  \usepackage{cite}
\fi
\ifCLASSINFOpdf
  \usepackage[pdftex]{graphicx}
\else
\fi
\usepackage{url}

\usepackage{color}
\definecolor{orange}{rgb}{1,0.5,0}

\begin{document}
%
\title{Human Action Localization \\ with Sparse Spatial Supervision}

\author{Philippe~Weinzaepfel,
        Xavier~Martin,
        and~Cordelia~Schmid,~\IEEEmembership{Fellow,~IEEE}
\IEEEcompsocitemizethanks{\IEEEcompsocthanksitem P. Weinzaepfel is with Xerox Research Centre Europe, Meylan, France. \protect\\ E-mail: philippe.weinzaepfel@xrce.xerox.com
\IEEEcompsocthanksitem X. Martin and C. Schmid are with Inria, LJK,  Grenoble, France.  \protect\\
E-mail: firstname.lastname@inria.fr}
}

\IEEEtitleabstractindextext{%

\begin{abstract}

We introduce an approach for spatio-temporal human action localization
using sparse spatial supervision.  Our method leverages the large
amount of annotated humans available today and extracts human tubes by
combining a state-of-the-art human detector with a
tracking-by-detection approach. Given these high-quality human tubes
and temporal supervision, we select positive and negative tubes with
very sparse spatial supervision, i.e., only one spatially annotated
frame per instance. The selected tubes allow us to effectively learn a
spatio-temporal action detector based on dense trajectories or CNNs.  
We conduct experiments on existing action localization benchmarks:
  UCF-Sports, J-HMDB and UCF-101. Our results show that our approach, despite
  using sparse spatial supervision, performs on par with methods using
  full supervision, i.e., one bounding box annotation per frame.
To further validate our method, we introduce \textbf{DALY} (Daily Action
Localization in YouTube), a dataset for realistic
action localization in space and time. 
It contains high quality temporal and spatial annotations for 
3.6k instances of 10 actions in 31 hours of videos (3.3M frames). It is an
order of magnitude larger than existing datasets, with more diversity
in appearance and long untrimmed videos. 

\end{abstract}

}

\maketitle

\IEEEdisplaynontitleabstractindextext

%
\IEEEpeerreviewmaketitle

\IEEEraisesectionheading{\section{Introduction}\label{sec:intro}}

\IEEEPARstart{A}{ction} classification has been widely studied over the past decade and state-of-the-art
methods~\cite{wang:hal-01145834,2streamsCNN,tran2014c3d,yue2015beyond,TSN2016ECCV} now achieve excellent performance. 
However, to analyze video content in more detail, we need to localize actions in space and time. 
Detecting actions in videos is a challenging task which has received increasing attention over the past few years. 
Recently, significant progress has been achieved in supervised action localization, see for example~\cite{fat,weinzaepfelICCV15,WangQT14,Saha2016,MR2RCNN}.
However these methods require a large amount of annotation, i.e., bounding box annotations in every frame.
Such annotations are, for example, used to train Convolutional Neural Networks (CNNs)~\cite{fat,weinzaepfelICCV15,Saha2016,MR2RCNN} 
at the bounding box level.
Several works have suggested to generate action proposals before classifying them~\cite{marian2015unsupervised,van2015apt},
however they generate hundreds of proposals for a video, thus supervision is still required to label them in order to train a classifier.
Consequently, all these approaches require full supervision, where action localization needs to be annotated in every frame.
This makes scaling up to a large dataset difficult. The goal of this
paper is to move away from full supervision, similar in spirit to
recent work on weakly-supervised object localization~\cite{multifoldMIL,bilen2016weakly}.

\begin{figure}
 \centering
 \includegraphics[width=\linewidth]{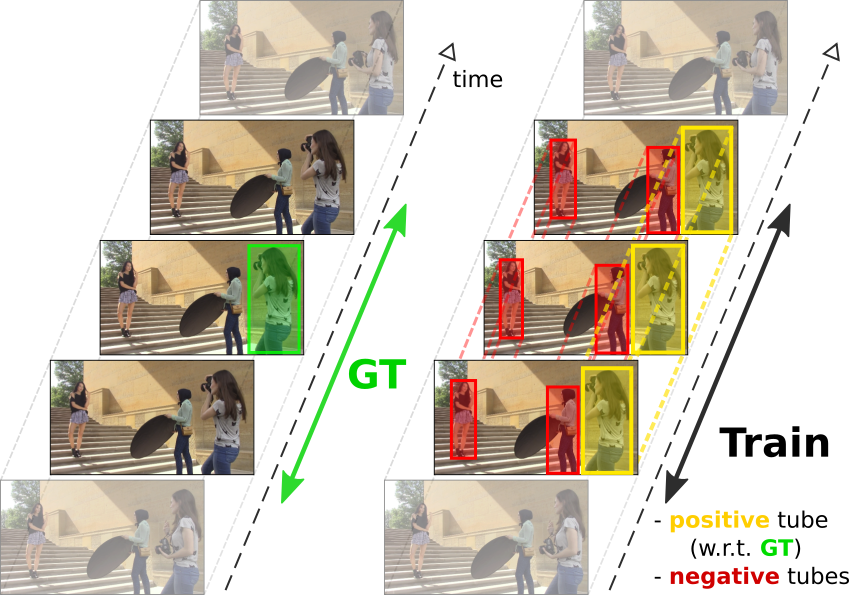}
 \caption{We consider sparse spatial supervision: the temporal extent of the action as well as one box per instance are annotated in the training videos (left).
 To train an action detector, we extract human tubes and select positive and negative ones (right) according to the sparse annotations.
 }
 \label{fig:intro}
\end{figure}

Recently, Mettes~et al.~\cite{mettes2016spot} have addressed action
localization with another annotation scheme, e.g. with pointly-supervised
proposals. A large number of candidate proposals are obtained using
APT~\cite{van2015apt}, a method based on grouping dense trajectories.  
They show that Multiple Instance Learning (MIL) applied directly on these
proposals performs poorly. They thus introduce point supervision and 
incorporate an overlap measure between annotated points and
proposals into the mining process.  
This requires annotating a point in every frame.
In this paper we go a step further and
significantly reduce the number of frames to annotate.
To this end, we leverage the
fact that actors are humans and extract human tubes. Given these human
tubes, our approach uses only one spatial annotation per action instance,
see Figure~\ref{fig:intro}.  We show that such a sparse annotation scheme is 
sufficient to train state-of-the-art action detectors.

\begin{figure*}
 \centering
 \small
  \begin{tabular}{c@{ }c@{ }c@{ }c@{ }c}
   \includegraphics[width=0.195\textwidth,height=0.11\textwidth]{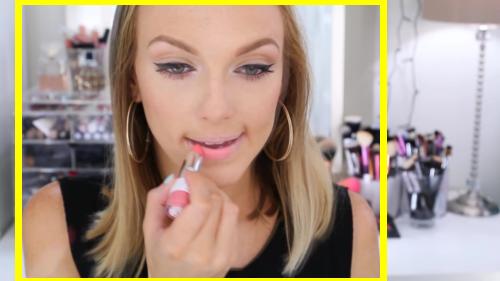} &
   \includegraphics[width=0.195\textwidth,height=0.11\textwidth]{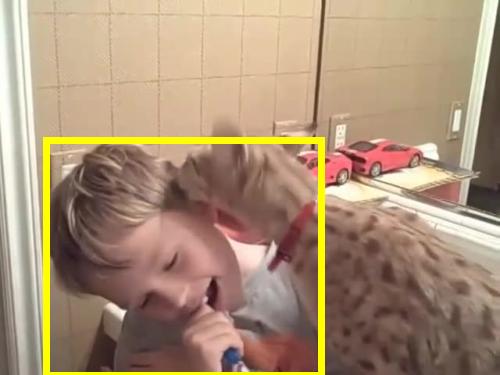} &
   \includegraphics[width=0.195\textwidth,height=0.11\textwidth]{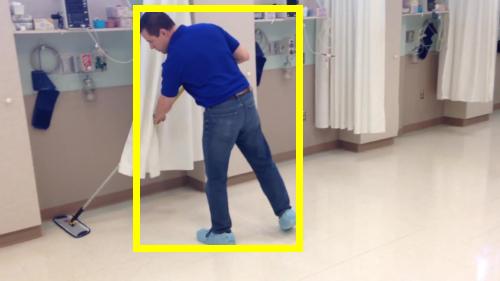} &
   \includegraphics[width=0.195\textwidth,height=0.11\textwidth]{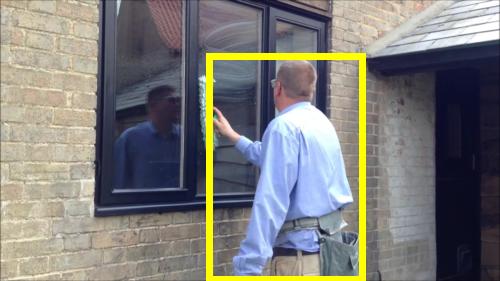} &
   \includegraphics[width=0.195\textwidth,height=0.11\textwidth]{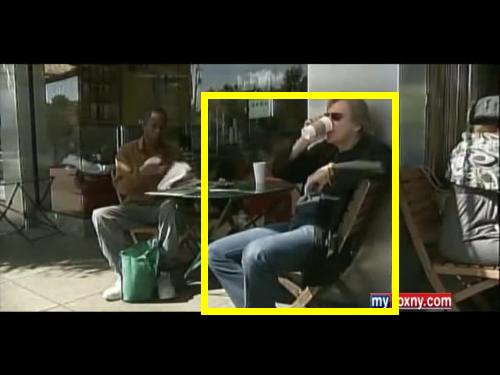} \\
   \emph{applying make up on lips} & \emph{brushing teeth} & \emph{cleaning floor} & \emph{cleaning} & \emph{drinking} \\[0.2cm]
   \includegraphics[width=0.195\textwidth,height=0.11\textwidth]{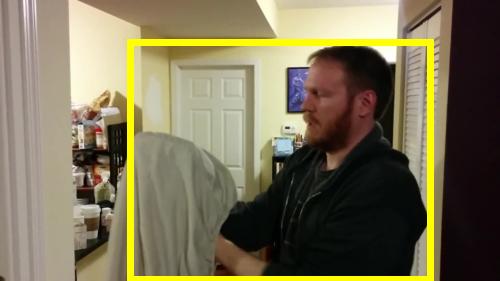} &
   \includegraphics[width=0.195\textwidth,height=0.11\textwidth]{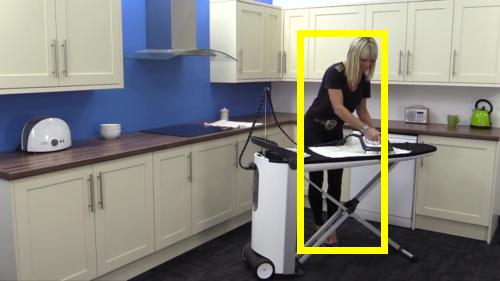} &
   \includegraphics[width=0.195\textwidth,height=0.11\textwidth]{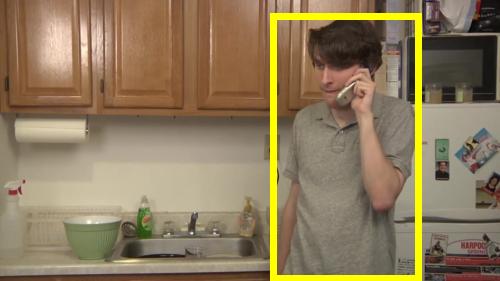} &
   \includegraphics[width=0.195\textwidth,height=0.11\textwidth]{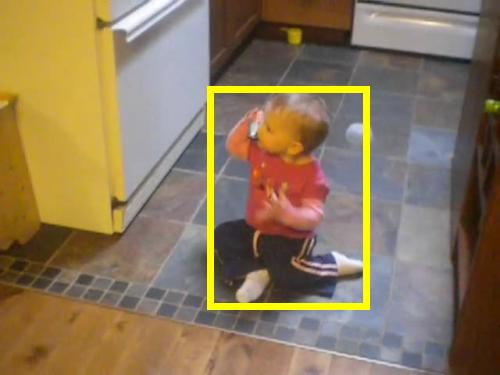} &
   \includegraphics[width=0.195\textwidth,height=0.11\textwidth]{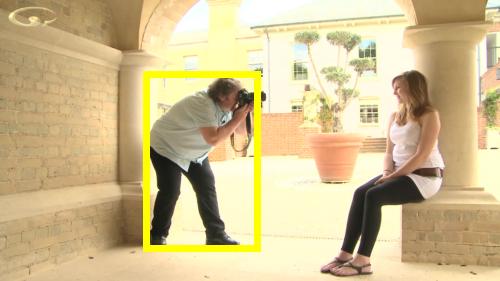} \\
   \emph{folding textile} & \emph{ironing} & \emph{phoning} & \emph{playing harmonica} & \emph{taking photos/videos} \\
  \end{tabular}
  \normalsize
 \caption{Illustration of the 10 classes of our DALY dataset. One annotated frame is shown per class.}
 \label{fig:daly}
\end{figure*}

Our approach first extracts human tubes from videos.
Using human tubes for action recognition is not a novel idea~\cite{klaser:inria-00514845,prest:hal-00720847,Yu_2015_CVPR}. 
However, we show that extracting high quality 
human tubes is possible by leveraging a recent
state-of-the-art object detection approach (Faster R-CNN~\cite{fasterRCNN}), a large
annotated dataset of humans in a variety of poses (MPII Human Pose~\cite{mpiipose}) and 
a state-of-the-art tracking-by-detection approach~\cite{struck,tld}.
Our experiments demonstrate that a small number of human tubes per video is sufficient to
obtain a recall of  95\% on the UCF-Sports and J-HMDB datasets, and 65\% on the more challenging UCF-101 benchmark.

Assuming that the temporal extent of the action is given at train time,
we study how much spatial supervision is required for training an
action detector, when sparse spatial  annotations are used to 
label human tubes as positive or negative, see Figure~\ref{fig:intro}. 
We show that spatial supervision can be 
reduced to only one annotated frame without impacting performance
significantly. 
This is observed both for learning a SVM on dense trajectories (IDT)~\cite{wang:hal-01145834} and for training CNNs~\cite{2streamsCNN}.
Our action detector combining IDT and CNN obtains a mAP of 96\%, 64\% and 57\% on UCF-Sports, J-HMDB and UCF-101 respectively, with only one annotated frame. 
The performance is comparable to full supervision, i.e., the drop is at most 
 2\% while the annotation cost is drastically reduced.
We also significantly outperform other methods that aim at reducing supervision. 
For example, we obtain 57\% mAP on the UCF-101 dataset with
one annotated bounding box, i.e., 2 points annotated per action instance,
whereas Mettes et al.~\cite{mettes2016spot} 
obtain 32\% mAP but use one point annotation in every frame. 
This represents 4030 bounding box annotations (i.e., 8060 points) for our sparse annotation scheme compared to almost 1.9 million points for~\cite{mettes2016spot}.

To further validate our method, we introduce the  Daily
Action Localization in YouTube (\textbf{DALY}) dataset. It is designed to correct the  
drawbacks of existing datasets, which are trimmed (UCF-Sports, J-HMDB) or
almost-trimmed videos (UCF-101) with specific action types,
e.g. sports only, showing in most cases only one human per video. 
DALY is a large dataset with diverse actions in untrimmed videos,
sampled from real-world data. 
It consists of more than 31 hours of videos (3.3M frames) from
YouTube with 3.6k spatio-temporal action instances for 
10 realistic daily actions, see Figure~\ref{fig:daly}.  Annotations indicate the
start and end time of each action instance, with spatial
annotations for a sparse subset of frames. The task is to localize
relatively short actions (8 seconds in average) in long untrimmed
videos (3min 45s in average).  Furthermore, it includes videos with
multiple humans performing actions simultaneously.  On the DALY
dataset our human tubes obtain a spatial recall of 95\%, but the
detection task is extremely challenging, we obtain a mean Average Precision of 14\%.

This paper is organized as follows. We first review related work in Section~\ref{sec:related}.
We then describe our approach for human action localization with
sparse spatial supervision in Section~\ref{sec:overview}. 
Section~\ref{sec:dataset} presents the datasets used in
experimental evaluation and introduces our DALY dataset. 
Next, we describe and evaluate our approach for extracting human tubes from videos
in Section~\ref{sec:tube}. 
Finally, Section~\ref{sec:xp} presents experimental results for action
 localization with sparse spatial supervision.
The DALY dataset is available online at
\url{http://thoth.inrialpes.fr/daly/}.

\section{Related work}
\label{sec:related}

In this section we review related work on action localization 
with full supervision (Section~\ref{sub:actionloc})
and partial supervision (Section~\ref{sub:supervision}).

\subsection{Fully-supervised action localization}
\label{sub:actionloc}

Initial attempts for temporal and spatio-temporal action localization are based on a sliding-window scheme and focus on improving 
the search complexity~\cite{LaptevP07,msr2,yuan2009discriminative,actoms}.
Other approaches rely on figure-centric models. 
For instance, Lan et al.~\cite{LanWM11} consider the human position as a latent variable and infer it jointly with the action label.
Kl\"{a}ser et al.~\cite{klaser:inria-00514845} use a human detector and build human tracks using KLT features tracks. 
The human tracks are then classified with HOG-3D descriptors~\cite{hog3d}.
Our approach is also based on human tracks but is significantly more robust to huge variations in pose and appearance. 

Several recent methods for action localization are based on action proposals to reduce the search complexity. 
Jain et al.~\cite{JainCVPR2014} build action localization candidates by hierarchically merging supervoxels and use dense trajectory features for tube classification.
Similarly, van Gemert et al.~\cite{van2015apt} cluster trajectories and use the resulting tubes for action detection. 
In~\cite{Yu_2015_CVPR}, proposals are based on an actionness measure~\cite{actionness}
which requires localized training samples. 
In parallel, several works~\cite{oneata:hal-01021902,marian2015unsupervised} have attempted to further improve the quality of tubes.
Most of these methods generate thousands of proposals for a short video and require ground-truth annotations
to label the proposals in order to learn a proposal classifier.
In contrast, our approach relies on only a few human tube proposals per video and can, thus, reduce spatial supervision to one annotated frame without drop of performance.  

Recently, CNNs for human action localization have emerged~\cite{fat,weinzaepfelICCV15,MR2RCNN,Saha2016}.
These approaches rely on R-CNNs for both appearance and motion, classifying region proposals in individual frames. 
Detection tubes are obtained by combining class-specific detections with either temporal linking based on proximity~\cite{fat}, or with a class-specific tracking-by-detection approach~\cite{weinzaepfelICCV15}.
Both strategies need to be run independently for each action.
State-of-the-art approaches~\cite{MR2RCNN,Saha2016} rely on Faster R-CNN trained on appearance and flow. 
Note that all these methods make extensive use of bounding box annotations in every frame for training.

\subsection{Action localization with partial supervision}
\label{sub:supervision}

Annotating all videos with bounding boxes in every frame is unrealistic for large-scale datasets,
yet reducing spatial supervision has received little attention so far.
 Weakly-supervised temporal localization was studied in~\cite{Bojanowski14weakly,duchenne2009automatic,hoai2014learning}. 
 Bojanowski et al.~\cite{Bojanowski14weakly} assume an ordered list of actions in each video as input.
 Duchenne et al.~\cite{duchenne2009automatic} use a discriminative clustering on short video segments to identify
 the temporal localization in the training set and learn a classifier.
The detection is then performed using a sliding window.
 Hoai et al.~\cite{hoai2014learning} extend a Multiple Instance SVM to time series, 
allowing for discontinuities in the positive samples.
In the context of object detection, Prest~et al.~\cite{prestCVPR12} propose to extract spatio-temporal tubes and then perform tube selection before training a classifier.
Siva and Xiang~\cite{siva2011weakly} apply multiple instance learning (MIL) on cuboids of various time lengths around detected humans, described by STIPs~\cite{laptev2005space}.
Their method is thus limited to static human actions, with a bounding box that does not move or change over time.
Recently, Mettes~et al.~\cite{mettes2016spot} propose to extract hundreds of action proposals and then apply MIL.
Given the huge number of proposals, good performance requires pruning of the proposals. 
To this end they consider spatial supervision in the form of 2D points annotated for each frame, which they refer to as pointly-supervised.

Some other works also detect actions without spatial supervision.
Mosabbeb et al.~\cite{mosabbeb2014multi} use a subspace segmentation clustering approach applied on groups of trajectories, in order to segment videos into parts.
A low-rank matrix completion method estimates the contribution of each cluster to the different labels, hence the approach detects several
disjoint action parts and not one consistent spatio-temporal localization. 
Ma et al.~\cite{STsegments} first extract a per-frame hierarchical segmentation, 
which is tracked over the video. Using foreground scoring, they obtain a hierarchy of spatio-temporal segments 
where the upper level corresponds to human body location candidates.
However, they rely on parts segmentation which is challenging in low-quality real-world videos, e.g. with strong occlusion and compression artifacts.
More recently, Chen and Corso~\cite{chen2015action} propose to generate unsupervised proposals by clustering intentional motion based on dense trajectories.
Their method can handle only one action per training video and is not robust to nearby motions.

\section{Our approach}
\label{sec:overview}

In this section we describe our approach for learning an action
detector with sparse spatial supervision.
Sparse means that ground-truth bounding boxes around the actors are annotated in a few frames only.
Sparse spatial annotation significantly reduces annotation cost,
an important factor to create real-world large-scale datasets.  

Our main hypothesis is that since actions are performed by humans, we can take advantage of existing human localization datasets.
Thanks to the large amount of human annotations at our disposal~\cite{mpiipose}, we can extract high-quality action-agnostic human tubes,
which we then label as negative or positive samples using the sparse spatial annotations, see Figure~\ref{fig:intro}.
The obtained 
labeling is then used to train action detectors. In this paper
we rely on two detectors, one based on dense  
trajectories~\cite{wang:hal-01145834} and the other on
CNNs~\cite{fasterRCNN}. At test time, we also extract human tubes and  
score them  with the trained detectors.

In this section, we first present how human tubes are extracted in
Section~\ref{sub:humantubes}. Next, we describe how these human
tubes are labeled from the few annotated frames required to train a
classifier (Section~\ref{sub:learn}) and the different approaches we consider for 
human tube scoring (Section~\ref{sub:features}).
We finally present how we perform temporal detection for untrimmed dataset in Section~\ref{sub:tempo}.

\begin{table*}[t]
 \centering
 \begin{tabular}{|c||c||c|c|c|}
 \hline
 & ~~~~~~\textbf{DALY}~~~~~~ & ~~UCF-Sports~\cite{ucfsports}~~ & ~~J-HMDB~\cite{jhmdb}~~ & ~~UCF-101~\cite{ucf101}~~ \\ 
 \hline
 \hline
 \#classes & 10 & 10 & 21 & \textbf{24} \\
 \hline
 action types & \textbf{everyday} & sports & \textbf{everyday} & sports \\ 
 \hline
 \#clips & \textbf{8133} & 150 & 928 & 3207 \\ 
 \hline
 avg resolution & \textbf{1290x790} & 690x450 & 320x240 & 320x240 \\
 \hline
 total \#frames & \textbf{3.3M} & 10k & 32k & 558k \\
 \hline
 avg video dur. & \textbf{3min 45s} & 5.8s & 1.4s & 5.8s \\
 \hline
 avg action dur. & \textbf{7.9s} & 5.8s & 1.4s & 4.5s \\ 
 \hline
 action dur. / video dur. & \textbf{4\%} & 100\% & 100\% & 78\% \\ 
 \hline
 \#instances & 3637 & 154 & 928 & \textbf{4030} \\
 \hline
 avg \#instances/class & \textbf{364} & 15 & 44 & 168 \\ 
 \hline
 spatial annotation & subset & all & all & all \\ 
 \hline
 \end{tabular}
 \caption{Comparison of our DALY dataset with existing action localization datasets. }
\label{tab:datasets}
\end{table*}

\subsection{Extracting human tubes} 
\label{sub:humantubes}

The first step of our approach is to extract human tubes, where human
tubes are sequences of bounding boxes following a particular person.
We propose to extract human tubes by relying on additional 
training data, i.e., the pose annotations from the MPII Human Pose
dataset~\cite{mpiipose}.  
This human pose dataset contains people in a wide range of poses,
which allows us to detect humans even when they perform actions that involve
unusual poses.

Our approach starts by detecting humans at the frame level. The human
detector relies on  the state-of-the-art Faster
R-CNN~\cite{fasterRCNN} detector trained with the MPII Human Pose
dataset.  
Once humans are detected in every frame, we track them throughout the
video. 
Relying on a tracking strategy similar to Weinzaepfel et al.~\cite{weinzaepfelICCV15},
we combine the human detection score from Faster R-CNN with an instance-level detector. 
The instance-level detector is a linear SVM learned on the features from the last fully-connected layer of Faster R-CNN. 
The tracker performs a sliding window search in the next frame 
around the location of the tracked box,
and selects the highest scored box according to a combined score of the per-frame human and instance-level detectors.
This box is refined according to the regressor branch of Faster R-CNN learned for human detection.
At every frame, the instance-level detector is updated with the selected region.

We obtain an initial human tube by tracking the highest scoring human
detection in the video sequence, both forward and backward in time. 
Having tracked this detection, we remove the human
detections that have an Intersection Over Union (IoU) above $0.3$ with
any box of this track. 
We repeat this process by selecting  and tracking the highest-scoring human detection
among remaining ones. We stop when no detections are left to examine. 

\subsection{Learning with sparse spatial supervision}
\label{sub:learn}

In this work, we consider that for each training video, only a subset of $N$ frames are annotated in each action instance. 
For our experiments we regularly sample them from the ground-truth
tubes, except in the case of DALY where only up to 5 annotations are
provided per action instance.
The human tubes are labeled as positive if they have an Intersection-over-Union (IoU) over $0.5$ with any ground-truth, 
the IoU being computed only over the $N$ annotated frames, see
Figure~\ref{fig:intro}.
In our experiments, we vary the value of $N$ from $1$ to $5$ frames. 

\subsection{Scoring human tubes}
\label{sub:features}

We consider different approaches to score the human
tubes: improved dense trajectories (IDT)
and two-stream R-CNNs (CNN), as well their combination with late fusion (CNN+IDT).

\subsubsection*{Improved Dense Trajectories}

For each human tube, we extract Improved Dense Trajectories (IDT) and aggregate them with a Fisher Vector
representation~\cite{wang:hal-01145834}. 
In more details, we start by extracting IDT for the entire 
video\footnote{\url{https://lear.inrialpes.fr/people/wang/improved_trajectories}}. 
For each descriptor type (HOG, HOF, MBHx, MBHy), we reduce its initial dimension by a factor of 2 using PCA 
 and learn a codebook of 256 Gaussians.
For each tube, we build a  Fisher Vector per descriptor type, using only the trajectories that start inside the tube.
Each of the 4 Fisher Vectors is independently power-normalized and L2-normalized~\cite{FVnorm}.
A tube is finally described by the concatenation of the 4 normalized Fisher Vectors, resulting in 102400 dimensions. 
To classify human tubes described with IDT, we learn a linear Support
Vector Machine (SVM), which have demonstrated an excellent performance
on the Fisher Vector representation~\cite{wang:hal-01145834}. We
convert SVM scores into probabilities~\cite{wu2004probability} in order to
make them comparable to CNN softmax scores.
We use as negatives all human tubes from negative videos, as well as human tubes
for which the Intersection-over-Union over human tubes labeled as positives is below
$0.5$.

\subsubsection*{Two-stream R-CNNs}

We also learn a two-stream Fast R-CNN~\cite{fastRCNN} on the tubes selected with sparse spatial supervision. 
We use as input proposals to Fast R-CNN the ones from the human tubes, i.e., the Region Proposals Network of the human detector.
We label them as positive or negative according to the overlap with the human tubes: proposals with IoU over $0.5$ with positive human tubes are labeled as positive proposals, the others as negative. 
Thanks to the human tubes, we can thus label the human proposals from all frames for training, which increases significantly the training data compared to 
training on the sparse annotated frames only.

We learn the two streams independently, one on RGB images and the
 other on flow images. Flow is computed with~\cite{Bro04a} and converted to a jpg file following~\cite{fat,weinzaepfelICCV15}.
As~\cite{TSN2016ECCV}, we initialize the weights with ImageNet pretraining~\cite{ILSVRC15} for both streams.
We perform late fusion of the scores from the RGB and the flow streams.
During testing, we score the human tubes using the softmax probability
scores averaged over all boxes of a tube. 
Note that we do not need to detect, but just to apply the classifier on the boxes from the human tubes. Thus, we do not need to learn region proposals.
Furthermore, we do regress boxes, as (a)
the human tubes are already regressed, and (b) the sparse supervision
limits the number of ground-truth boxes, i.e., of regression targets at training.

\subsubsection*{Fusion of the features}
We finally consider the fusion of the IDT and the CNN features, using late fusion.
More precisely, we score a human tube using the average of the probabilities output by (a) the SVM learned on IDT features, (b) the RGB stream of Fast R-CNN and, (c) its flow stream.

\subsection{Temporal detection}
\label{sub:tempo}

We now present how we perform temporal detection at test time for untrimmed dataset.
During training, we assume that the temporal extent of the actions is
given, and extract the human tubes for this temporal extent.
At test time, for datasets with long videos that can contain
  multiple clips such as DALY, we first split videos into clips using
an automatic shot
detector\footnote{\url{https://github.com/johmathe/Shotdetect}}. 
For datasets with only one shot per video, e.g. the UCF-Sports, J-HMDB and UCF-101 datasets, we define clips as the full videos.
We then extract the human tubes in each clip.
The temporal detection is performed using a multi-scale temporal sliding window inside each tube.
We use the same temporal lengths as~\cite{weinzaepfelICCV15} 
($\{20,30,40,...,90,100,150,300,450,600\}$ frames) and the same stride ($10$).
For the DALY dataset, which contains longer actions, we add temporal lengths of $\{900,1200,1500,1800\}$ and $\{2400,3000,3600,...,120000\}$.
In order to penalize short action detections, we score a snippet of a tube using its
CNN+IDT score minus  $\alpha / L$ where $\alpha$ is a parameter
experimentally set to $20$ and $L$ is the length of the detection.

\section{Datasets}
\label{sec:dataset}

This section describes existing action localization datasets used
in our experiments (Section~\ref{sub:datasets}) and their
limitations. We then introduce our \textbf{DALY} (Daily Action
Localization in YouTube) dataset (Section~\ref{sub:daly}).
We finally present the evaluation protocol (Section~\ref{sub:protocol}).

\begin{figure}
 \includegraphics[width=0.49\linewidth]{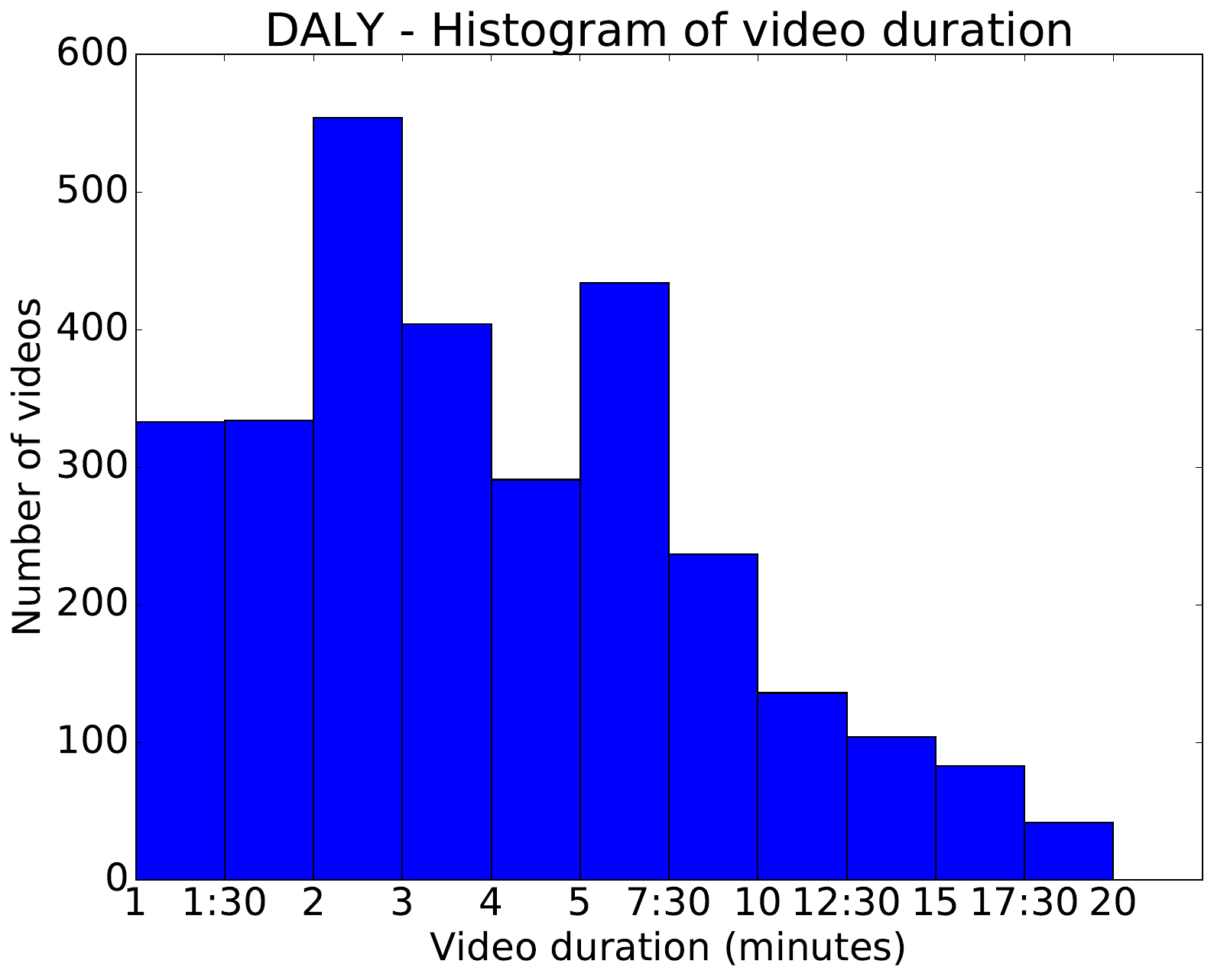}
 \hfill
 \includegraphics[width=0.49\linewidth]{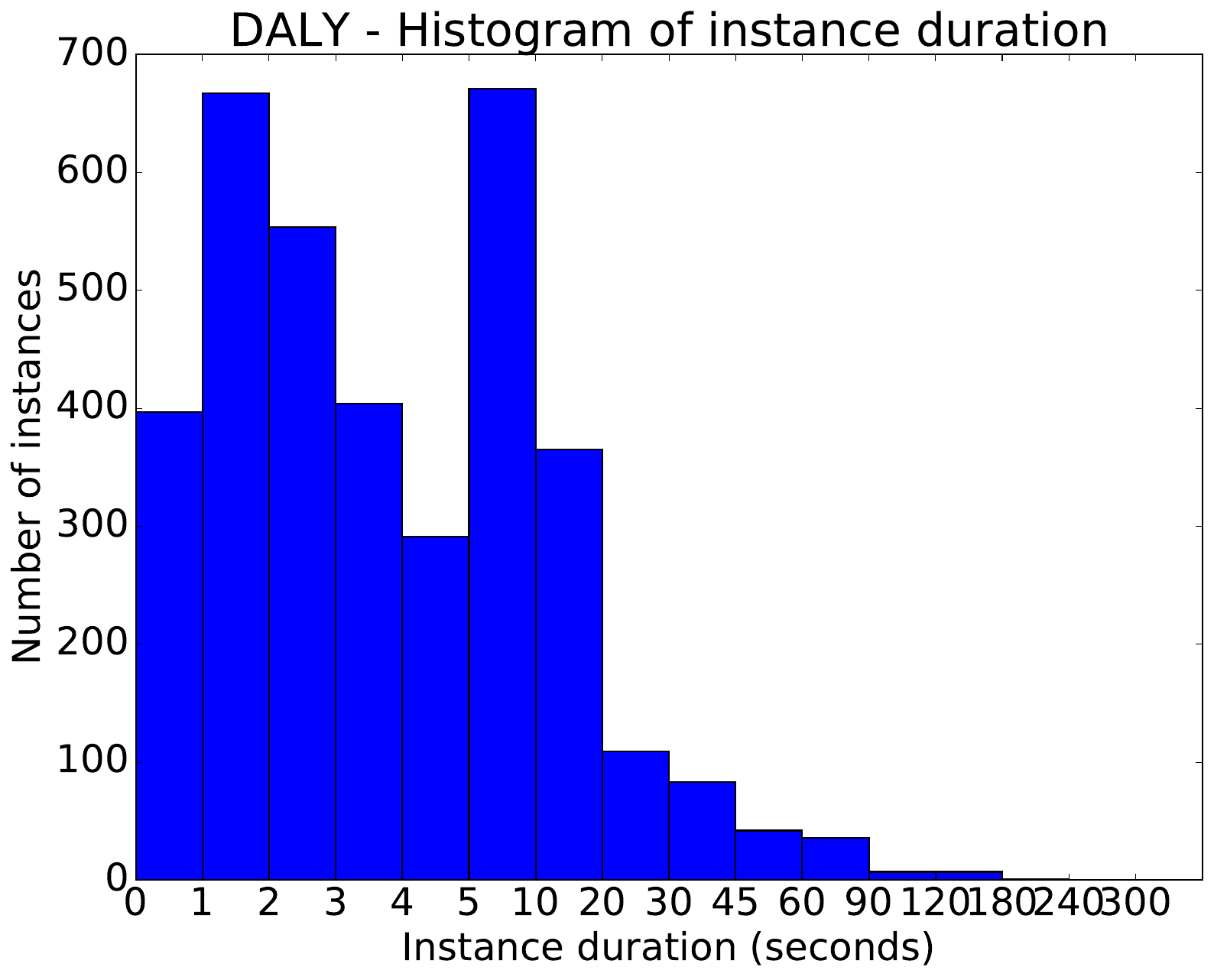}
 \caption{Histogram of duration of the videos (left) and instances (right).}
 \label{fig:histograms}
\end{figure}

\subsection{Action localization datasets}
\label{sub:datasets}

\noindent $\bullet$ The {\bf UCF-Sports} dataset~\cite{ucfsports} is limited to 150 short sports videos with 10 actions, such as \textit{diving} or \textit{running}. 
It contains an average of 15 instances per class.
Videos are trimmed to the action and every frame is annotated with a bounding box.
For each class, sequences present similarities in background, camera viewpoint and actors, reducing diversity.
We use the train/test split \mbox{defined} in~\cite{LanWM11}.

\noindent $\bullet$ The {\bf J-HMDB} dataset~\cite{jhmdb} is a subset of the HMDB benchmark~\cite{hmdb}. It contains 928 videos with 21 actions, including \textit{stand up}, \textit{run} and \textit{pour}.
The videos are trimmed to the action, are very short (1.4 sec on average), and most of them contain a single human. 
On average 44 instances are annotated per action class.
The annotations are human silhouettes in every frame. 
We use the bounding boxes around these silhouettes as ground-truth.
The dataset has 3 train/test splits. 
  
\noindent $\bullet$ The {\bf UCF-101} dataset~\cite{ucf101} contains spatio-temporal 
annotations for 24 actions in 3207 sports videos. Many videos are similar in terms of actors or background.
In contrast to UCF-Sports and J-HMDB, the detection is also temporal but the videos remain short; for half of the classes, the action lasts for more than 80\% of the video duration.
There are 3 train/test splits. Results are reported for split 1 only, as in~\cite{weinzaepfelICCV15,Saha2016,MR2RCNN}.

In summary, existing action localization datasets are limited in either diversity of actions (mainly sports),
diversity of videos (similar backgrounds and/or actors across videos), video duration (short, trimmed to the action) or number of samples per class, see Table~\ref{tab:datasets} for details.
There is a clear need for a dataset that overcomes all of these limitations if we want to 
improve spatio-temporal action localization in real-world scenarios.

\subsection{DALY dataset}
\label{sub:daly}

We introduce DALY, a dataset for {\bf D}aily {\bf A}ction {\bf L}ocalization in {\bf Y}ouTube. 
The DALY dataset consists of 31 hours of YouTube videos, with spatial
and temporal annotations for 10 everyday human actions with a total of  3.6k instances. 

We describe in this section how the DALY dataset was collected.
More precisely, we first explain the action class selection and
the filtering of the videos. We then present the spatio-temporal annotation of action instances.

\subsubsection*{Picking action classes}

In order to obtain a dataset that fairly evaluates action localization methods, action classes must have clearly defined temporal boundaries.
Ambiguities introduce noise in label and temporal annotation, which makes the evaluation unreliable.

We thus select classes for which temporal boundary guidelines can be stated precisely and concisely.
For instance, the \textit{brushing teeth} action is defined as `toothbrush inside the mouth'.
Another example is \textit{cleaning windows} for which the moment where `the tool is in contact with the window' is annotated.
The selected action categories are \textit{applying make up on lips, brushing teeth, cleaning floor, cleaning windows, drinking, folding textile, ironing, phoning, playing harmonica and taking photos/videos}, 
see Figure~\ref{fig:daly}. 

Some of those action classes were picked to have similar body movement, in order to make them hard to distinguish.
For instance,  several action classes imply motion of the hands near the head (\textit{taking photos, phoning})
or the mouth (\textit{playing harmonica, drinking, brushing teeth, applying make up on lips}).

\subsubsection*{Video collection} 
The videos are retrieved from YouTube using manually defined queries related to the action labels.
For example, the class \textit{cleaning floor} relies on queries such as `sweeping floor', `mopping floor', `cleaning floor', etc.
We only collect videos that last between 1 and 20 minutes.
A minimum duration of 1 minute ensures that temporal localization will be meaningful (in most cases, shorter videos contain only one action from the beginning to the end).
The maximum duration of 20 minutes 
is to promote ease of use and, for some future methods, avoid disproportionate computational time.
The average video duration is 3min 45s. Figure~\ref{fig:histograms} (left) shows an histogram of video duration,
and Table~\ref{tab:statclass} displays per-class average video duration.
Videos are longest for \textit{ApplyingMakeUpOnLips}, 
mainly because this action tends to be present in long and detailed make-up tutorials.

\begin{table}
  \centering
  \resizebox{\linewidth}{!}{
    \begin{tabular}{|c|c|r|r@{ }c@{ }l|}
     \hline
     class & avg video dur. & \#inst. & \multicolumn{3}{c|}{avg inst. dur.} \\
     \hline
ApplyingMakeUpOnLips & 376.8s $\pm$ 265.1 & 409 & 3.8s & $\pm$ & 3.4 \\
BrushingTeeth & 176.0s $\pm$ 120.3 & 257 & 9.2s & $\pm$ & 16.2 \\
CleaningFloor & 194.2s $\pm$ 128.7 & 187 & 14.9s & $\pm$ & 15.4 \\
CleaningWindows & 196.2s $\pm$ 131.9 & 468 & 7.1s & $\pm$ & 10.0 \\
Drinking & 202.4s $\pm$ 130.9 & 291 & 2.6s & $\pm$ & 3.0 \\
FoldingTextile & 184.1s $\pm$ 150.1 & 257 & 14.6s & $\pm$ & 22.4 \\
Ironing & 233.2s $\pm$ 183.8 & 424 & 7.2s & $\pm$ & 8.2 \\
Phoning & 217.9s $\pm$ 140.7 & 509 & 10.0s & $\pm$ & 30.3 \\
PlayingHarmonica & 190.2s $\pm$ 139.8 & 289 & 14.6s & $\pm$ & 21.8 \\
TakingPhotosOrVideos & 283.0s $\pm$ 207.3 & 546 & 3.1s & $\pm$ & 3.4 \\
\hline
all & 225.4s $\pm$ 175.7 & 3637 & 7.9s & $\pm$ & 16.6  \\
     \hline
    \end{tabular}
  }
 \caption{Statistics for each class showing the video duration (average and standard deviation), the number of instances, and the instance duration (average and standard deviation). }
 \label{tab:statclass}
\end{table}

\begin{figure}
 \centering
 \includegraphics[width=\linewidth]{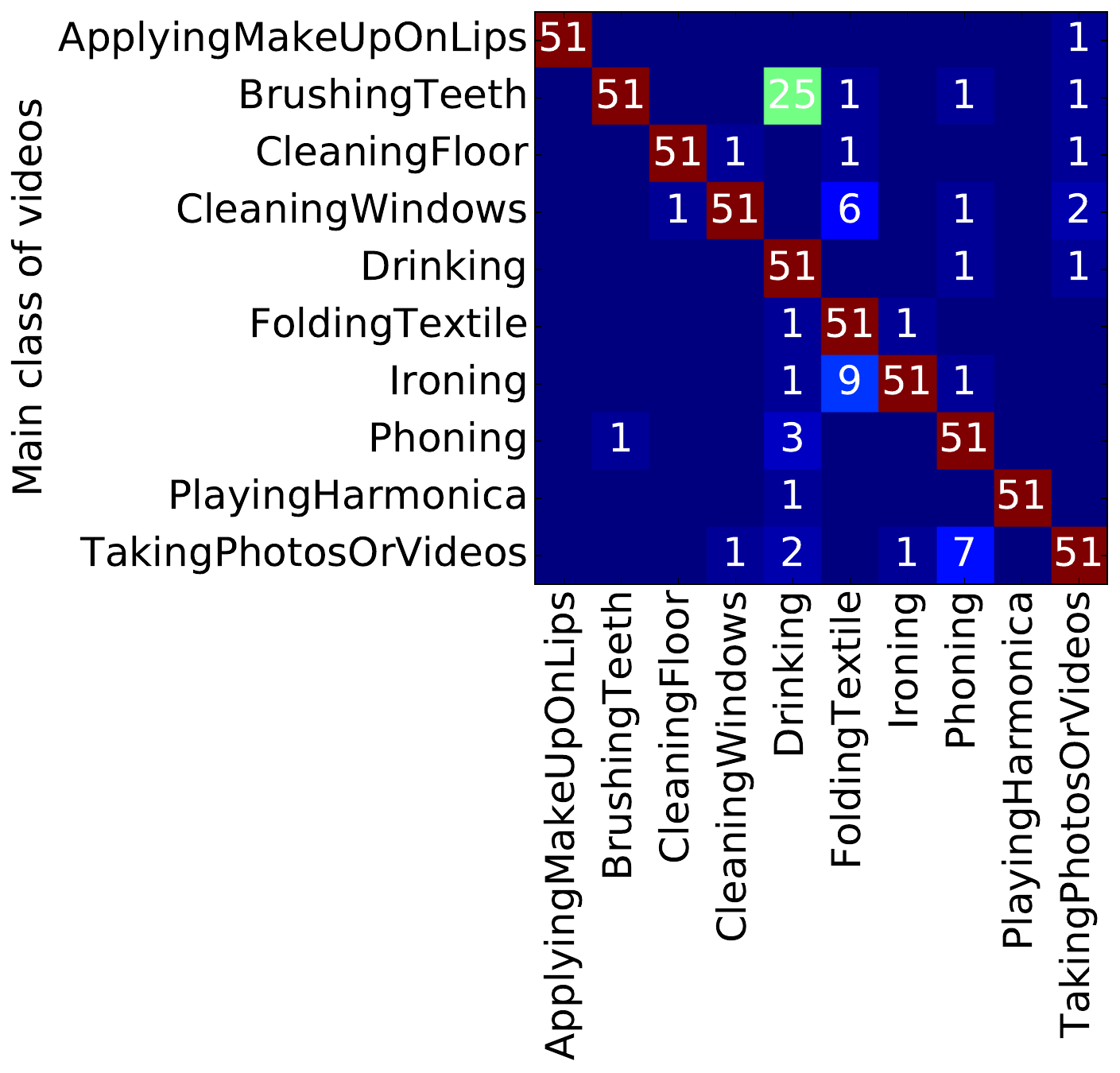}
\caption{Statistics of multiple classes per video. 
 Each row comprises the 51 videos downloaded for a given class, each column counts the videos containing at least one instance of the column class.}
 \label{fig:labelmixing}
\end{figure}

\begin{figure*}
 \centering
 \includegraphics[width=0.245\linewidth]{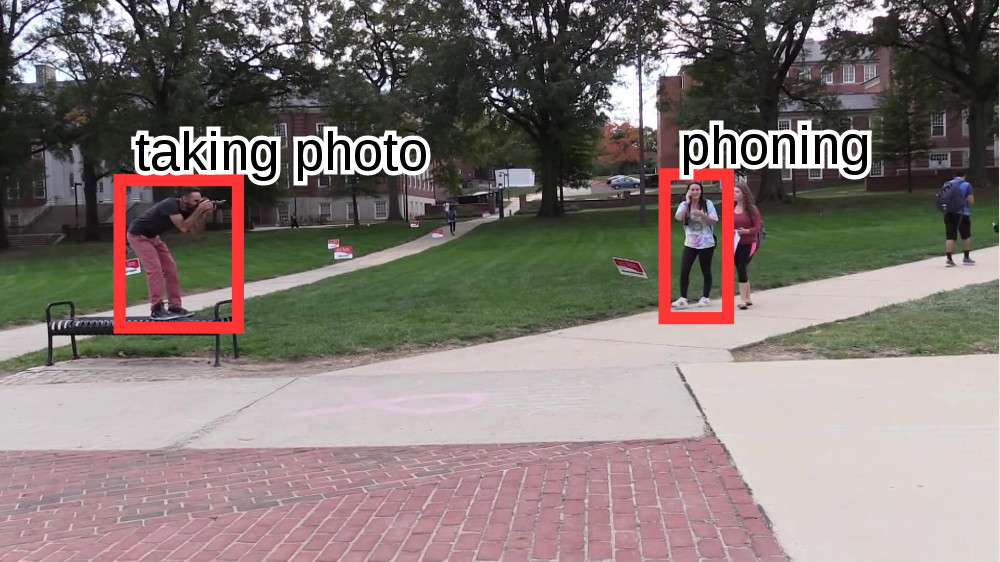}
 \hfill
 \includegraphics[width=0.245\linewidth]{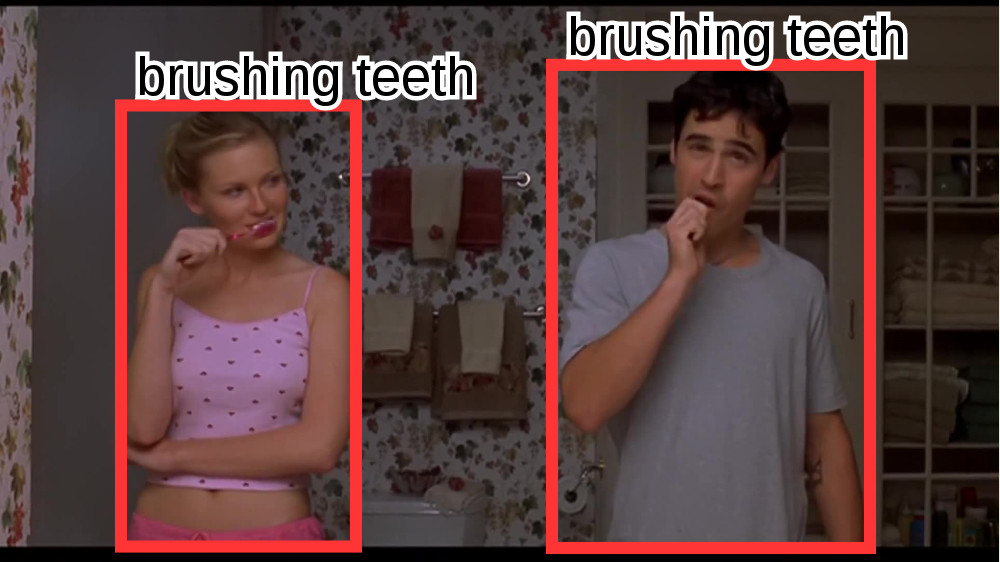} \hfill
 \includegraphics[width=0.245\linewidth]{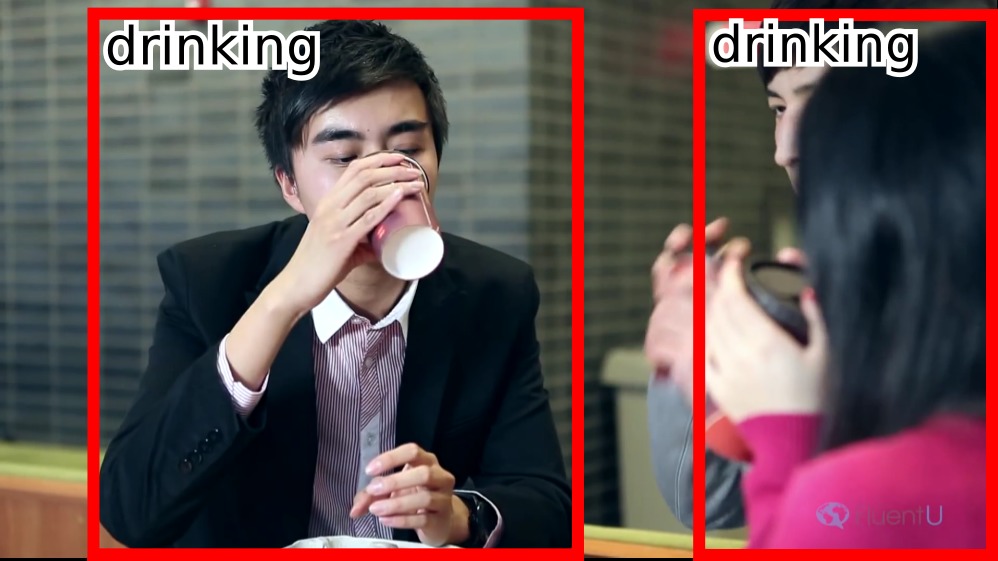}
 \hfill
 \includegraphics[width=0.245\linewidth]{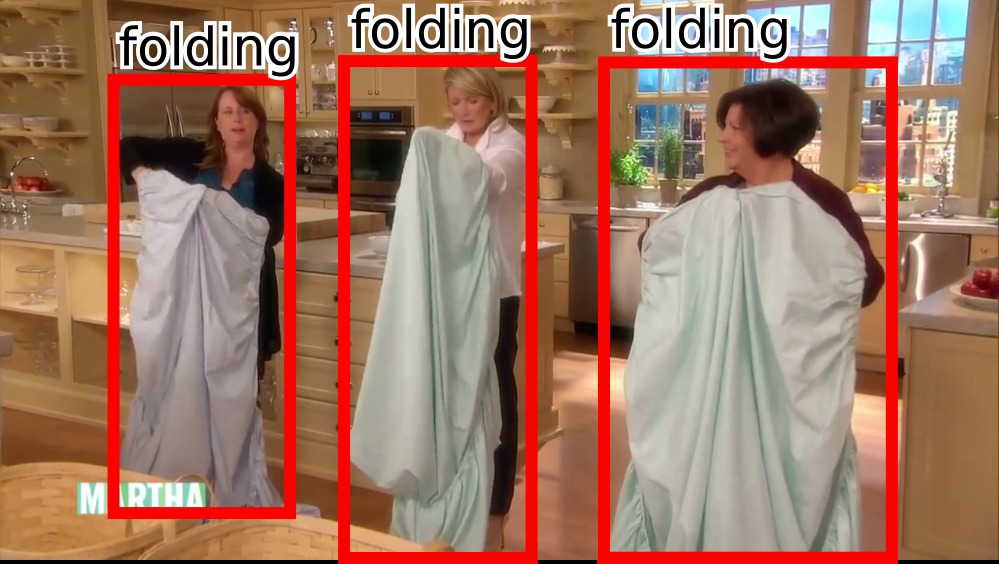}
 \caption{Example frames from the DALY dataset with simultaneous actions.}
 \label{fig:mult_actions}
\end{figure*}

\begin{figure*}
\includegraphics[width=0.245\linewidth]{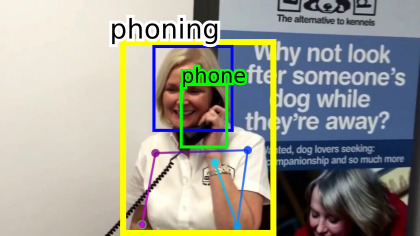} \hfill
\includegraphics[width=0.245\linewidth]{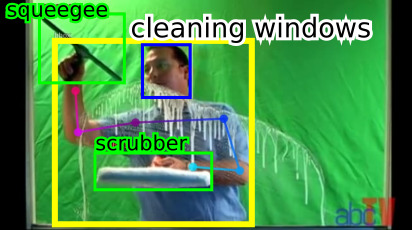} \hfill
\includegraphics[width=0.245\linewidth]{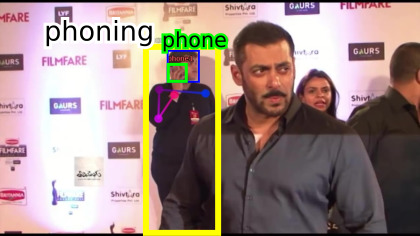} \hfill
\includegraphics[width=0.25\linewidth]{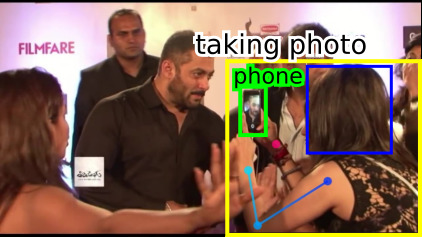} 
\caption{Example of spatial annotation from the DALY dataset. In addition to the bounding box around the actor (yellow), we also annotate the objects (green) and the pose of the upper body (bounding box around the head in blue and joint annotation for shoulders, elbows and wrists).}
\label{fig:spatial}
\end{figure*}

Videos are filtered to remove cartoons, slideshows, actions performed by animals and first-person viewpoints.
We also remove videos in which the human is not visible when the action occurs,
for instance when the camera focuses on the mop while performing the \textit{cleaning floor} action.

We keep 51 videos for each action class, where each video contains at least one instance of the action class.
In total, this corresponds to 31 hours of video or 3.3 million frames.
This represents 300 times more frames than UCF-Sports, 10 times more than J-HMDB
and 6 times more than UCF-101, see Table~\ref{tab:datasets}.
Our dataset contains 510 videos in total, 
downloaded from YouTube and used as-is. Thus, those videos may
contain many shots in contrast to existing datasets that have trimmed or almost-trimmed videos.
Using an off-the-shelf automatic shot detector\footnote{\url{https://github.com/johmathe/Shotdetect}}, we obtain a total of 8133 clips.
We generate a split with 31 training videos and 20
test videos for each class, ensuring that videos uploaded by the same user are in the same set.

The selected action classes are sufficiently common such that multiple action classes can be found in a single video, see Figure~\ref{fig:labelmixing}.
Each row of the matrix displays the 
presence of other action classes in the 51 videos of a given class.
For example, out of the 51 videos selected for the class
\textit{brushing teeth}, 25 videos also contain \textit{drinking} instances.
There is overlap between \textit{ironing} and \textit{folding textile};
\textit{Taking photos or videos}, \textit{phoning} and \textit{drinking} also occur together, as \textit{taking photos} is mostly performed outdoors, where other people are \textit{phoning} or \textit{drinking}. 
In some cases, several instances are happening simultaneously, see Figure~\ref{fig:mult_actions}.
We annotate all occurrences of the 10 classes exhaustively.

\subsubsection*{Temporal annotation} 

Videos are carefully annotated by members of our research team
with the \textit{begin} and \textit{end} time for all action instances.
Precise guidelines are established before annotation. 
For example, the \textit{phoning} action lasts
as long as the phone remains close to the ear.
In case of a shot change during an action, we annotate it as two separate instances and set a `shotcut' flag on the second instance.
DALY contains 3637 action instances in total.
Compared to the average video duration of about 4 minutes, actions are short with an average duration of 8 seconds. Figure~\ref{fig:histograms} (right) shows an histogram of instance duration and per-class statistics are shown in Table~\ref{tab:statclass}.
Most instances are shorter than 10 seconds, however DALY also contains instances of several minutes.
Some classes have very brief instances (e.g. \textit{drinking}), others are longer on average (e.g. \textit{brushing teeth}).  
For actions that usually last many seconds (such as \textit{cleaning window} or \textit{brushing teeth}), a short instance duration can be explained by video editing: the uploader may have edited the videos, creating small shots that contain the action.
The videos are untrimmed, 75\% of the frames do not contain any of our 10 actions.

For each action instance, we add a set of `flags' that state if an action is:
(a)~small compared to the image, (b)~very big compared to the image (zoomed in), (c)~largely occluded at some point, (d)~outside the camera's field of view at some point.
These flags will allow future work to focus on these challenging cases.

Not included in the above count are around 200 instances annotated as ambiguous or mirror reflections. 
The `ambiguous' flag is applied when it is unclear
whether the action is genuinely performed or not, e.g. the \textit{toothbrush} is put inside the mouth without actually \textit{brushing the teeth}, or the \textit{squeegee} is in contact with the window but the actor is mainly talking instead of \textit{cleaning windows}.
In addition to the main actor performing the action, his
reflection can sometimes be seen in a mirror or a window. In this
case,  we have added the flag `mirror reflection'. 
This is in particular the case for \textit{brushing teeth} that often
occurs in a bathroom in front of a mirror.
These annotations of ambiguous cases and mirror reflections are ignored during evaluation, following the Pascal VOC protocol~\cite{Everingham10}.
In other words, we do not register these cases as missing positives or false positives.

\subsubsection*{Spatial annotation}

An action is present in 700k frames out of 3.3M frames total.
Annotating all of them would be very time consuming and
clearly does not scale up to even larger collections.

Thus, we subsample the frames for spatial annotation.
For each temporal instance, we pick 5 frames uniformly sampled over
time, with a maximum of 1 frame per second. 
For each frame, annotators are asked to draw a bounding box around the
actor, a bounding box around the object(s) involved in the action (e.g.
the glass/cup for \textit{drinking}), and the pose of the upper body
of the actor (bounding box around the head and keypoints for
shoulders, elbows and wrists). 
Some of the spatial annotations are completed by external workers, but
all of them are reviewed in-house and adjusted when necessary.
Figure~\ref{fig:spatial} shows a few examples of spatial annotation.

\subsection{Evaluation protocol}
\label{sub:protocol}

We measure detection performance using the standard mean Average Precision (mAP) metric.
Following the Pascal VOC protocol~\cite{Everingham10}, a detection is considered `correct' if
(a) the intersection over union (IoU) with the ground-truth
is above a threshold $\delta$, and (b) the detection is correctly classified.
Duplicate detections are considered as false positives.
The IoU between two spatio-temporal tubes is defined as the IoU in the temporal domain 
multiplied by the average of the spatial IoU between boxes averaged over all frames in the temporal intersection.
For the DALY dataset, the averaged spatial overlap is computed only over the annotated frames in the temporal interval. 
Average Precision (AP) is computed for each class, and the mean over all classes (mAP) is reported.
The IoU threshold $\delta$ is set to $0.5$ when measuring spatial localization in trimmed clips and $0.2$ for detection in space and time unless stated otherwise. We denote by mAP@$\delta$ the mAP computed at an IoU threshold $\delta$.

\section{Evaluating human tubes}
\label{sec:tube}

\begin{figure*}
 \centering
 \includegraphics[width=0.24\linewidth,height=0.15\linewidth]{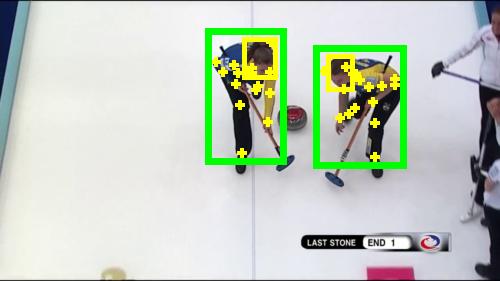}
 \hfill
 \includegraphics[width=0.24\linewidth,height=0.15\linewidth]{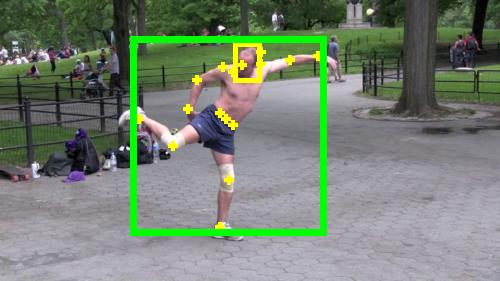}
 \hfill
 \includegraphics[width=0.24\linewidth,height=0.15\linewidth]{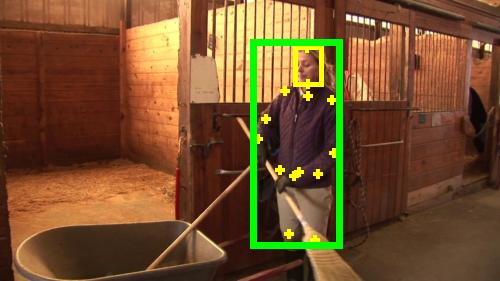}
 \hfill
 \includegraphics[width=0.24\linewidth,height=0.15\linewidth]{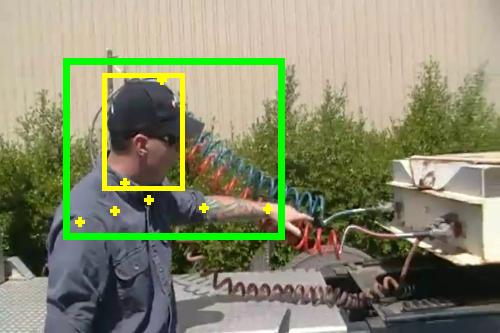}
 \caption{Example of training examples from the MPII Human Pose dataset~\cite{mpiipose}. We display the joint annotation in yellow and the bounding box used for training in green.}
 \label{fig:mpii}  
\end{figure*}

\begin{figure*}
 \centering
 \includegraphics[width=0.24\linewidth,height=0.15\linewidth]{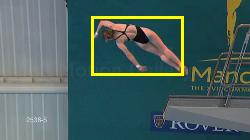}
 \hfill
 \includegraphics[width=0.24\linewidth,height=0.15\linewidth]{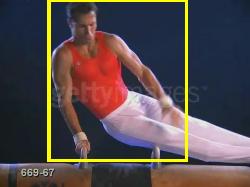}
 \hfill
 \includegraphics[width=0.24\linewidth,height=0.15\linewidth]{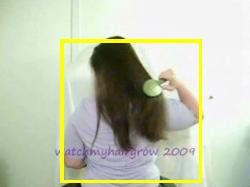}
 \hfill
 \includegraphics[width=0.24\linewidth,height=0.15\linewidth]{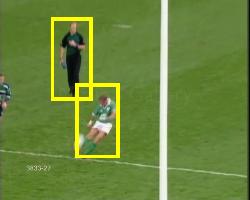}
 \caption{Example results of our human detector which consists of Faster R-CNN~\cite{fasterRCNN} trained on the Human MPII Pose dataset~\cite{mpiipose}. The first, second and fourth examples come from the UCF-Sports dataset, the third one from J-HMDB.
}
 \label{fig:humans}  
\end{figure*}

In this section we experimentally evaluate our human tubes.
We first evaluate the human detector in Section~\ref{sub:humandetect},
and then the human tubes in Section~\ref{sub:humaneval}. 
We also give implementation details.

\subsection{Human detector}
\label{sub:humandetect}

\subsubsection*{Implementation details} 
We use the state-of-the-art detector Faster R-CNN~\cite{fasterRCNN} to
train our human detector. 
We use the approximate joint training\footnote{\url{https://github.com/rbgirshick/py-faster-rcnn}} 
with the VGG-16 network architecture~\cite{vgg}. 
In Faster R-CNN, proposals are generated from anchors of size 8, 16 and 32 pixels on features maps with a stride of 16 pixels. We add a smaller scale of 4 pixels to
the anchors, as it improves robustness to
small humans. 
We found that this consistently boosts performance, i.e., the recall at an IoU threshold 0.5 for detections with a score over 0.5 is increased by 0.2\% on UCF-Sports, 1.3\% on J-HMDB and 3.7\% on UCF-101.
Other parameters are kept similar to the original implementation.

\subsubsection*{External training data} 
We use the MPII Human Pose dataset~\cite{mpiipose} to obtain training data with sufficient variability.
The publicly available training set used here contains 28k annotated poses,
including a bounding box around the head and joint positions. 
The images come from around 4000 videos, selected to contain around 500 different activities.
We obtain a bounding box for each person
by taking the box containing the head and all visible joints, with a fixed additional margin of 20 pixels.
The bounding boxes are thus not perfect, see Figure~\ref{fig:mpii}.
For instance, they can be slightly too large (top of bounding boxes from left image)
or may not cover the extremity of the limbs (second image).
The bounding boxes may also be cropped if some joints are not visible (last two images).
Nevertheless, this dataset remains large enough and offers a huge variability in term of poses.
It is thus well suited for training an accurate human detector.

\begin{table}
\resizebox{\linewidth}{!}{
\begin{tabular}{|c||c|c|c|c|}
\hline
  score th. & UCF-Sports & J-HMDB & UCF-101 & DALY \\
 \hline
 0.5 & 90.5\% (33) &  91.4\% (27) &  58.9\% (15) &  81.8\% (19) \\
 0.1 & 94.0\% (51) &  95.1\% (43) &  66.7\% (24) &  90.1\% (56) \\  
\hline
 \end{tabular}
}
 \caption{Recall@0.5 of the human detections when thresholding them at a score over $0.5$ or $0.1$, on all annotated frames from the UCF-Sports, J-HMDB, UCF-101 and DALY datasets. The number in parenthesis indicates the average number of detections per frame before non-maximum suppression.}
 \label{tab:humaneval}
\end{table}

\subsubsection*{Evaluation}
Figure~\ref{fig:humans} shows some results of our human detector.
It is robust to unusual poses (first two examples), to humans that
are not fully visible (third image), and can detect multiple
people (right example).  
To numerically evaluate the detection performance, we measure 
Recall@0.5, i.e., the ratio of ground-truth boxes for which at least
one human detection has an Intersection Over Union (IoU) over $0.5$. 
We consider as human detections the boxes for which the human probability is over $0.5$. 
Table~\ref{tab:humaneval} reports the Recall@0.5 on several datasets.
We can see that our human detector performs well on all of them, with over $90\%$ on the UCF-Sports and J-HMDB datasets.
On UCF-101 we obtain a lower recall, which can be explained by the
low quality (and high compression) of the videos. On the DALY
dataset, we obtain a recall of around $80\%$ and find that most missed humans are
small or occluded.
We also report the Recall@0.5 at a lower score threshold of $0.1$ in Table~\ref{tab:humaneval} and we can see that the recall increases by $4$ to $9\%$.

\subsection{Evaluation of human tubes}
\label{sub:humaneval}

\begin{table}
\centering
\begin{tabular}{|c@{ }c|r@{ }l|r@{ }l|r@{ }l|}
  \hline
  &  & \multicolumn{2}{c|}{UCF-Sports} & \multicolumn{2}{c|}{J-HMDB} & \multicolumn{2}{c|}{UCF-101} \\
  \hline
        ImageNet & (I) &         29.9\%  & (5.1) &   63.1\% & (2.5) &   10.4\% & (4.3) \\
      Pascal VOC & (I) &         69.5\%  & (3.4) &   92.2\% & (1.5) &   30.7\% & (2.8) \\
            MPII & (I) & \textbf{97.4\%} & (2.3) &   95.2\% & (1.2) &   \textbf{65.5\%} & (2.6) \\
            MPII & (H) &         96.8\%  & (3.2) &   95.3\% & (1.4) &   59.1\% & (3.2) \\
          MPII & (I+H) & \textbf{97.4\%} & (2.7) &   \textbf{95.5\%} & (1.3) &   65.0\% & (3.0) \\ 
  \hline
	     \multicolumn{2}{|c|}{Linking} &   96.1\% & (267.6) &   97.7\% & (276.2) &   53.9\% & (223.3) \\        
  \hline 
  \end{tabular}
 \caption{Recall@0.5 of our human tubes with variants of our tracker on all videos from the UCF-Sports, J-HMDB and UCF-101 datasets. The number in parenthesis indicates the average number of tubes per video.
 We study the impact of the training data, as well as the impact of
 the human detection score (H) and the instance-specific detection
 score (I) in the tracking strategy. We also compare our tracker to a linking approach. }
 \label{tab:tubefeat}
\end{table}

\begin{figure*}
\includegraphics[width=0.24\textwidth]{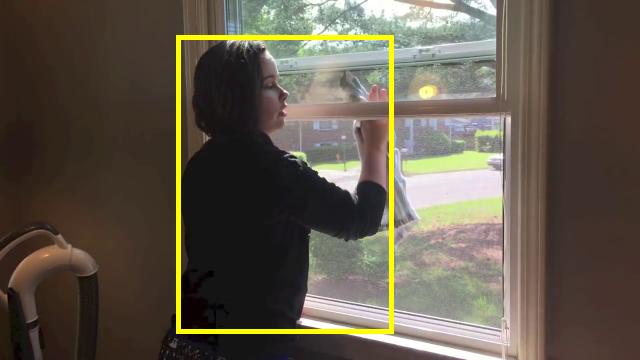} \hfill
\includegraphics[width=0.24\textwidth]{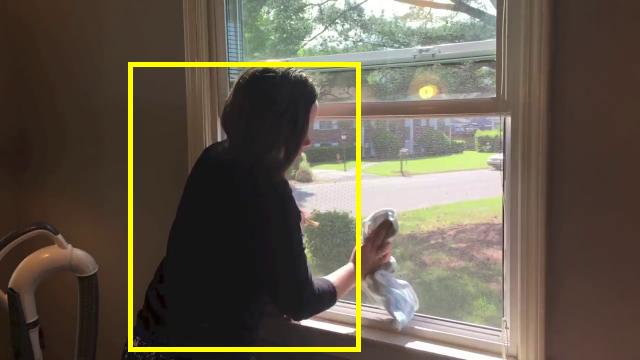} \hfill
\includegraphics[width=0.24\textwidth]{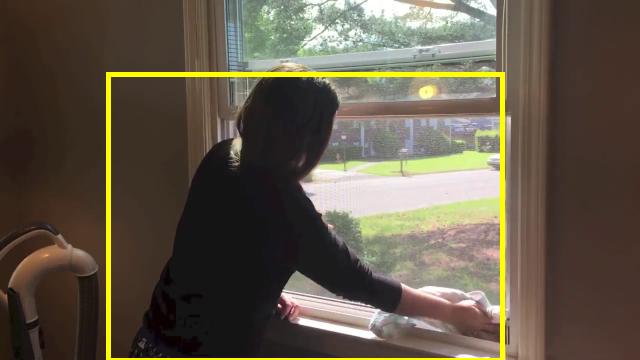} \hfill
\includegraphics[width=0.24\textwidth]{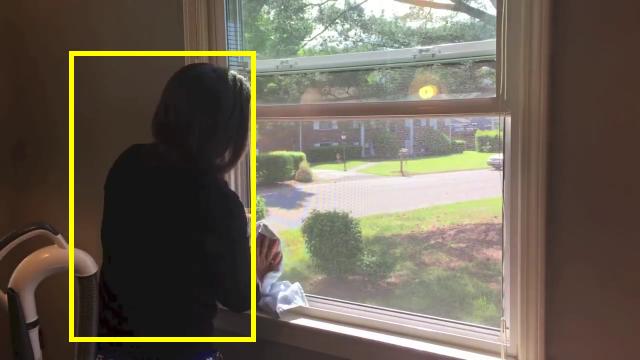} \\[0.2cm]
\includegraphics[width=0.24\textwidth]{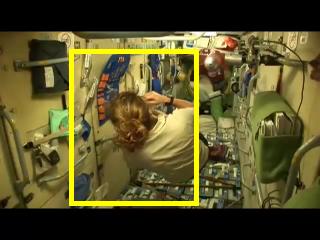} \hfill
\includegraphics[width=0.24\textwidth]{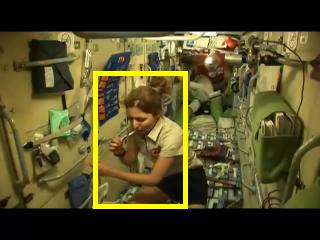} \hfill
\includegraphics[width=0.24\textwidth]{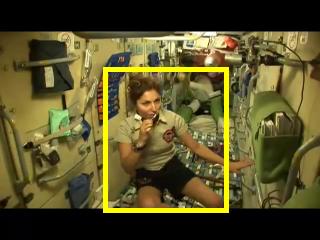} \hfill
\includegraphics[width=0.24\textwidth]{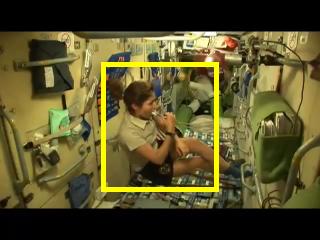} \\[0.2cm]
\includegraphics[width=0.24\textwidth]{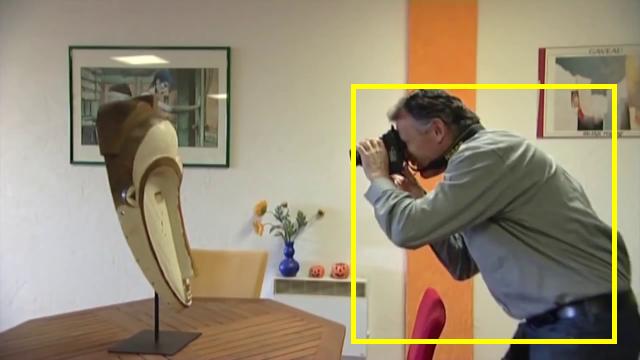} \hfill
\includegraphics[width=0.24\textwidth]{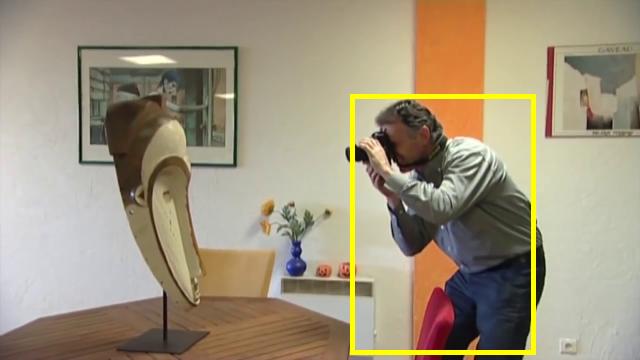} \hfill
\includegraphics[width=0.24\textwidth]{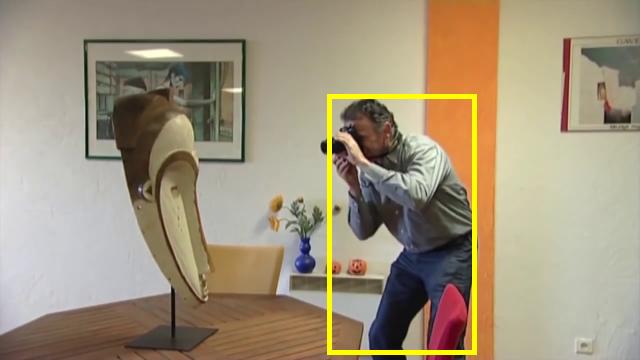} \hfill
\includegraphics[width=0.24\textwidth]{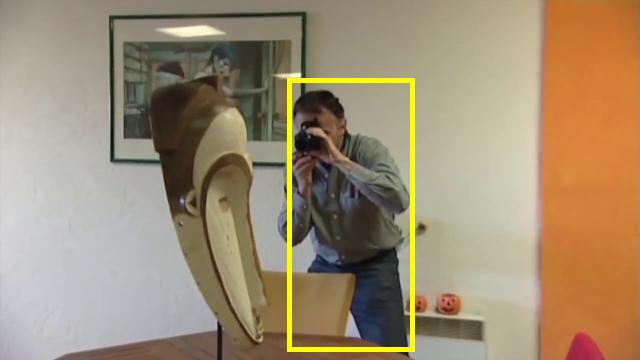} \\[0.2cm]
\includegraphics[width=0.24\textwidth]{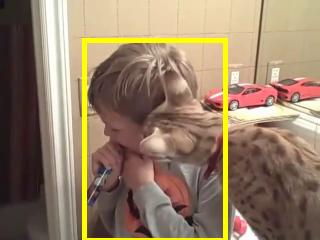} \hfill
\includegraphics[width=0.24\textwidth]{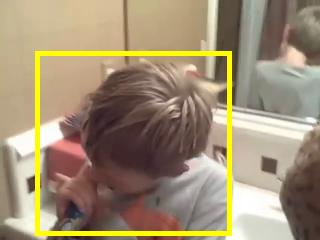} \hfill
\includegraphics[width=0.24\textwidth]{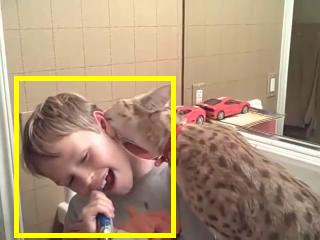} \hfill
\includegraphics[width=0.24\textwidth]{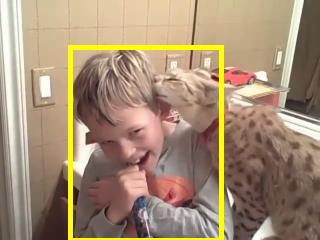} \\[0.2cm]
\includegraphics[width=0.24\textwidth]{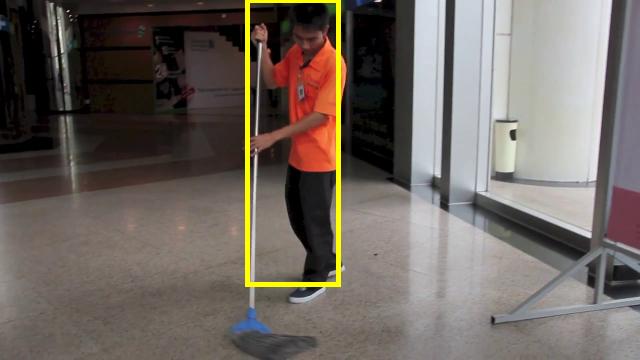} \hfill
\includegraphics[width=0.24\textwidth]{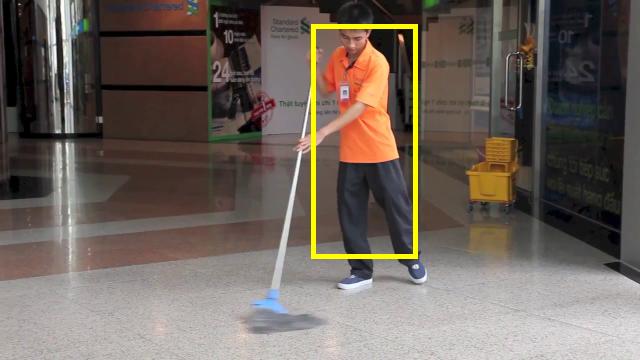} \hfill
\includegraphics[width=0.24\textwidth]{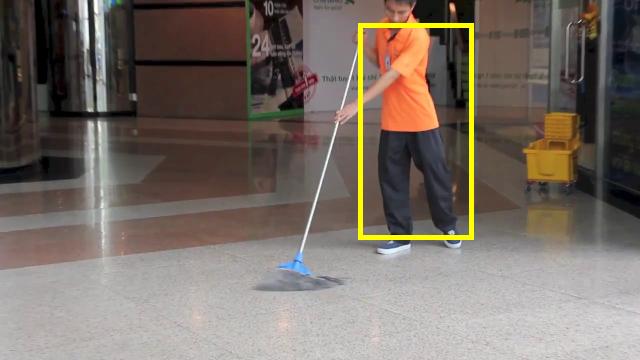} \hfill
\includegraphics[width=0.24\textwidth]{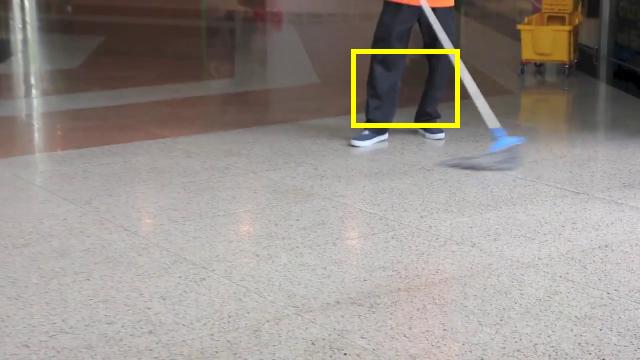} \\[0.2cm]
\includegraphics[width=0.24\textwidth]{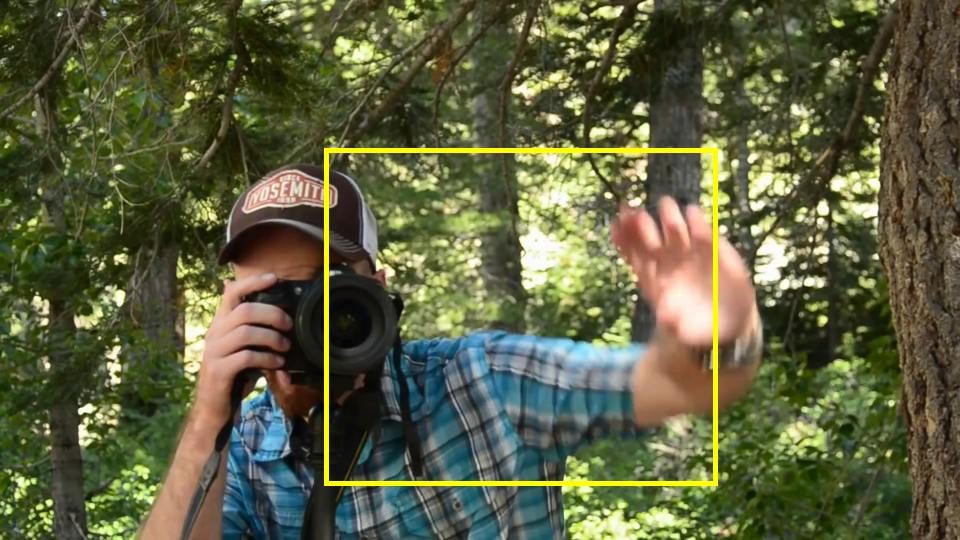} \hfill
\includegraphics[width=0.24\textwidth]{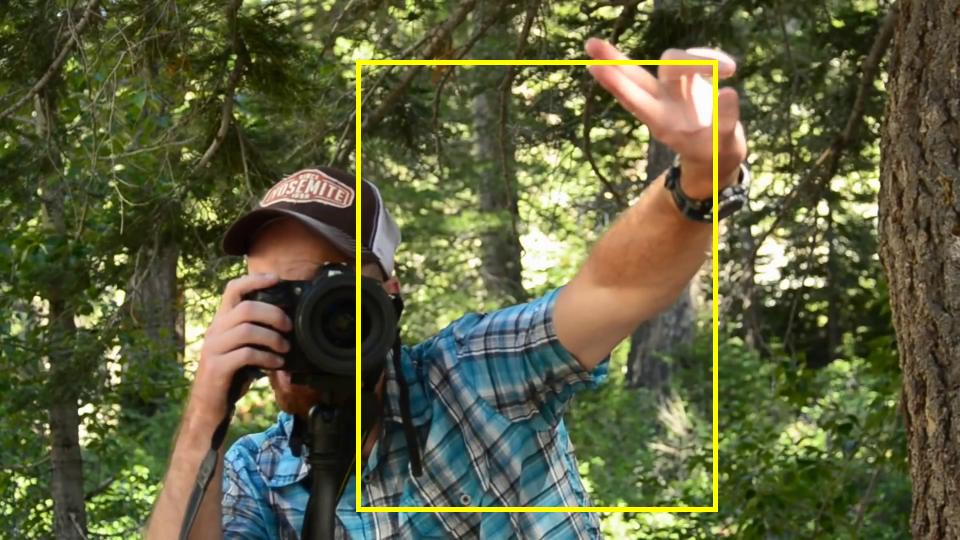} \hfill
\includegraphics[width=0.24\textwidth]{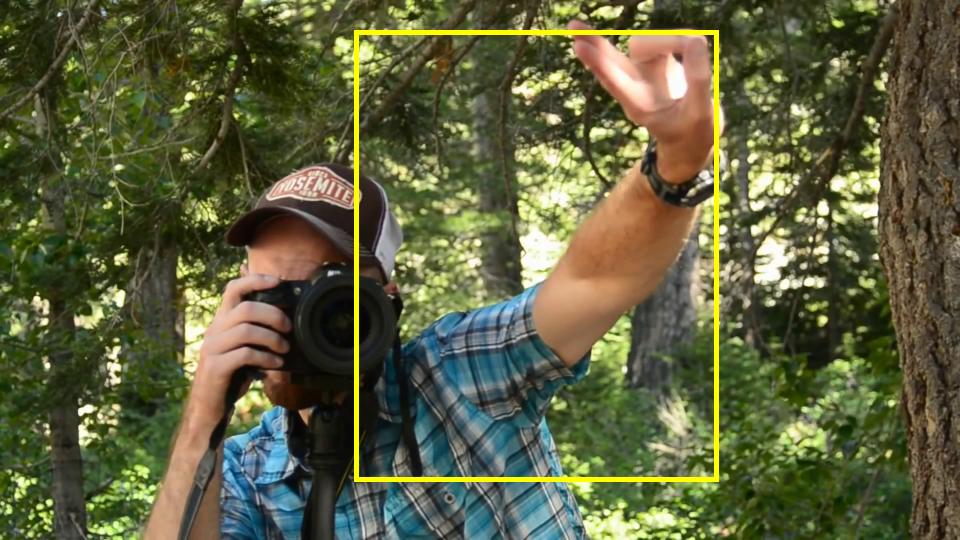} \hfill
\includegraphics[width=0.24\textwidth]{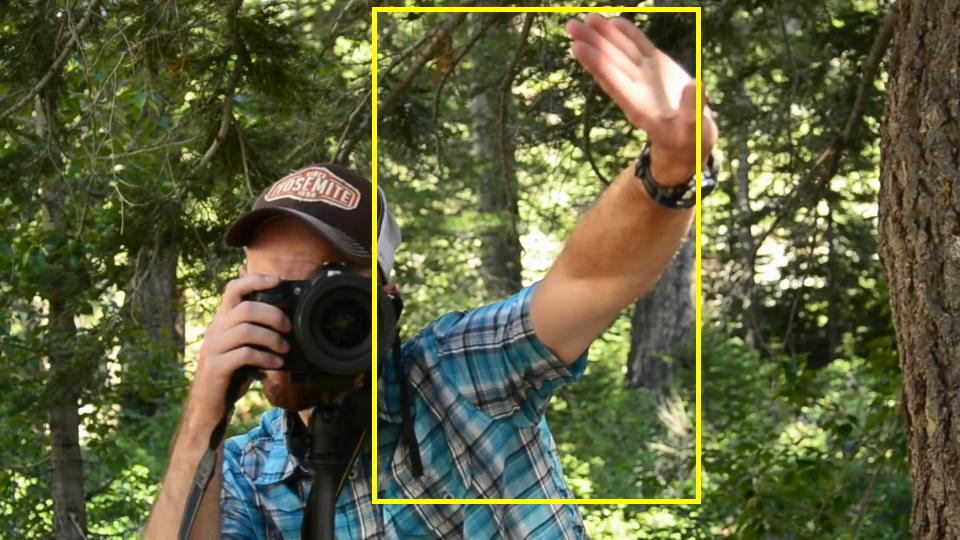} \\
\caption{Example of human tubes with successful human tube extraction in the first four rows, 
and some failure cases in the last two rows. 
Failures are caused by partial visibility of the human (end of fifth row) and missed human detection caused by an occluding camera (last row). }
\label{fig:humantubes}
\end{figure*}

\subsubsection*{Implementation details}

The instance-level detector is a linear SVM on the last fully-connected layer of Faster R-CNN which has 4096 dimensions.
The scores are converted to probabilities~\cite{wu2004probability}.
During initialization and update, we use the tracked box as positive samples.
As negatives, we use the human proposals that have (a) an IoU below $0.1$ and (b) a probability score over $0.1$, i.e., hard negatives.
For the sliding window performed in each frame, we use 5 different widths and heights equal to $\{80\%,90\%,100\%,110\%,120\%\}$ of the original size.
In space, we use a grid of size $9 \times 9$ with stride $16~pixels$ at each scale.
Launching the tracker multiple times on a clip can be performed efficiently thanks to shared convolutional layers between all regions of a frame in Faster R-CNN. 
When detecting the humans, we can keep in cache the last convolutional layer (the one just before the Region-of-Interest pooling layer) as well as the features of the last fully connected layer. The first one will be used each time we perform a sliding window in the same frame, as the computation until the RoI pooling will be exactly the same. The second one will be used as the negative features for training the SVM.

\subsubsection*{Evaluation}
To evaluate our human tubes we use Recall@0.5 at the tube level, i.e., the ratio of ground-truth instances for which at least one tube has an IoU over 0.5.
We first measure the impact of the human-level detector and the instance-level detector, as well as the importance of training data. 
For the untrimmed UCF-101 dataset, we truncate all tracks to the ground-truth duration of the actions.
Table~\ref{tab:tubefeat} shows a comparison of multiple variants of
the tracker with different pre-training sets for the network:
(a) ImageNet (image classification task)~\cite{ILSVRC15}, (b) Pascal VOC (object detection task)~\cite{Everingham10} and (c) MPII Human Pose.
The tags (I) and (H) indicate that the instance-level and/or the human-level detectors are used in the tracking process.
First, we can see that using features trained for human detection (I.e., on the MPII Human Pose dataset) is crucial:
the performance significantly drops when using
features from ImageNet or Pascal VOC.
The fact that the network is pre-trained for human detection allows to
effectively learn an instance-level detector for human tracking.
Removing the instance-level detector decreases the performance by a few percent.
Indeed, when multiple humans are present, relying only on the human-level detector may lead to drifting.
The instance-level detector alone and its combination with the human-level detector give a similar performance.
We are able to reach a recall of more than 95\% with only 2 tubes on
average per video, for UCF-Sports and J-HMDB.  
On the UCF-101 dataset, we obtain a Recall@0.5 of $65\%$ with 3 tubes on average. 
We explain this lower recall by the lower performance of the human detector, itself caused by the low quality of the videos that contain huge compression artifacts,
as well as the fact that the humans are smaller.
We also compare to a linking strategy similar
to~\cite{fat,Saha2016,MR2RCNN}. We can observe that the tracker reaches a higher recall, in particular for UCF-101. 
There is almost no difference on J-HMDB, which can be explained by the fact that the videos are extremely short and most of them contain only one human.

\begin{figure*}[t]
 \includegraphics[width=0.24\linewidth]{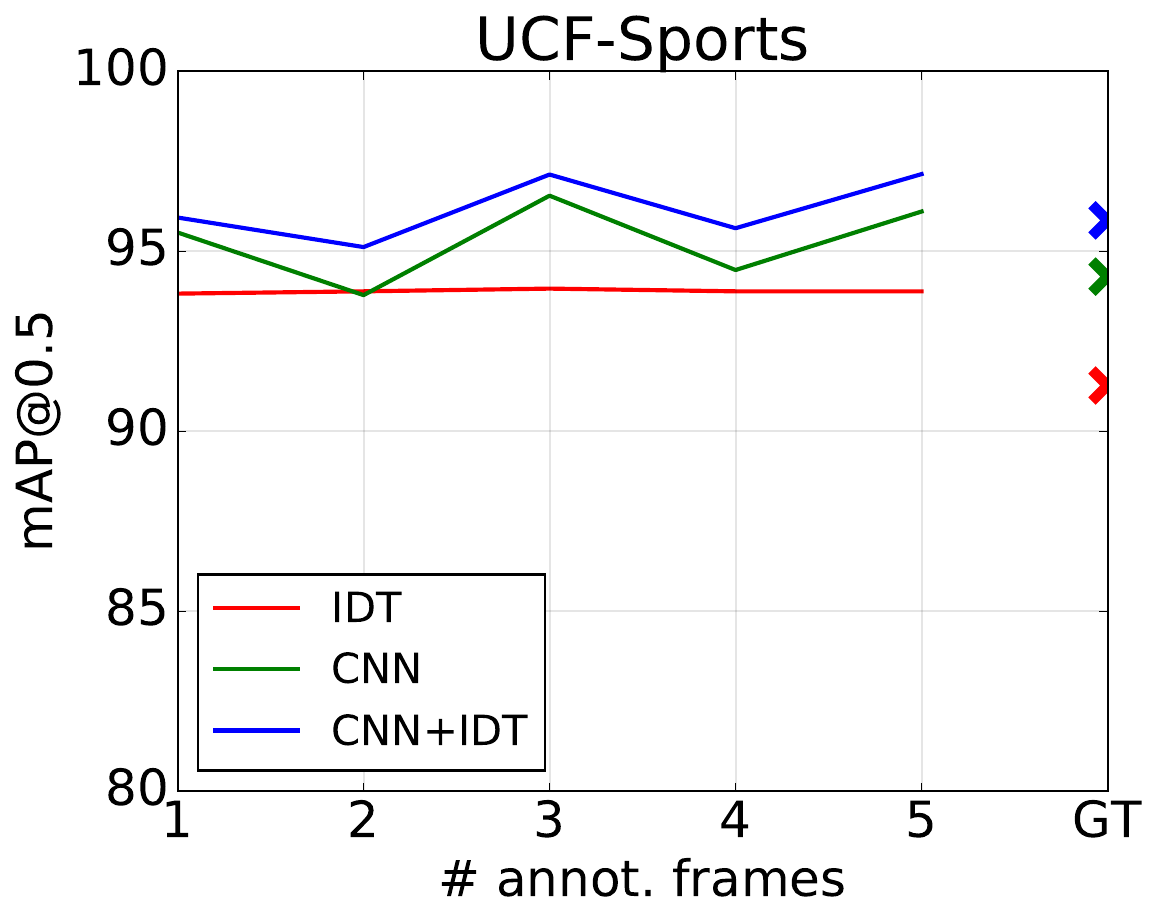}
 \hfill
 \includegraphics[width=0.24\linewidth]{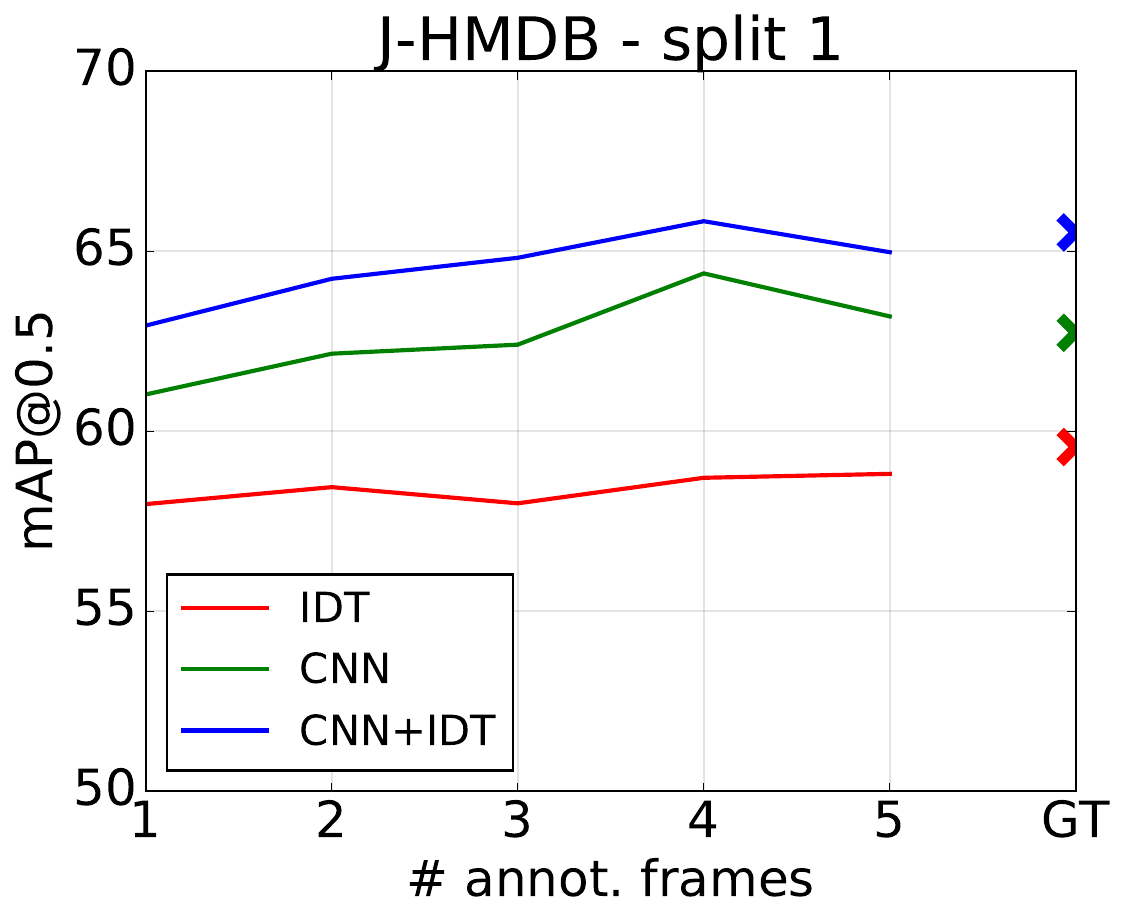}
 \hfill
 \includegraphics[width=0.24\linewidth]{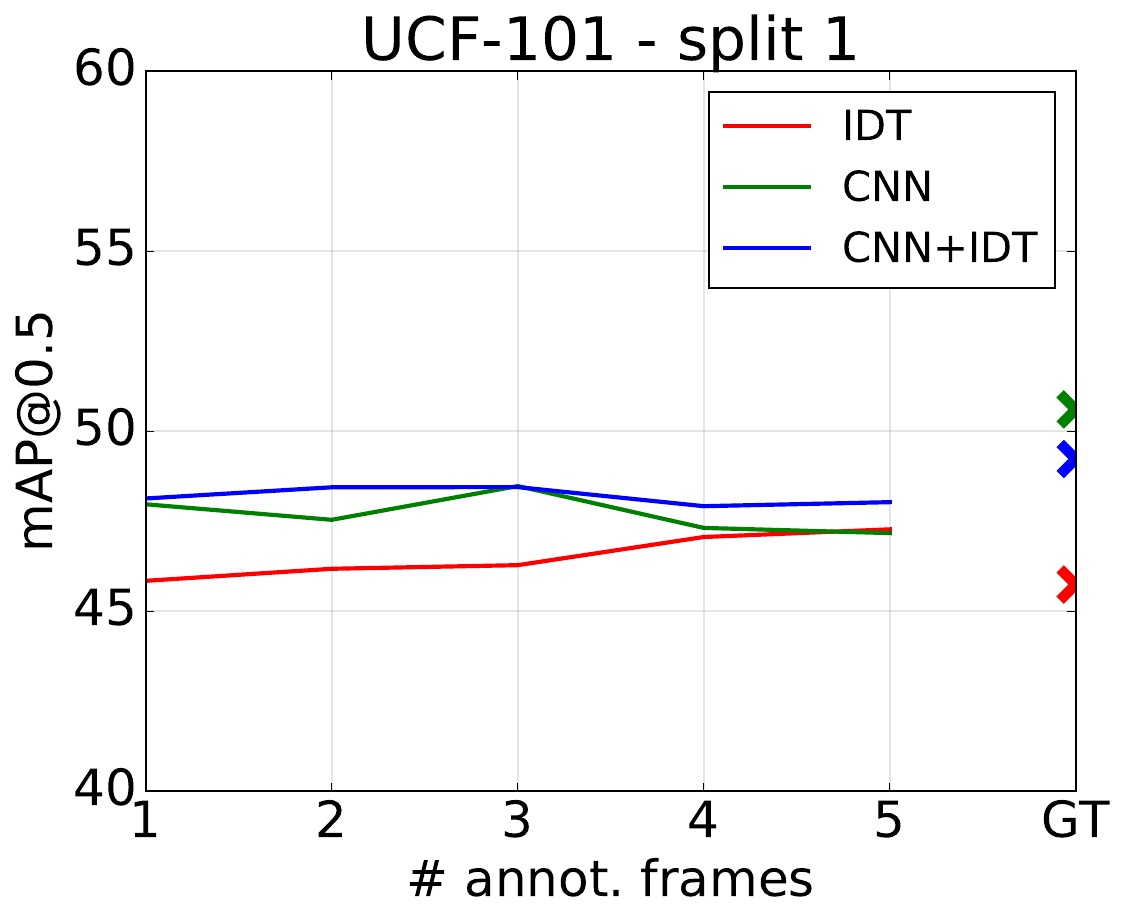}
 \hfill
 \includegraphics[width=0.24\linewidth]{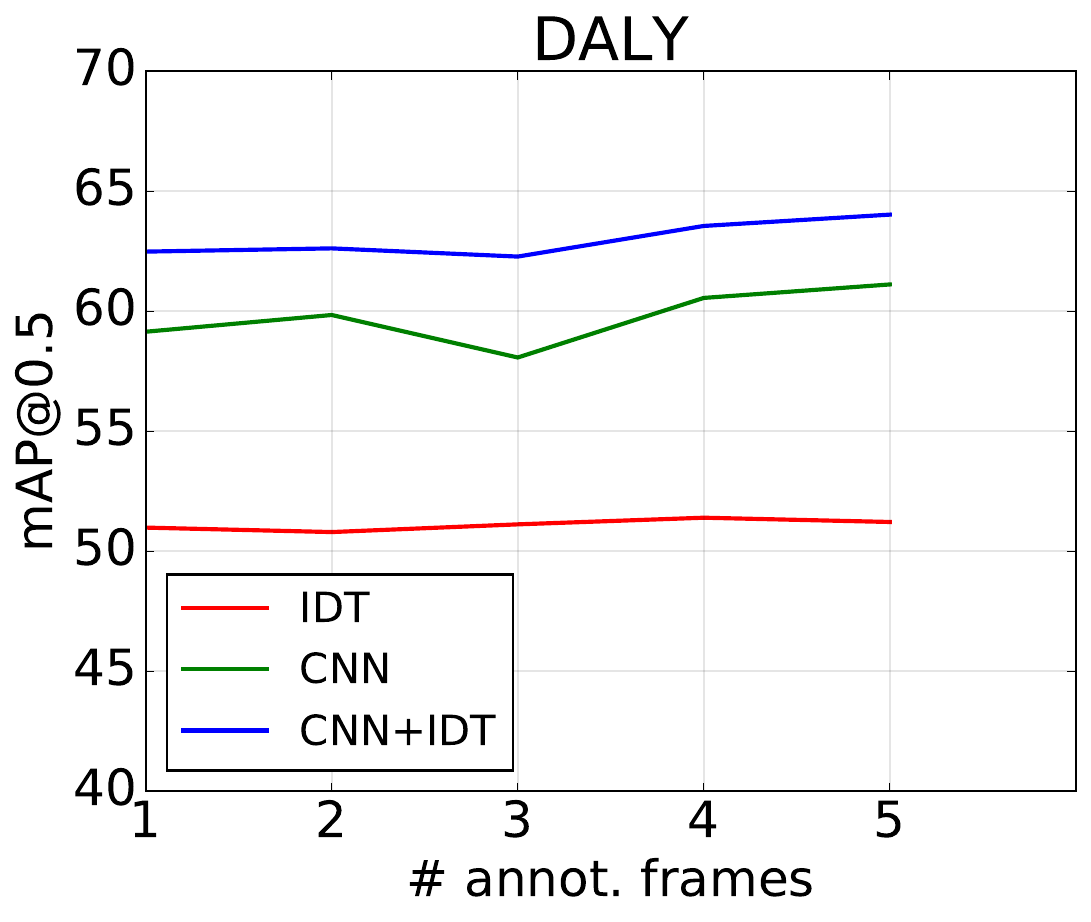}
 \caption{
Video-mAP@0.5 on trimmed clips when selecting positive tubes according to a varying number of frames. The cross indicates the performance when training on ground-truth tubes instead. Note that for the DALY dataset there is no ground-truth tube, and 5 frames means `up to 5', as short instances may have less annotated frames than 5. We report results for CNN and IDT separately, and their late fusion of scores (CNN+IDT).}
 \label{fig:nframes}
\end{figure*}

On the DALY dataset, the combination of the instance-level and human-level detectors in the tracker (MPII, I+H) obtains a Recall@0.5 of 91\% with an average of 5 tubes per clip on the test set when measuring the spatial IoU on the annotated frames. 
We observe that most of the failure cases correspond to small humans or occlusions.
Figure~\ref{fig:humantubes} shows the highest scoring human tube for several sequences of the DALY dataset.
In the first four examples, we can see that the human tube performs well despite motion of one arm (first row), turning of the person (second and third row),
camera motion (third row) or presence of an animal close to the human (fourth row).
Nevertheless, there are some failure cases due to the fact that the full body disappears (fifth row). 
In this case, only the feet remain visible causing the failure of the human tracker, as
the human detector performs poorly and the instance-level detector is
trained on previous frames where the full body is visible.
Another failure case is due to a partial occlusion by the camera
(sixth row), which causes the human detector to fail.

\section{Experimental results on action localization with sparse spatial supervision}
\label{sec:xp}

In this section, we present experimental results measuring the performance of action localization when using sparse spatial supervision.
In Section~\ref{sub:sparse} we show that using few annotated frames and our human tube-based approach, we achieve results similar to fully-supervised methods.
We compare this result to the impact of the number of annotated frames on Faster R-CNN used by state-of-the-art action localization methods in Section~\ref{sub:faster}.
Next, we compare our work with the state of the art on existing datasets in Section~\ref{sub:eval}.
We present experimental results on the DALY dataset in Section~\ref{sub:xpdaly}.

\subsection{Impact of sparse spatial supervision}
\label{sub:sparse}

To measure the impact of sparse spatial supervision, we evaluate the video-mAP@0.5 on the trimmed clips using IDT, CNN and CNN+IDT features.
Results are reported in Figure~\ref{fig:nframes} for the UCF-Sports, J-HMDB, UCF-101 and DALY datasets.
We compare to a baseline in which we train on the ground-truth tubes.
More precisely, we use as positives the ground-truth tubes, and as negatives the human tubes that do not match any ground-truth (IoU below 0.5).
The performance when training on the ground-truth is shown with the crosses.

We measure the impact of varying the number of annotated frames from 1 to 5
and compare to the performance with ground-truth annotations. 
Overall, the performance is not significantly impacted by the number of annotated frames, with variations limited to a maximum of 3\% for all datasets. 
More interestingly, the difference between training on the ground-truth tubes and training with sparse supervision is also limited to this range.
This is despite a significant decrease of annotation cost from 1 box per frame (i.e., 100 boxes for an instance of 100 frames) to 1 per action instance. 
In other words, for a fraction of the annotation cost we obtain similar results.

In more details, we can observe that the performance is almost constant in all cases for the UCF-Sports dataset, with a variation of less than 2\% for all features. The slight variations when using the CNN features can be explained by the randomness when learning a CNN, e.g. weight initialization for the last layers or the random shuffling of the frames.
On the other datasets, we can see a small increase in performance when moving from 1 annotated frame to 5, or when training with the ground-truth. 
However, this small increase is limited to around 2\% in all cases.

In summary, the drop in performance when using only 1
annotated frame per instance is very small. This holds true for both CNN and IDT features as well as their combination.
This demonstrates the effectiveness of leveraging human tubes in order to drastically reduce annotation effort when training an action detector.

So far, we consider that the sparse subset of the $N$ annotated frames are uniformly sampled in time. 
The following experiments examine the impact of this choice, using IDT features and randomly choosing $N$ annotated frames.
We run this experiment 50 times and report the mean and the standard deviation (std) in Table~\ref{tab:var} for different numbers of annotated frames. 
We obtain a std of around 0.5\% on UCF-101 and J-HMDB. It is a bit higher with around 1.5\% on UCF-Sports, which can be explained by the small number of videos resulting in higher variance. 
We can observe that the standard deviation decreases slightly as the number of annotated frames increases.
The low standard deviation shows that the choice of the annotated frames has negligible impact, which can be explained by the high quality of our human tubes.

\subsection{Impact of training data on Faster R-CNN}
\label{sub:faster}

In this section, we measure the gain due to the additional annotations
obtained by the human tubes, i.e., we train CNNs only on the sparse
annotations of the actions without any additional human data. We
resort to  Faster R-CNN, which has shown state-of-the-art results for
action detection at the frame level~\cite{MR2RCNN,Saha2016}.
We run a baseline two-stream Faster R-CNN on the same annotated frames
as in our sparse supervision scheme, ranging from 1 to 5 annotated frames per action instance, uniformly sampled.
To combine the two streams, we use the union of the Region-of-Interests from both streams~\cite{MR2RCNN} and fuse the scores of both streams using late fusion.
 We report the frame-mAP@0.5 in Figure~\ref{tab:faster} when varying
 the number of annotated frames. We also report the performance when
 training on all frames (GT).

\begin{table}
\centering
\begin{tabular}{|c|c|c|c|}
\hline
\#nframes    & UCF-Sports & J-HMDB & UCF-101 \\
\hline
     1 &  92.2 $\pm$ 1.7 &  58.1 $\pm$ 0.6 &  45.5 $\pm$ 0.6 \\
     2 &  92.6 $\pm$ 1.6 &  58.5 $\pm$ 0.5 &  46.3 $\pm$ 0.4 \\
     3 &  93.0 $\pm$ 1.3 &  58.5 $\pm$ 0.4 &  46.7 $\pm$ 0.4 \\
     4 &  93.1 $\pm$ 1.4 &  58.5 $\pm$ 0.4 &  46.8 $\pm$ 0.5 \\
     5 &  93.0 $\pm$ 1.4 &  58.6 $\pm$ 0.3 &  46.9 $\pm$ 0.4 \\
\hline
\end{tabular}
\caption{Impact of random selection of annotated frames, the number of annotations per video ranges from 1 to 5. We report 
mean and standard deviation for video-mAP@0.5 on trimmed tracks using IDT features.  Results on J-HMDB and UCF-101 are reported for the first split only.}
\label{tab:var}
\end{table}

We can observe a clear drop between training on all frames (GT) and on
one annotated frame per instance. The frame-mAP@0.5 decreases by 20\%
on the UCF-Sports, 8\% on J-HMDB and 10\% on UCF-101. 
The drop is lowest for J-HMDB, which can be explained by extremely
short videos and thus reduced diversity in appearance (fewer unrelated
images). The significant decrease of 20\% on UCF-Sports is due
to (a) a low number of videos in the training set and (b) longer
instances with significant variation for some classes such as
\textit{Diving} or \textit{Swinging}. 
When training on 5 annotated frames per video, there still exists a gap compared to
training, in particular for small datasets (6\% on UCF-Sports and 3\%
on J-HMDB). For larger datasets such as UCF-101, the frame-mAP@0.5 is
on par with training on full ground-truth.

\subsection{Comparison to the state of the art}
\label{sub:eval}

We now compare our CNN+IDT approach to the state of the art on UCF-Sports, J-HMDB and UCF-101, see Table~\ref{tab:ap}.
We report our results with two settings: when training on sparse spatial annotations (first two rows) with 1 or 5 annotated frames per instance, and when training on the ground-truth tubes.

We first compare our results to Mettes et al.~\cite{mettes2016spot}, a recent method which uses sparse supervision based on one point per frame instead of ground-truth tubes. 
We can observe that our approach outperforms theirs substantially by
25\% despite the fact that their approach uses
significantly more annotations: 1 point per frame compared to 1 box (i.e., 2 points) per instance. 
On the training set of the first split of UCF-101 this represents 1.9M points with their point annotation compared to 8k points in our approach.
To examine the cause of the gain we 
report results with the same features and the same annotations.
We use their point annotations, which are the center point of the 
bounding boxes, to select positive human tubes. 
We use as positives the human tubes for which at least 80\% of the points are inside the tracks, and as negative the human tubes that have IoU below $0.5$ with the positives. 
For a fair comparison, the results of our approach are reported using IDT features only. We report our results in Table~\ref{tab:ap}.
We obtain a similar performance with the point annotations scheme, 57.5\% video-mAP@0.2 on UCF-101, than with our sparse supervision scheme: 57.1\% when using IDT alone. These performances are still significantly higher than~\cite{mettes2016spot} with an improvement of 24\% in video-mAP@0.2 on UCF-101.
The remaining source of difference lies in the human tubes.
To examine this, we compare the APT proposals~\cite{van2015apt} used by Mettes et al.~\cite{mettes2016spot} to our human tubes in Table~\ref{tab:prop} and report Recall@0.5.
We can observe that the quality of our human tube is significantly higher, i.e., our human tubes obtain a significantly higher recall with only a few tubes.
In contrast, APT outputs thousands of proposals and reaches a lower recall.
This clearly explains the gap in performance. 

\begin{table}
\centering
\begin{tabular}{|c|c|c|c|}
\hline
\#nframes   & UCF-Sports & J-HMDB & UCF-101 \\
\hline
     1 &  67.9 &  50.9 &  53.6 \\
     2 &  77.4 &  50.7 &  59.1 \\
     3 &  79.6 &  55.8 &  61.2 \\
     4 &  81.0 &  56.2 &  62.5 \\
     5 &  81.2 &  56.5 &  63.8 \\
\hline
    GT &  87.6 &  59.1 &  63.1 \\
\hline
\end{tabular}
\caption{
Frame-mAP@0.5 of two-stream Faster R-CNN when training on the annotated frames only. GT refers to training on all frames. Results on J-HMDB and UCF-101 are reported for the first split only.
}
\label{tab:faster}
\end{table}

\begin{table*}
 \centering
 \resizebox{\textwidth}{!}{
 \begin{tabular}{|l|c|c||c||c||c|c||c|}
  \hline
  \multirow{2}{*}{Method} & \multirow{2}{*}{Annot.} & \multirow{2}{*}{features} & UCF-Sports & J-HMDB & \multicolumn{2}{c||}{UCF-101 (split 1)} & DALY \\ 
  \cline{4-8}
  &  &  & mAP@0.5 & mAP@0.5 & mAP@0.05 & mAP@0.2 & mAP@0.2 \\
  \hline
%
\hline  
  \multirow{2}{*}{\textbf{ours (sparse)}} & 1 frame & \multirow{2}{*}{CNN (Fast R-CNN) + IDT} & \textbf{95.9\%} & \textbf{63.9\%} & \textbf{70.0\%} & \textbf{57.4\%} & \textbf{14.5\%} \\
  \cline{2-2} \cline{4-8}
  & 5 frames &  & \textbf{97.1\%} & \textbf{64.0\%} & \textbf{67.1\%} & \textbf{57.3\%} & \textbf{13.9\%} \\
  \hline  
%
  \hline
  Mettes et al.~\cite{mettes2016spot}  & \multirow{2}{*}{points} & \multirow{3}{*}{IDT} & - & - & - & 32.4\% & - \\
  \cline{1-1} \cline{4-8}
  \textbf{ours}  &  &  & \textbf{94.3\%} & 59.7\% & - & \textbf{57.5\%} & - \\
  \cline{1-2} \cline{4-8}
  \textbf{ours (sparse)} & 1 frame &  & 93.9 & \textbf{59.8} & - & 57.1\% & 14.2\% \\
  \hline
%
  \hline
  \textbf{ours} & \multirow{7}{*}{GT} & CNN (Fast R-CNN) + IDT & \textbf{95.9\%} & 65.8\% & 71.1\% & 58.9\% & - \\
  \cline{1-1} \cline{3-8}
  Gkioxari and Malik~\cite{fat} &  & CNN (R-CNN) & 75.8\% & 53.3\% & - & - & - \\
  \cline{1-1} \cline{3-8}
  Weinzaepfel~et al.~\cite{weinzaepfelICCV15} &  & CNN (R-CNN) + Handcrafted & 90.5\% & 60.7\% & 54.3\% & 46.8\% & - \\
  \cline{1-1} \cline{3-8}
  van Gemert~et al.~\cite{van2015apt} &  & IDT & - & - & 58.0\% & 37.8\% & - \\
  \cline{1-1} \cline{3-8}
  Yu and Yuan~\cite{Yu_2015_CVPR} &  & IDT & - & - & 42.8\% & - & - \\
  \cline{1-1} \cline{3-8}
  Saha~et al.~\cite{Saha2016} &  & CNN (Faster R-CNN) & - & 71.5\% & \textbf{79.1\%} & 66.7\% & - \\
  \cline{1-1} \cline{3-8}
  Peng and Schmid~\cite{MR2RCNN} &  & CNN (Faster  R-CNN) & 94.8\% & \textbf{73.1\%} & 78.8\% & \textbf{72.9\%} & - \\
  \hline
 \end{tabular}
 }
 \caption{Comparison to the state of the art with video-mAP@0.5 on spatial localization datasets (UCF-Sports and J-HMDB) and video-mAP@0.2 for spatio-temporal action localization benchmarks (UCF-101 and DALY).
 For UCF-101 we also report video-mAP@0.05 to compare to~\cite{Yu_2015_CVPR} which also leverages a human detector.
We first present the results of our method with sparse spatial annotation (first two rows).
 For comparison to Mettes et al.~\cite{mettes2016spot}, we also report our performance using IDT only and their point annotation scheme that assumes that the center point of each box is given.
 We finally  report our results when training on ground-truth as well as state-of-the-art fully-supervised approaches (GT).
}
 \label{tab:ap}
\end{table*}

\begin{table}
\centering
\resizebox{\linewidth}{!}{
\begin{tabular}{|c||r@{ }l|r@{ }l|r@{ }l|}
\hline
 & \multicolumn{2}{c|}{UCF-Sports} & \multicolumn{2}{c|}{J-HMDB} & \multicolumn{2}{c|}{UCF-101} \\ 
\hline 
\textbf{Human Tubes} & \textbf{97.4\%} & (2.7) & \textbf{95.3\%} & (1.3) & \textbf{65.0\%} & (3.0) \\
APT~[\cite{van2015apt} & 89.4\% & (1449) & \multicolumn{2}{c|}{-} & 36.8 & (2299) \\
\hline
\end{tabular}
}
\caption{Comparison of our human tubes and APT~\cite{van2015apt} (used by Mettes et al.~\cite{mettes2016spot}) with Recall@0.5 over all videos for the UCF-Sports, J-HMDB and UCF-101 datasets. 
The average number of proposals is in parenthesis.}
\label{tab:prop}
\end{table}

We also compare our approach to state-of-the-art fully supervised approaches in Table~\ref{tab:ap}. 
On the UCF-Sports dataset, we obtain state-of-the-art video-mAP with 95.9\% when training on the ground-truth, and the same performance when using only one annotated frame per action instance.
This performance is explained by the high quality of the human tubes (Section~\ref{sub:humaneval}) and the fact that we combine IDT and CNN features.

For the J-HMDB dataset we obtain a video-mAP@0.5 of 65.8\% with full
supervision, which is slightly below the state of the art.  The drop
in performance compared to for example~\cite{MR2RCNN} can   
be explained by their additional multi-region and multi-flow
description. 
Furthermore, using human tubes required in a weakly supervised case 
are likely to be the cause for an additional drop, as the regions are not
selected and adapted to the action class.
When considering that only one frame (resp. 5 frames) is spatially annotated for each
action instance, we obtain a video-mAP of 63.9\% (resp. 64.0\%), which is less than 2\% below the fully-supervised variant.

For the UCF-101 dataset, we obtain a video-mAP@0.2 of 58.9\% with full
supervision and of 57.4\% with one box annotated per instance, i.e.,
there is almost no drop in performance due to using sparse supervision.
This performance is on-par with most existing approaches but below~\cite{Saha2016}, which can be explained by the relatively low recall of our tubes.
At a threshold of 0.05 we perform better than most state-of-the-art methods. 
In particular, we significantly outperform~\cite{Yu_2015_CVPR} which
also leverages a human detector.

\subsection{Evaluation on DALY}
\label{sub:xpdaly}

We now present experimental results on DALY.
We separately evaluate 
spatial detection in trimmed clips and spatio-temporal detection in full videos.

\subsubsection*{Action localization in trimmed clips}
We first evaluate action localization in trimmed clips.
To this end, we only test on the human tubes trimmed to the ground-truth temporal extent of the actions, i.e., we measure spatial detection performance, as in the case of the UCF Sports and J-HMDB datasets. 
Figure~\ref{fig:nframes} reports the results when varying the maximum number of annotated frames from 1 to 5 using IDT, CNN and CNN+IDT features.
We obtain a video-mAP@0.5 of 64.0\% with CNN+IDT when using all annotated frames, i.e., up to 5 per instance, compared to 62.5\% when using only one frame annotation.
Once again, this gap is extremely small and validates the effectiveness of our training from sparse supervision. On the DALY dataset, we observe that IDT performs significantly worse than CNN. This can be explained by the fact that many instances are short in time, and consequently contain a relatively small number of trajectories when building the Fisher Vector representation with IDT.

\subsubsection*{Spatio-temporal action localization} 

We finally evaluate our method on action localization in space and time.
We report a video-mAP@0.2 of 13.9\% when training on all annotated frames, i.e., 5 per instance, see Table~\ref{tab:ap}.
We obtain a similar performance of 14.5\% when considering only 1 annotated frame per action instance.
The drop compared to spatial localization can be explained 
by the difficulty of temporal detection as the dataset contains both short and long actions in long untrimmed videos.

\begin{figure*}
 \includegraphics[width=0.245\textwidth]{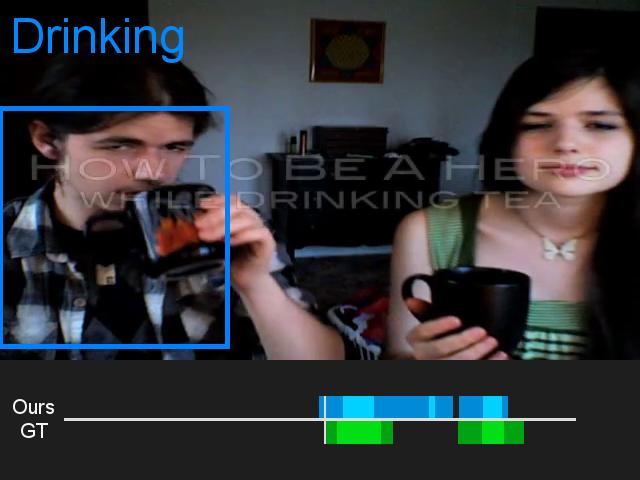} \hfill
 \includegraphics[width=0.245\textwidth]{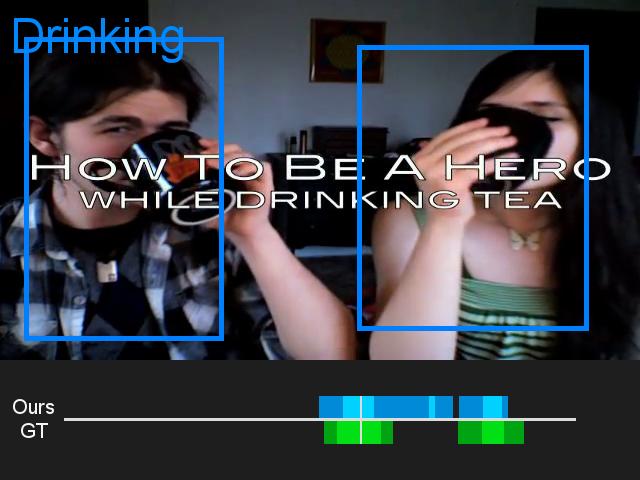} \hfill
 \includegraphics[width=0.245\textwidth]{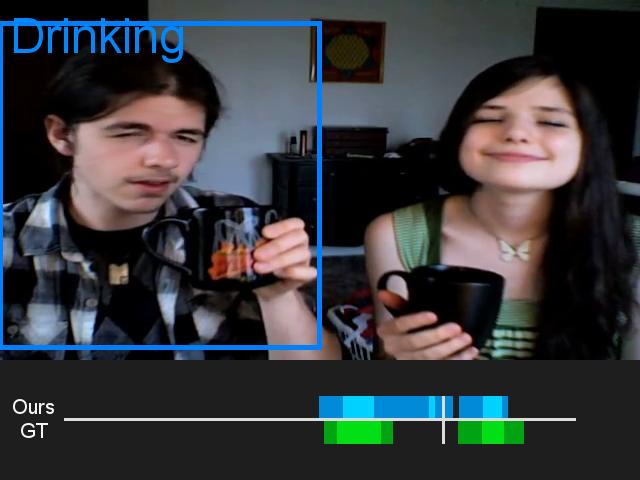} \hfill
 \includegraphics[width=0.245\textwidth]{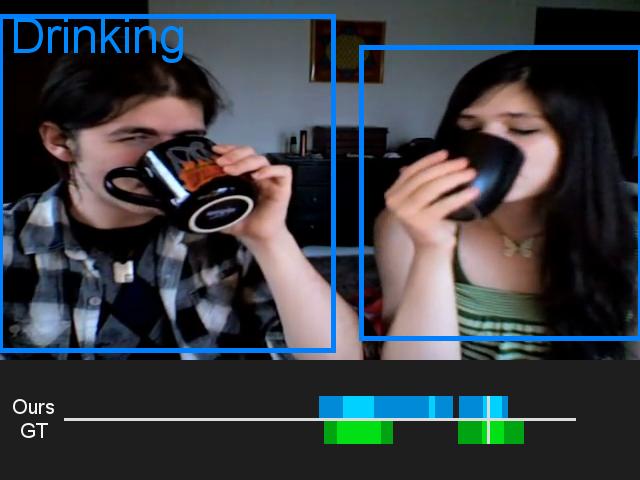} \\
 
 \includegraphics[width=0.245\textwidth]{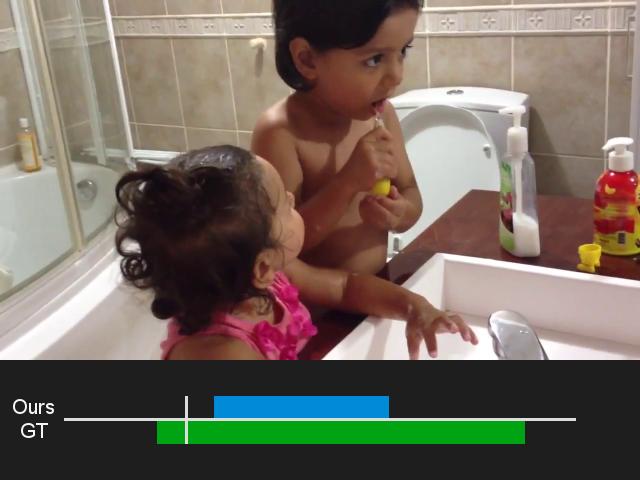} \hfill
 \includegraphics[width=0.245\textwidth]{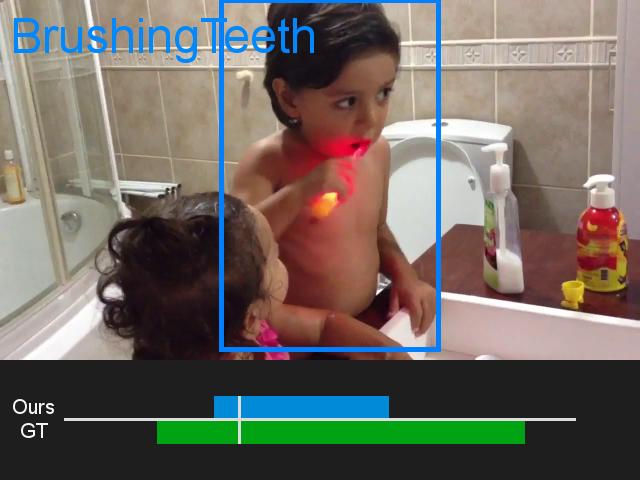} \hfill
 \includegraphics[width=0.245\textwidth]{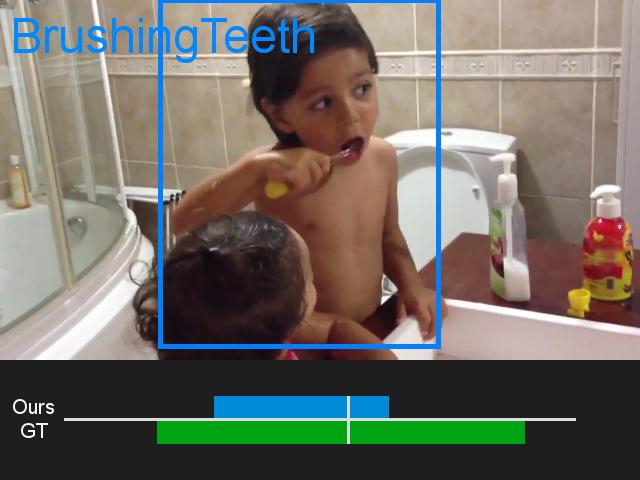} \hfill
 \includegraphics[width=0.245\textwidth]{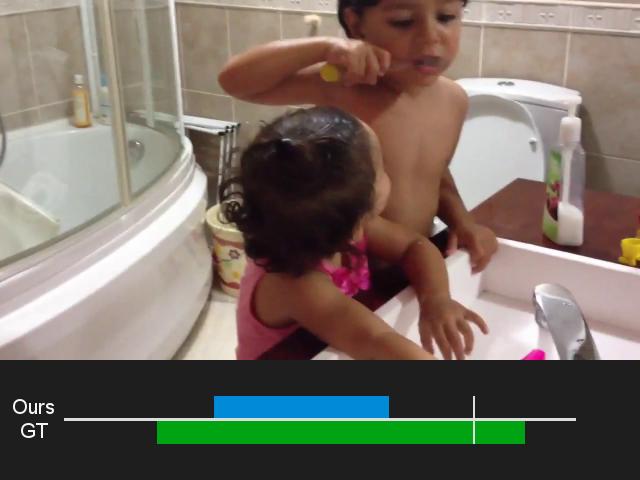} \\
 
 \includegraphics[width=0.245\textwidth]{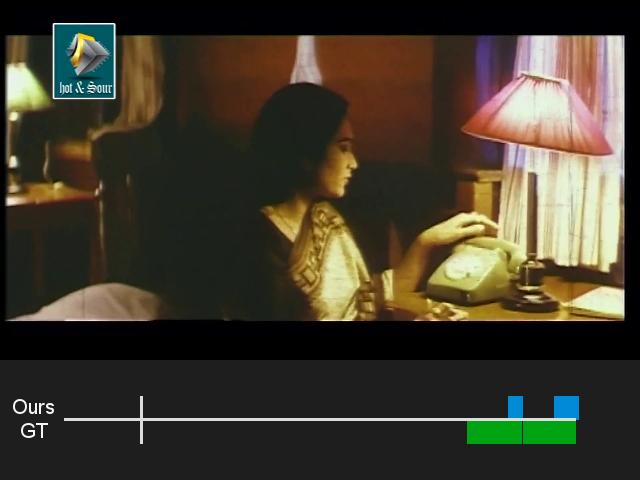} \hfill
 \includegraphics[width=0.245\textwidth]{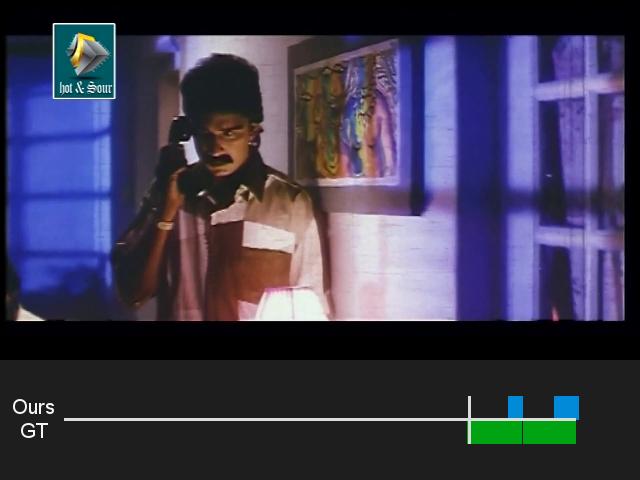} \hfill
 \includegraphics[width=0.245\textwidth]{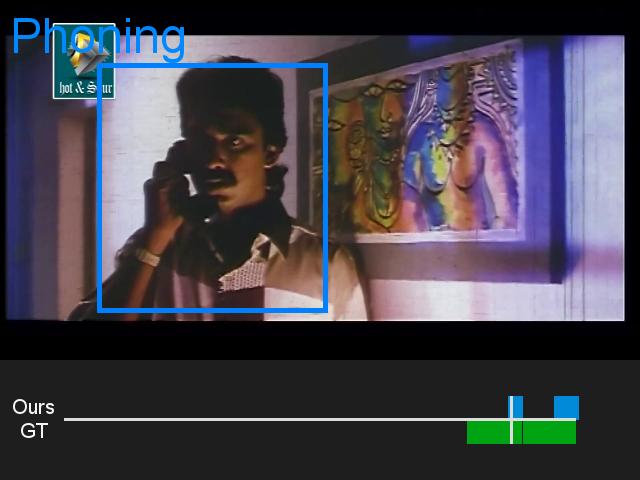} \hfill
 \includegraphics[width=0.245\textwidth]{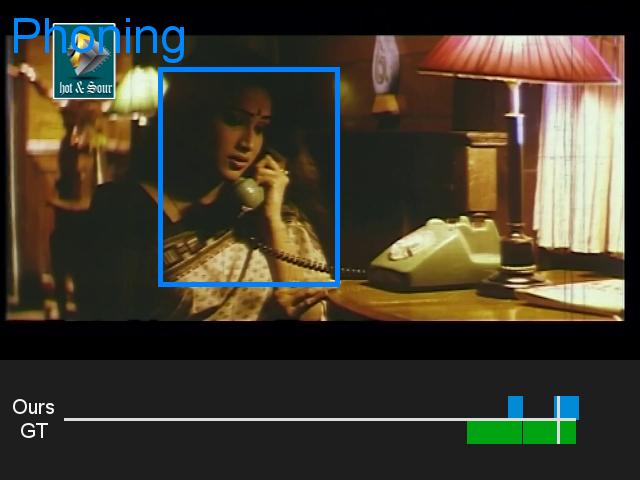} \\
 
 \includegraphics[width=0.245\textwidth]{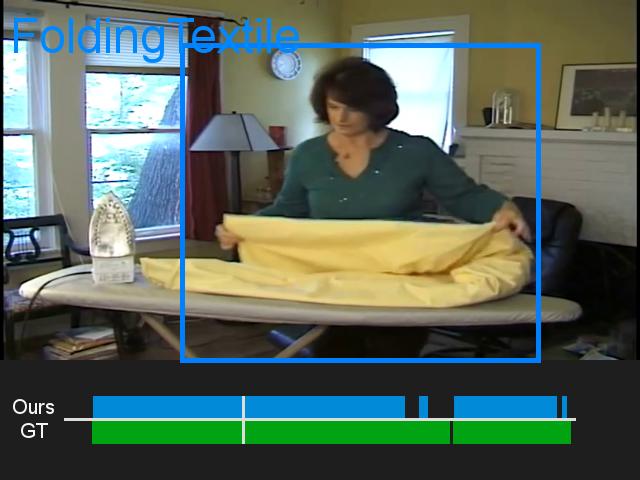} \hfill
 \includegraphics[width=0.245\textwidth]{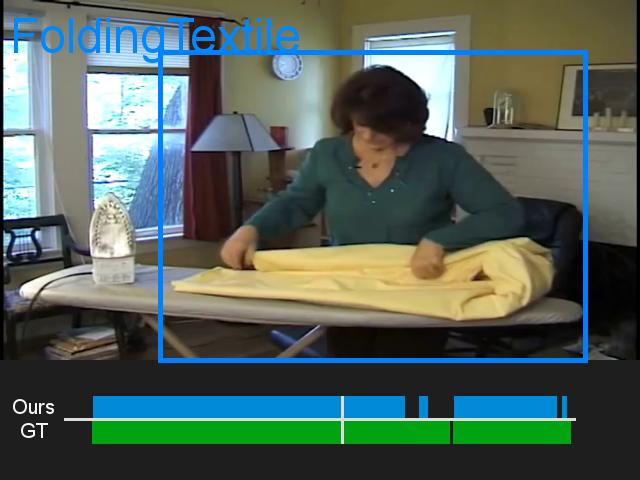} \hfill
 \includegraphics[width=0.245\textwidth]{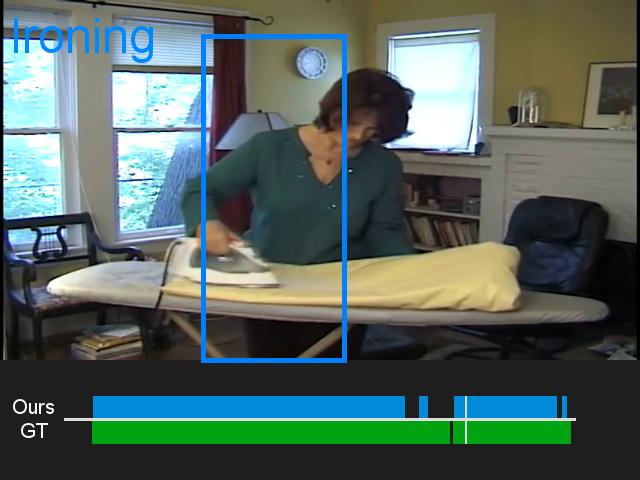} \hfill
 \includegraphics[width=0.245\textwidth]{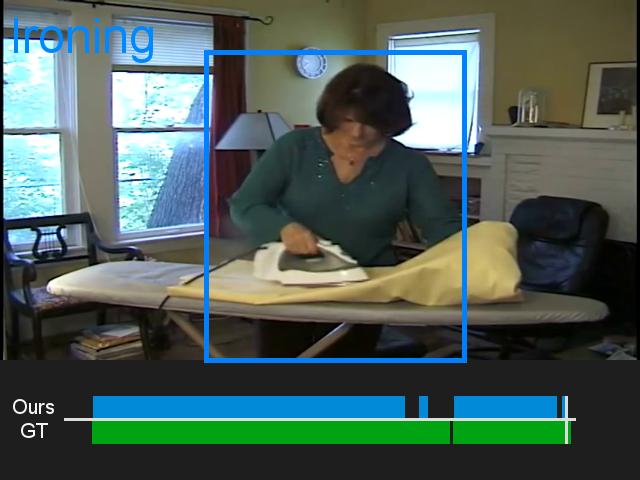} \\
 
 \includegraphics[width=0.245\textwidth]{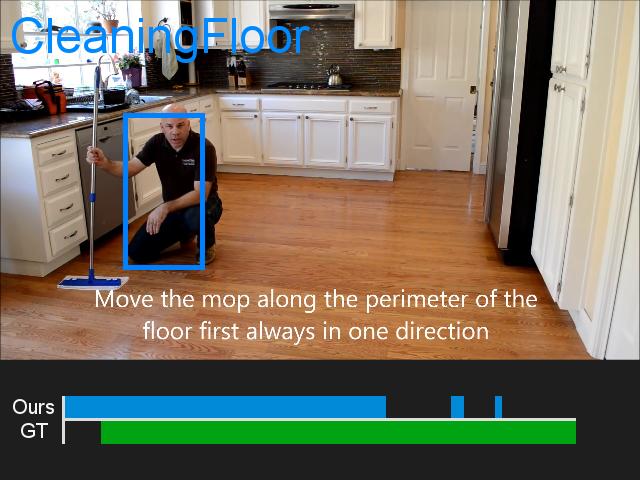} \hfill
 \includegraphics[width=0.245\textwidth]{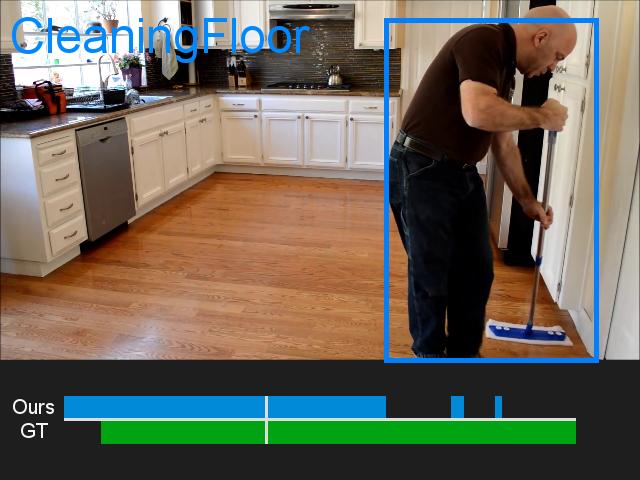} \hfill
 \includegraphics[width=0.245\textwidth]{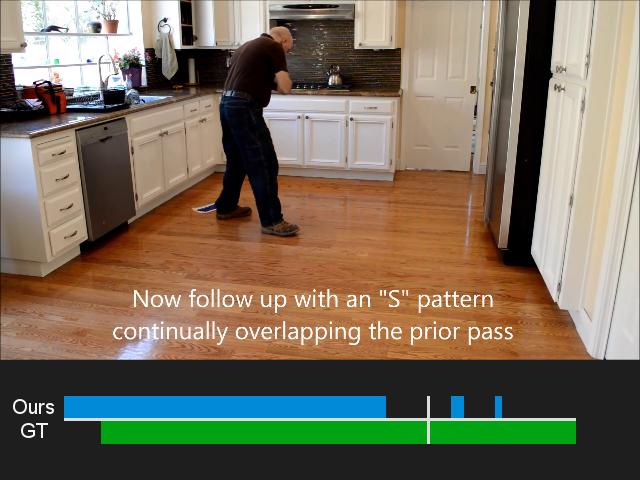} \hfill
 \includegraphics[width=0.245\textwidth]{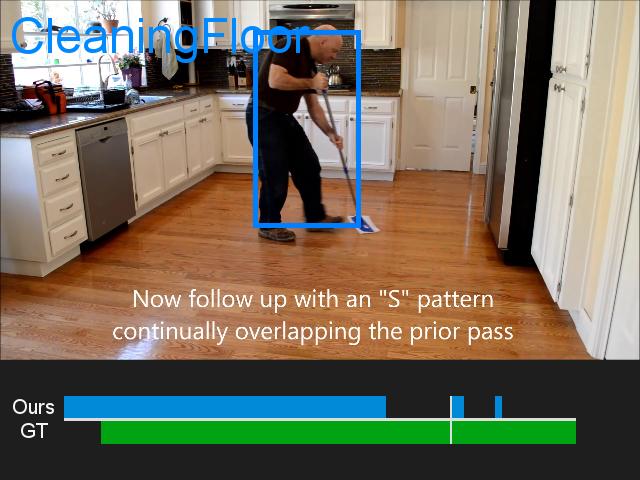} \hfill
 
 \includegraphics[width=0.245\textwidth]{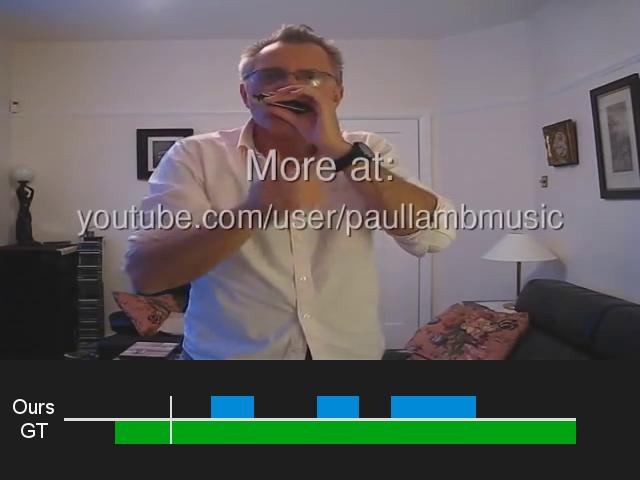} \hfill
 \includegraphics[width=0.245\textwidth]{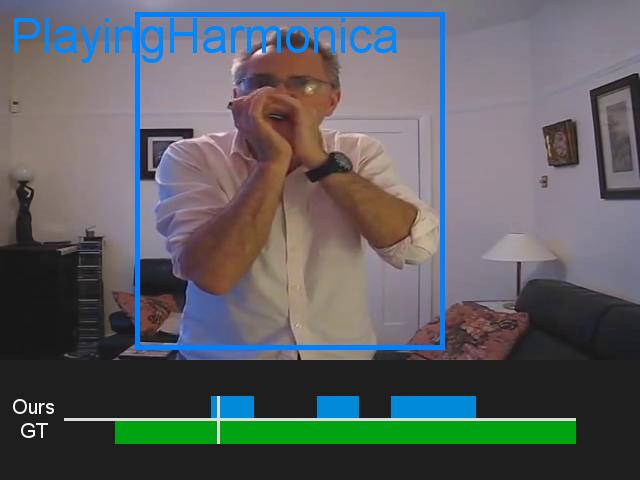} \hfill
 \includegraphics[width=0.245\textwidth]{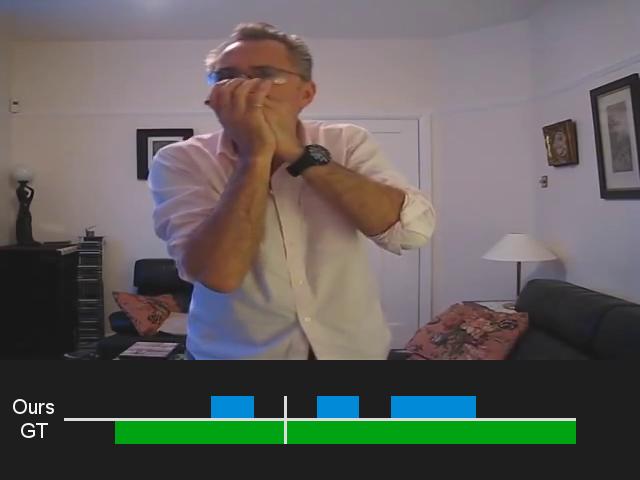} \hfill
 \includegraphics[width=0.245\textwidth]{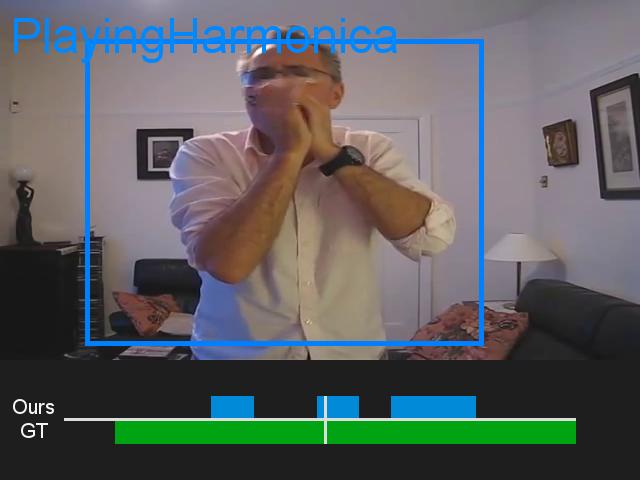} \\
 
 \caption{Spatio-temporal detection examples on the DALY dataset.
 The timeline shows the ground-truth and our detection for a clip. 
 The video cursor indicates the displayed frame, for which we show the action detection.}
 \label{fig:examples}
\end{figure*}

Figure~\ref{fig:examples} shows a few detection examples. 
In the first row, we are able to distinguish multiple instances of drinking performed by two different actors, with
only a small false positive detection in the third column. 
In the second row we can observe that our detection is a bit too short, but well localized. 
Similar findings are also valid for the third row with two instances of the phoning actions.
In the fourth row we detect both actions, folding textile and ironing, with an accurate time overlap. 
However, there is one
short-lived
false detection at the end of each detected instance.
In the fifth row, our human tube is able to track and detect the cleaning floor action despite significant motion in the scene, but the temporal detection is cut into parts towards the end of the action. 
The last row shows a failure case in which a long action, playing harmonica, is detected as many small chunks. We can observe that this dataset requires a more sophisticated approach  for localization in time. 

\section{Conclusion}

We have presented an effective approach for extracting human tubes using a generic human detector.
We have shown that this allows to significantly reduce the level of spatial supervision for training an action detector.
In particular, when considering only one annotated frame per instance, the drop in performance is
almost insignificant
while the annotation cost is drastically reduced.
We also introduced DALY, the first dataset for action localization in
space and time in real-world untrimmed videos.
It overcomes the limitations of existing datasets and will allow
to measure progress in the field over the next years.

\ifCLASSOPTIONcompsoc
  \section*{Acknowledgments}
\else
  \section*{Acknowledgment}
\fi

  This work was supported in part by
the ERC advanced grant ALLEGRO, the MSR-Inria joint
project, a Google research award and a Facebook gift. We
gratefully acknowledge the support of NVIDIA with the donation
of GPUs used for this research.

\ifCLASSOPTIONcaptionsoff
  \newpage
\fi



%
%
%

\bibliographystyle{IEEEtran}
\bibliography{biblio}

%

\begin{IEEEbiographynophoto}
{Philippe Weinzaepfel}
received a Master degree from Universit\'e Grenoble Alpes, France, and Ecole Normale Sup\'erieure de Cachan, France, in 2012.
He was a doctoral student in the THOTH
team, at INRIA Grenoble and Laboratoire Jean Kuntzmann, from 2012 until 2016,
and received a PhD degree in computer science
from Universit\'e de Grenoble, France, in 2016. ´
He is currently a research scientist at Xerox Research Centre Europe, France, in the computer vision group. His research interests include computer vision and machine learning, with special interest in video understanding and action recognition.
\end{IEEEbiographynophoto}

\begin{IEEEbiographynophoto}{Xavier Martin}
is currently a research engineer in the THOTH team (INRIA Grenoble) since 2014.
He received a Master's degree of Engineering from \'Ecole Nationale d'Informatique
et de Math\'ematiques Appliqu\'ees de Grenoble (ENSIMAG) in 2014.
He participated in the European project AXES, the ERC grant Allegro
and was an apprentice engineer in research team Inria  MOAIS while
preparing his degree (2011-2014).   
\end{IEEEbiographynophoto}


\begin{IEEEbiographynophoto}{Cordelia Schmid}
Cordelia Schmid holds a M.S. degree in Computer Science from the
University of Karlsruhe and a Doctorate, also in Computer Science,
from the Institut National Polytechnique de Grenoble (INPG). Her
doctoral thesis received the best thesis award from INPG in 1996. 
Dr. Schmid was a post-doctoral research assistant in the Robotics
Research Group of Oxford University in 1996--1997. Since 1997 she has
held a permanent research position at INRIA Grenoble Rhone-Alpes,
where she is a research director and directs an INRIA team. Dr. Schmid
has been an Associate Editor for IEEE PAMI (2001--2005) and for IJCV 
(2004--2012), editor-in-chief for IJCV (2013---), a program chair of
IEEE CVPR 2005 and ECCV 2012 as well as a general chair of IEEE CVPR
2015 and ECCV 2020. In 2006, 2014 and 2016, she was awarded 
the Longuet-Higgins prize for fundamental contributions in computer
vision that have withstood the test of time. She is a fellow of
IEEE. She was awarded an ERC advanced grant in 2013, the Humbolt
research award in 2015 and the Inria \& French Academy of Science
Grand Prix in 2016. She was elected to the German National
Academy of Sciences, Leopoldina, in 2017. 

\end{IEEEbiographynophoto}




\end{document}


\title{Human Action Localization with Sparse Spatial Supervision \\ Supplementary Material}

\author{First Author\\
Institution1\\
Institution1 address\\
{\tt\small firstauthor@i1.org}
\and
Second Author\\
Institution2\\
First line of institution2 address\\
{\tt\small secondauthor@i2.org}
}

\maketitle

We first give details on how the DALY dataset was assembled and show some statistics for each class (Section~\ref{sup:daly}).
We then present some examples of human tubes extracted on this dataset (Section~\ref{sup:tubes}).
Finally, we comment on the attached video that shows some results of our spatio-temporal action detector learned from sparse supervision (Section~\ref{sup:examples}).

\section{The DALY Dataset}
\label{sup:daly}

This section describes how the DALY dataset was assembled. 
We first explain the action class selection,
the filtering of the videos and the spatio-temporal annotation of action instances.
Next, per-class statistics are presented for the dataset.

\subsection{Developing an action localization dataset}

\paragraph{Picking action classes.}
In order to obtain a dataset that fairly evaluates action localization methods, action classes must have clearly defined temporal boundaries.
Ambiguities introduce noise in label and temporal annotation, which makes the evaluation unreliable.

We thus restrict ourselves to classes for which temporal boundary guidelines can be stated precisely and concisely.
For instance, the \textit{brushing teeth} action is defined as \textit{``toothbrush inside the mouth''}.
Another example is \textit{cleaning windows} for which the moment where \textit{``the tool is in contact with the window''} is annotated.
An example for an ambiguous action is \textit{washing hands}, for which temporal boundaries are hard to establish.
Such actions are avoided.

In order to make class distinction challenging,
some of the classes are chosen to contain similarities, for instance in terms of motion patterns.
Several of our action classes imply motion of the hands near the head (\textit{taking photos, phoning})
or the mouth (\textit{playing harmonica, drinking, brushing teeth, applying make up on lips}).

In summary, we collect data for the following 10 actions: \textit{applying make up on lips, brushing teeth, cleaning floor, cleaning windows, drinking, folding textile, ironing, phoning, playing harmonica and taking photos/videos}.

\paragraph{Video collection.} 
The videos are gathered from YouTube using manually picked queries related to the desired action labels.
For the class \textit{cleaning floor}, the queries include \textit{``sweeping floor''}, \textit{``mopping floor''}, \textit{``cleaning floor''}, etc.
We only collect videos that last between 1 and 20 minutes.
A minimum duration of 1 minute ensures that temporal localization will be meaningful (shorter videos contain only one action from the beginning to the end in most cases),
and a maximum duration of 20 minutes avoids disproportionate computational time. 

Videos are filtered to remove cartoons, slideshows, actions performed by animals and first-person viewpoints.
We also remove videos in which the human is not visible when the action occurs,
for instance when the camera focuses on the mop while performing the \textit{cleaning floor} action.

We keep 51 videos for each action class, where each video contains at least one instance of the action class.
In total, this corresponds to 31 hours of video or 3.3 million frames.
Videos collected for an action class often contain multiple instances of this class, and sometimes also other action classes.
In some cases, several instances are happening simultaneously.
We annotate all occurrences of the 10 classes exhaustively. 

\paragraph{Temporal annotation.} 
All videos are carefully annotated by members of our research team
with the \textit{begin} and \textit{end} time for all instances.
Precise guidelines are established before annotation. 
For example, the \textit{phoning} \mbox{action} lasts
as long as the phone remains close to the actor's ears.
In case of a shot change during an action, we annotate it as two separate instances and set a \textit{``shotcut''} flag on the second instance.
DALY contains 3724 action instances in total, with an average duration of 8 seconds.

For each action instance, we add a set of ``flags'' that state if an action is:
(a) small compared to the image, (b) very big compared to the image (zoomed in), (c) largely occluded at some point, (d) outside the camera's field of view at some point.

Not included in the above count are around 200 instances, which are annotated as ambiguous (unclear whether the action is genuinely performed) or mirror reflections. 
These annotations are ignored during evaluation, \ie, they neither count as missing positives nor as false positives.

\paragraph{Spatial annotation.}
An action is present 
in 700k frames out of 3.3M frames total.
Annotating all of them would be very time consuming and  clearly does not scale up
to large collections. Furthermore, as demonstrated by the approach
developed in this paper, sparse annotations are sufficient.   
Thus, we subsample the frames for spatial annotation.
For each temporal instance, we pick 5 frames uniformly sampled over
time, with a maximum of 1 frame per second. 
For each frame, annotators are asked to draw a bounding box around the
actor, a bounding box around the object(s) involved in the action (\eg
the glass/cup for \textit{drinking}), and the pose of the upper body
of the actor (bounding box around the head and keypoints for
shoulders, elbows and wrists). 
Some of the spatial annotations are completed by external workers, but
all of them are reviewed and adjusted when necessary.

\subsection{Dataset statistics}

\begin{figure}
 \centering
 \includegraphics[width=\linewidth]{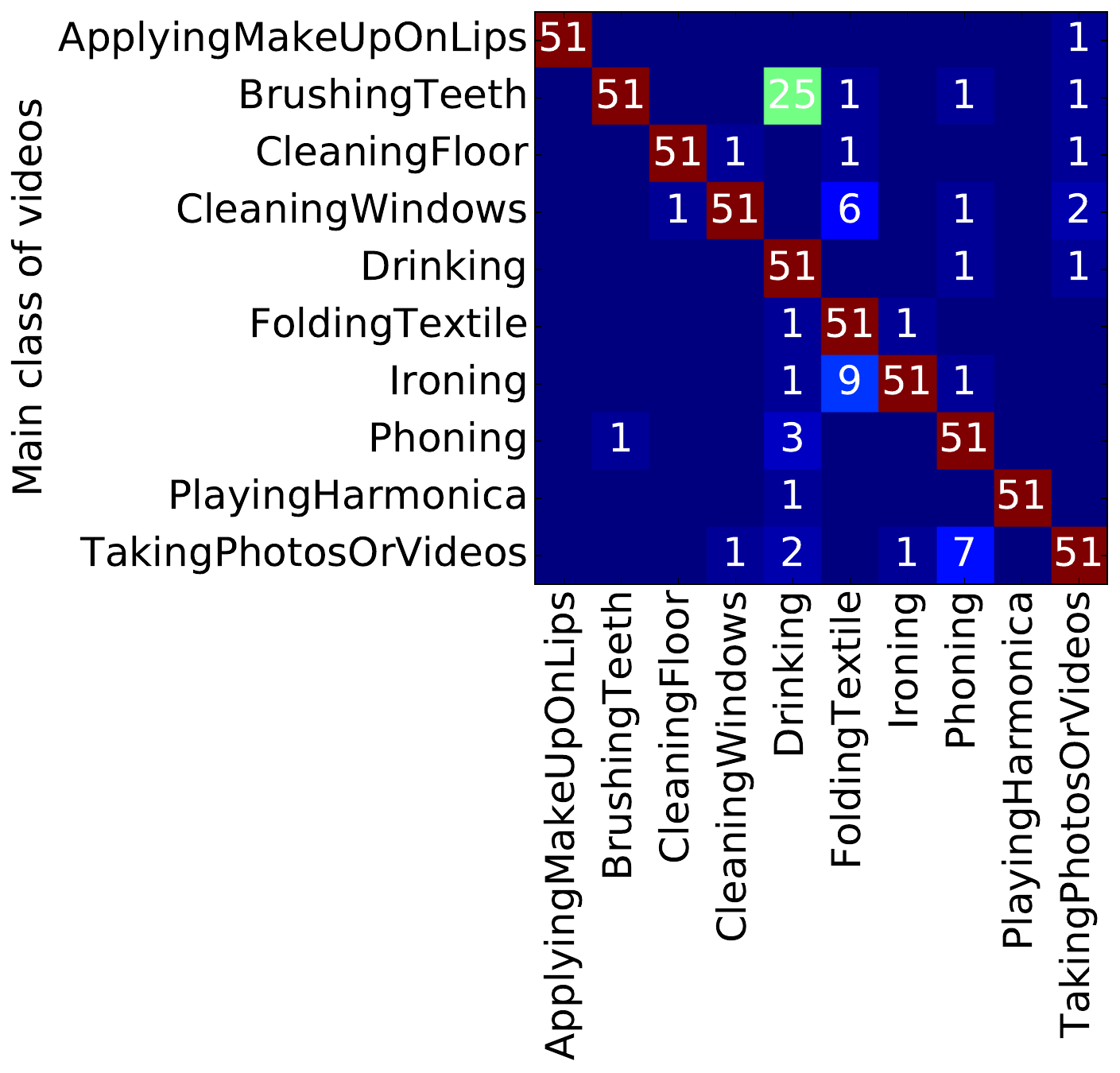}
\caption{Statistics of multiple classes per video. 
 Each row considers the 51 videos downloaded for a given class, each column counts the videos containing at least one instance of the column class.}
 \label{fig:labelmixing}
\end{figure}
 
 \begin{table}
  \centering
  \resizebox{\linewidth}{!}{
    \begin{tabular}{|c|c|c|c|}
     \hline
     class & avg video dur. & \#inst. & avg inst. dur.  \\
     \hline
ApplyingMakeUpOnLips & 376.8s $\pm$ 265.1 & 421s & 3.7 $\pm$ 3.3 \\
BrushingTeeth & 176.0s $\pm$ 120.3 & 277 & 9.3s $\pm$ 15.9 \\
CleaningFloor & 194.2s $\pm$ 128.7 & 200 & 14.1s $\pm$ 15.2 \\
CleaningWindows & 196.2s $\pm$ 131.9 & 468 & 7.1s $\pm$ 10.0 \\
Drinking & 202.4s $\pm$ 130.9 & 304 & 2.6s $\pm$ 3.0 \\
FoldingTextile & 184.1s $\pm$ 150.1 & 257 & 14.6s $\pm$ 22.4 \\
Ironing & 233.2s $\pm$ 183.8 & 424 & 7.2s $\pm$ 8.2 \\
Phoning & 217.9s $\pm$ 140.7 & 514 & 9.9s $\pm$ 30.2 \\
PlayingHarmonica & 190.2s $\pm$ 139.8 & 306 & 13.8s $\pm$ 21.4 \\
TakingPhotosOrVideos & 283.0s $\pm$ 207.3 & 553 & 3.1s $\pm$ 3.5 \\
\hline
all & 225.4s $\pm$ 175.7 & 3724 & 7.8s $\pm$ 16.4  \\
     \hline
    \end{tabular}
  }

 \caption{Statistics for each class showing the video duration (average and standard deviation), the number of instances, and the instance duration (average and standard deviation).}
  \label{tab:statclass}
  \end{table}

The selected action classes are sufficiently common such that multiple action classes can be found in a single video, see Figure~\ref{fig:labelmixing}.
Each row of the matrix displays the 
presence of other action classes in the 51 videos of a given class.
For example, out of the 51 videos selected for the class
\textit{brushing teeth}, 25 videos also contain \textit{drinking} instances.
There is overlap between \textit{ironing} and \textit{folding textile}, and 
between \textit{taking photos/videos} and \textit{phoning}, 
which can be explained by the fact that taking photos is mostly performed outdoors, where other people are \textit{phoning} or \textit{drinking}. 

\begin{figure}
 \includegraphics[width=0.49\linewidth]{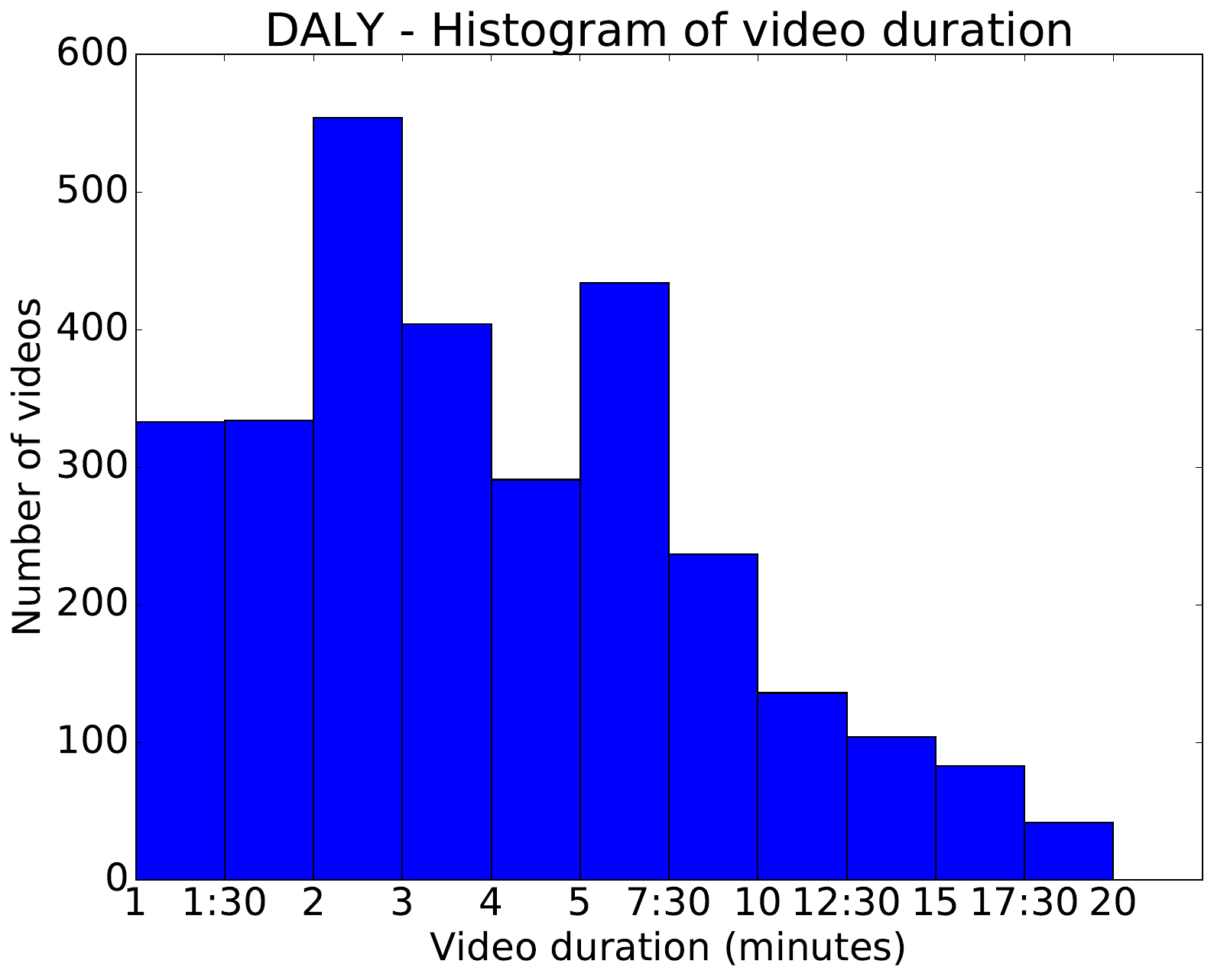}
 \hfill
 \includegraphics[width=0.49\linewidth]{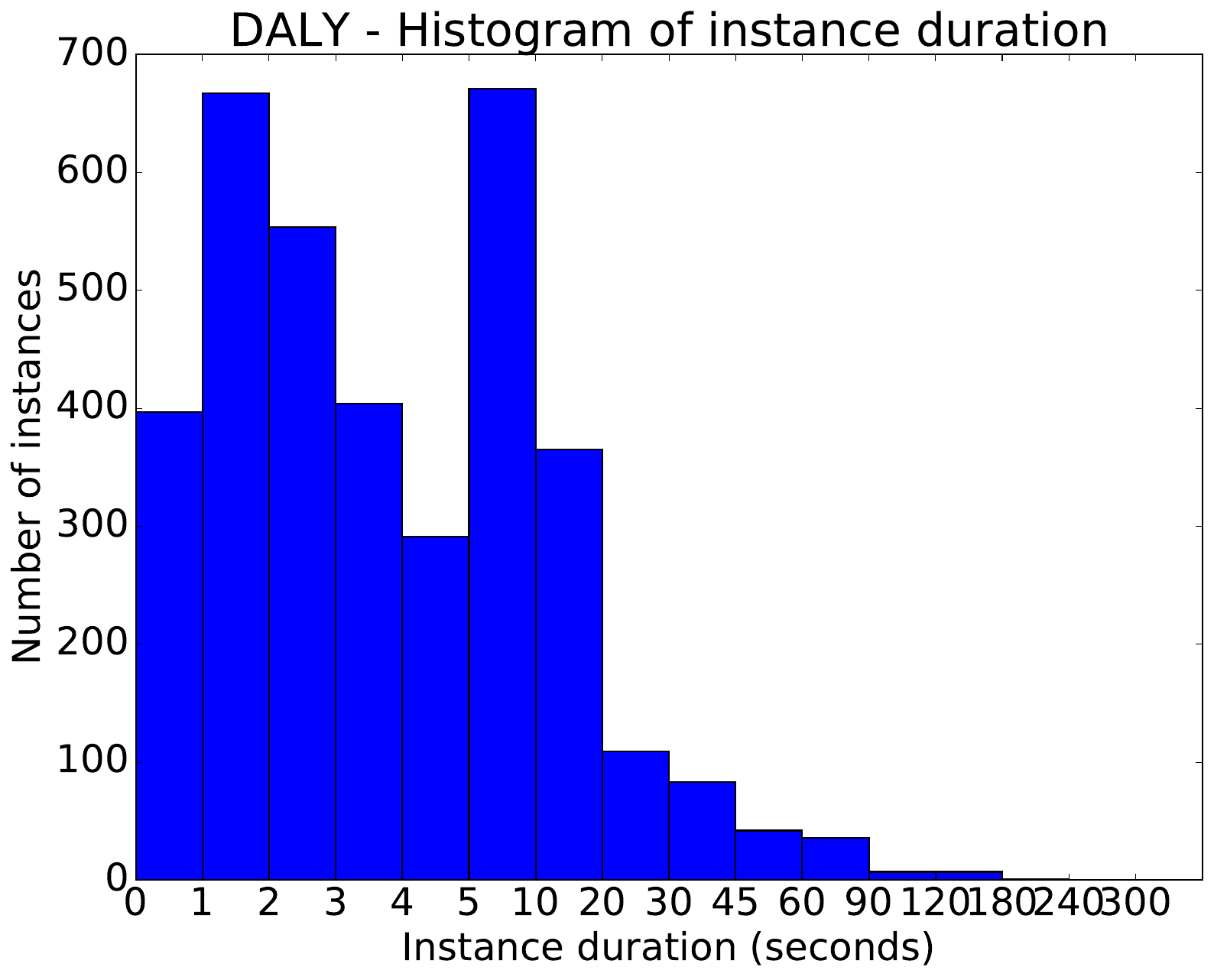}
 \caption{Histogram of duration of the videos (left) and instances (right).}
 \label{fig:histograms}
\end{figure}

Figure~\ref{fig:histograms} shows the histogram of video duration and instance duration. Per-class statistics are presented in Table~\ref{tab:statclass}.
One can see that most videos last several minutes, but less than 10. Videos are longest in average for \textit{ApplyingMakeUpOnLips}, 
mainly because this action tends to be present in face make-up tutorials.
Concerning instances, most of them are shorter than 10 seconds. Nevertheless, DALY also contains instances of several minutes.
In some cases, short instance duration for actions can be explained by video editing. 
The uploader may cut the action to a few seconds and include it in a long video.
Instances are shortest on average for \textit{drinking} and \textit{taking photos},
simply because drinking and taking a photo tend to take a short time. 

\section{Human tube examples}
\label{sup:tubes}

\begin{figure*}
\hfill
\includegraphics[width=0.24\textwidth]{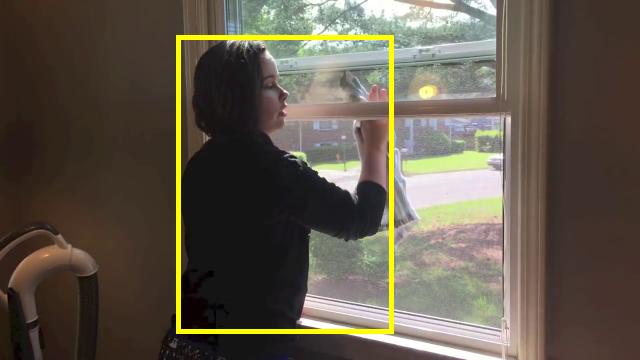}
\hfill
\includegraphics[width=0.24\textwidth]{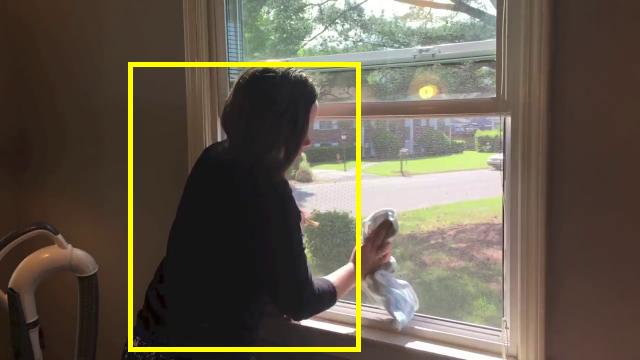}
\hfill
\includegraphics[width=0.24\textwidth]{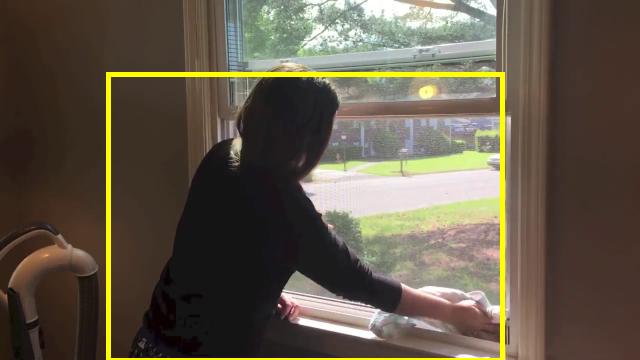}
\hfill
\includegraphics[width=0.24\textwidth]{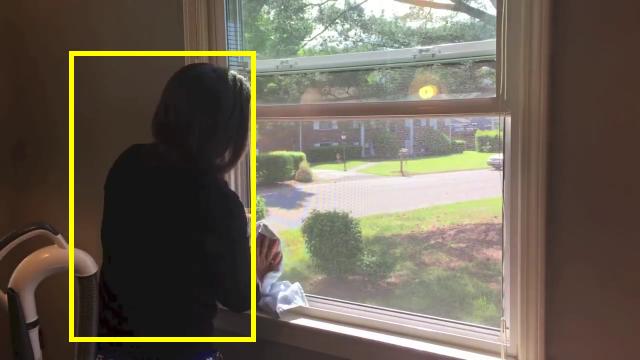}
\hfill \\
\hfill
\includegraphics[width=0.24\textwidth]{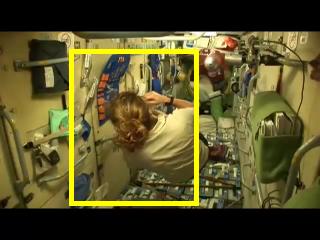}
\hfill
\includegraphics[width=0.24\textwidth]{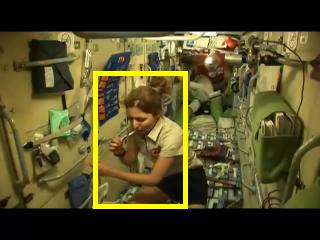}
\hfill
\includegraphics[width=0.24\textwidth]{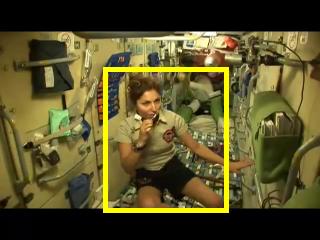}
\hfill
\includegraphics[width=0.24\textwidth]{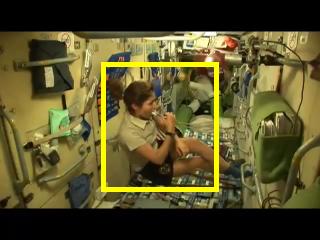}
\hfill \\
\hfill
\includegraphics[width=0.24\textwidth]{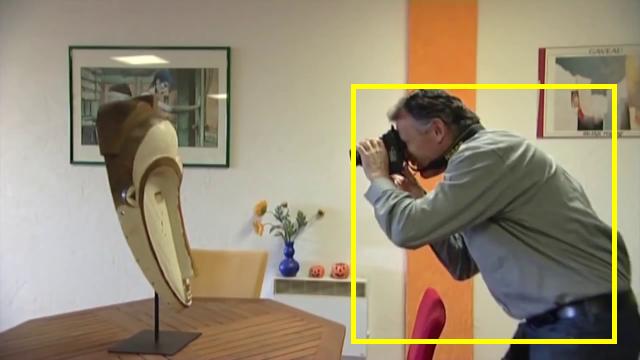}
\hfill
\includegraphics[width=0.24\textwidth]{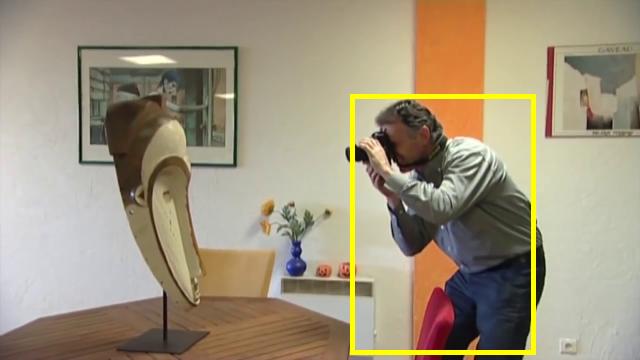}
\hfill
\includegraphics[width=0.24\textwidth]{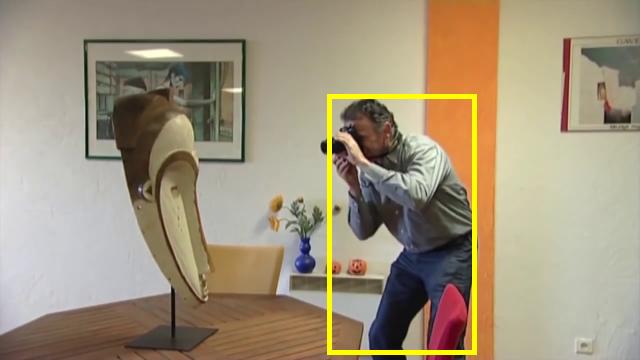}
\hfill
\includegraphics[width=0.24\textwidth]{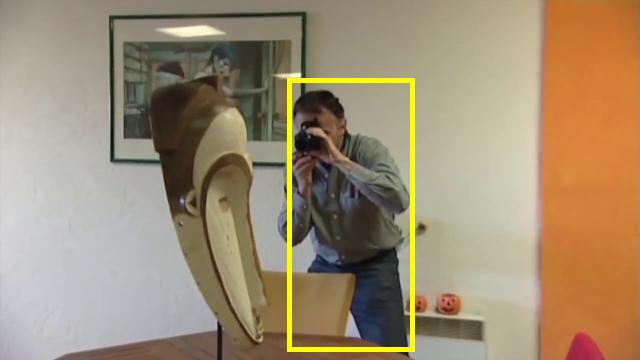}
\hfill\\
\hfill
\includegraphics[width=0.24\textwidth]{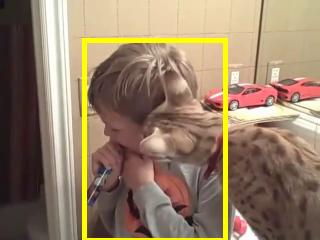}
\hfill
\includegraphics[width=0.24\textwidth]{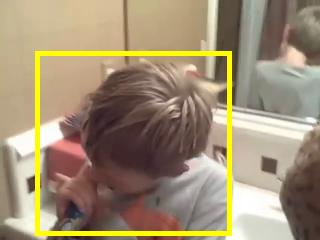}
\hfill
\includegraphics[width=0.24\textwidth]{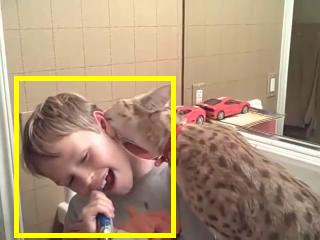}
\hfill
\includegraphics[width=0.24\textwidth]{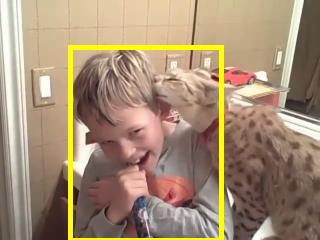}
\hfill \\
\hfill
\includegraphics[width=0.24\textwidth]{fig/daly_track/-BZKxQ99HZM_782.jpg}
\hfill
\includegraphics[width=0.24\textwidth]{fig/daly_track/-BZKxQ99HZM_1348.jpg}
\hfill
\includegraphics[width=0.24\textwidth]{fig/daly_track/-BZKxQ99HZM_1472.jpg}
\hfill
\includegraphics[width=0.24\textwidth]{fig/daly_track/-BZKxQ99HZM_1491.jpg}
\hfill \\
\hfill
\includegraphics[width=0.24\textwidth]{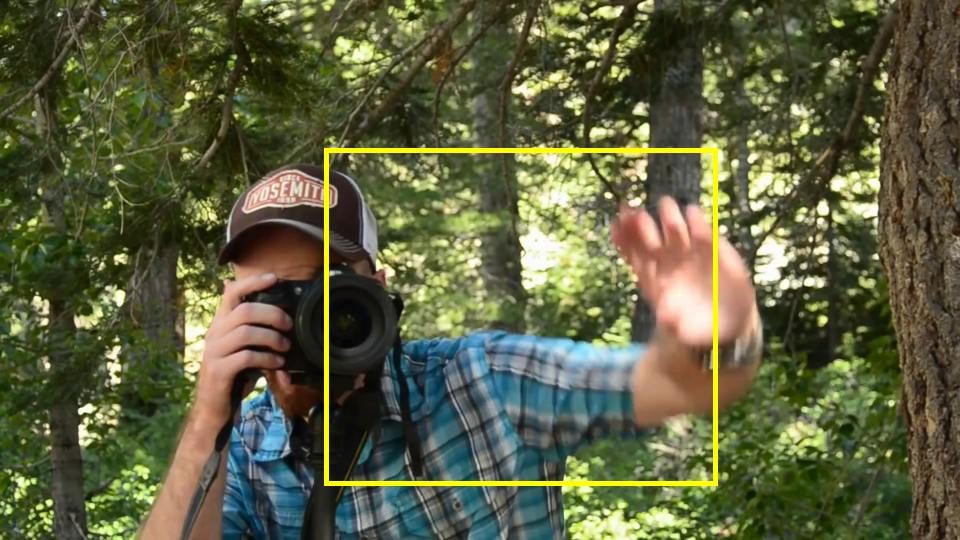}
\hfill
\includegraphics[width=0.24\textwidth]{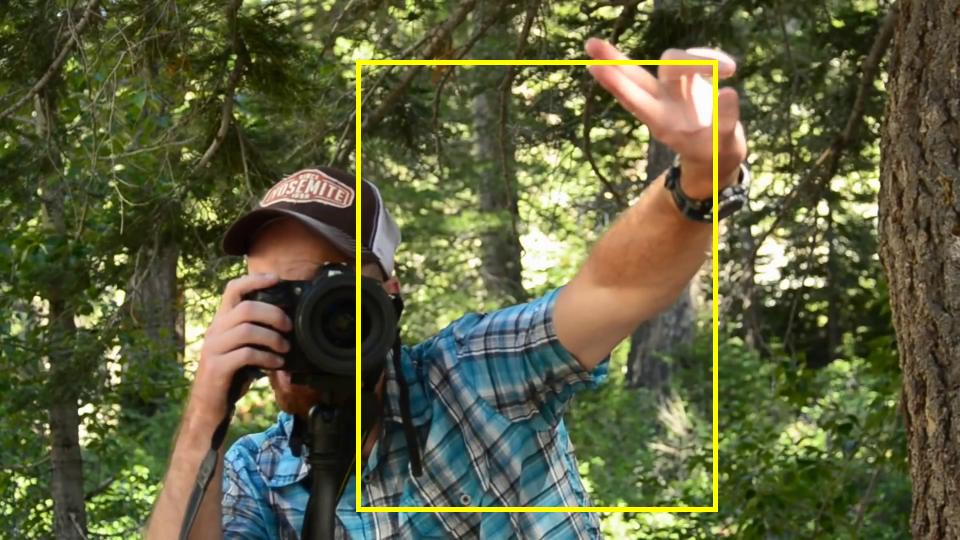}
\hfill
\includegraphics[width=0.24\textwidth]{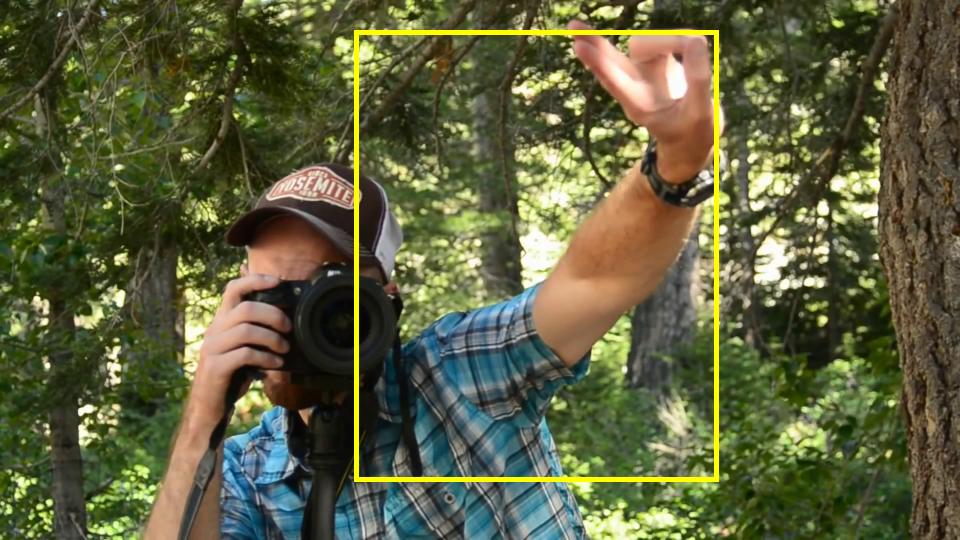}
\hfill
\includegraphics[width=0.24\textwidth]{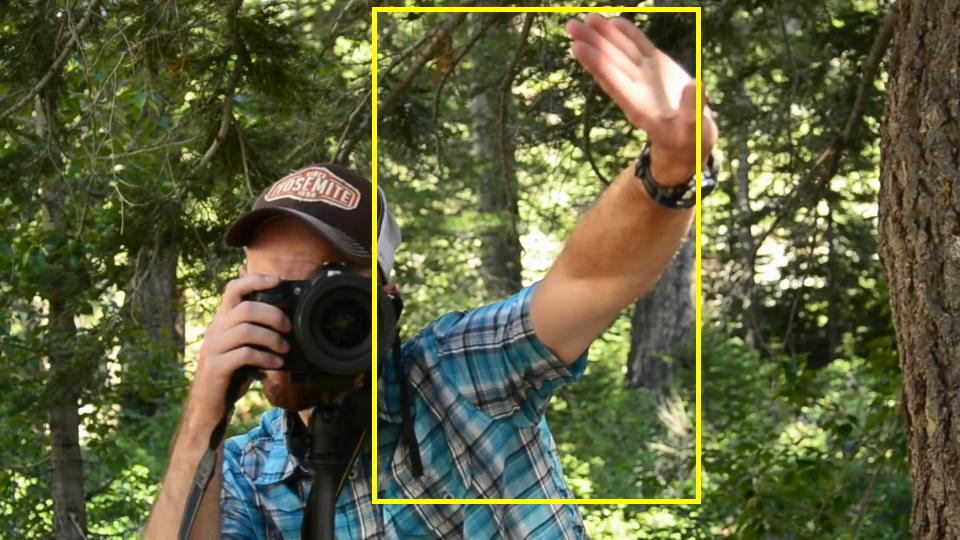}
\hfill \\

\caption{Example of human tubes with successful human tube extraction in the first four rows, 
and some failure cases in the last three rows. 
Failures are caused by partial visibility of the human (end of fifth row and last row) and missed human detection caused by an occluding camera (sixth row). }
\label{fig:humantubes}
\end{figure*}

Figure~\ref{fig:humantubes} shows the highest scoring human tube generated for several sequences of the DALY dataset.
In the first four examples, we can see that the human tube performs well despite motion of one arm (first row), turning of the person (second and third row),
camera motion (third row) or presence of an animal close to the human (fourth row).

Nevertheless, there are some failure cases due to the fact that the full body disappears (fifth row). 
In this case, only the feet remain visible causing the failure of the human tracker, as
the human detector performs poorly and the instance-level detector is
trained on previous frames where the full body is visible.
Another failure case is due to a partial occlusion by the camera
(sixth row), which causes the human detector to fail. 

\section{Detection examples}
\label{sup:examples}

\begin{figure*}
 \includegraphics[width=0.24\textwidth]{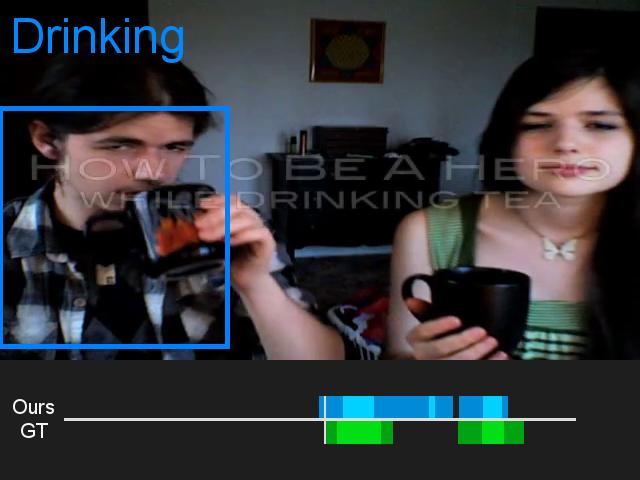} \hfill
 \includegraphics[width=0.24\textwidth]{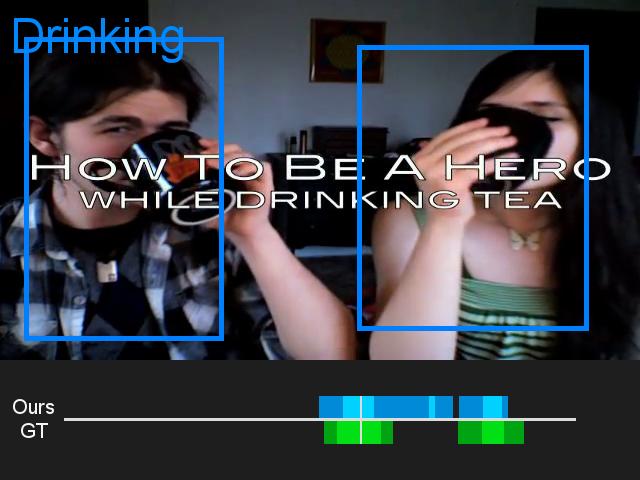} \hfill
 \includegraphics[width=0.24\textwidth]{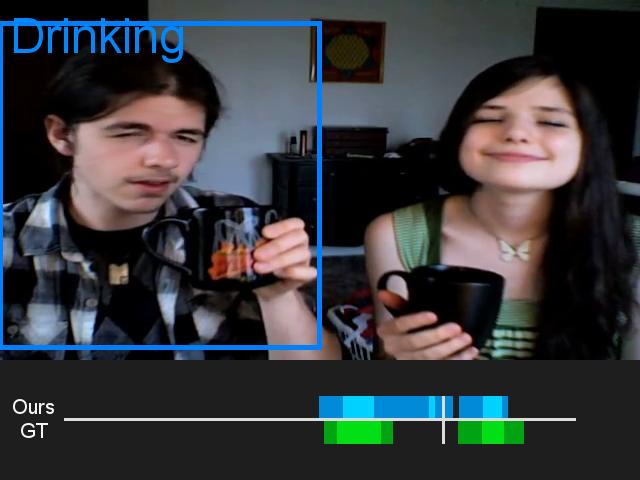} \hfill
 \includegraphics[width=0.24\textwidth]{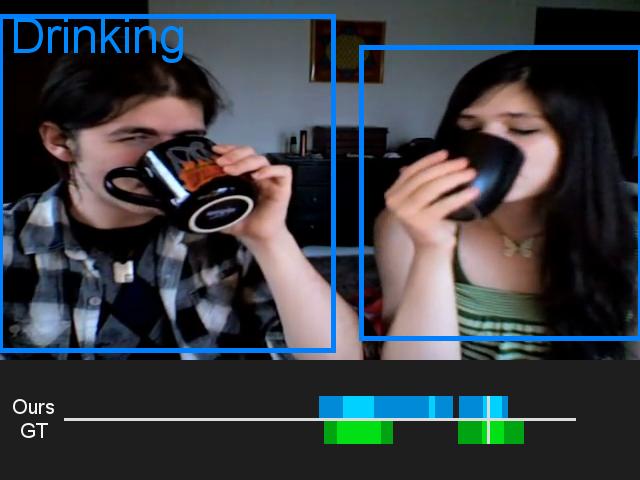} \\
 
 \includegraphics[width=0.24\textwidth]{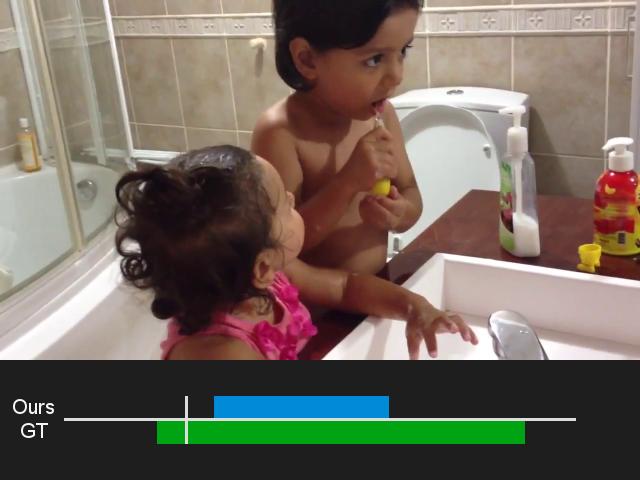} \hfill
 \includegraphics[width=0.24\textwidth]{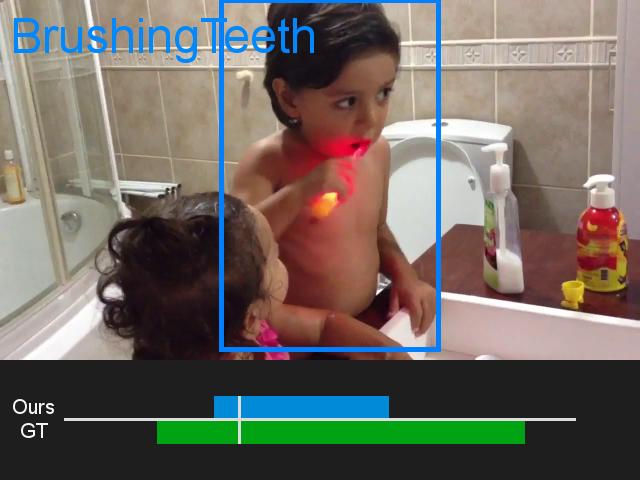} \hfill
 \includegraphics[width=0.24\textwidth]{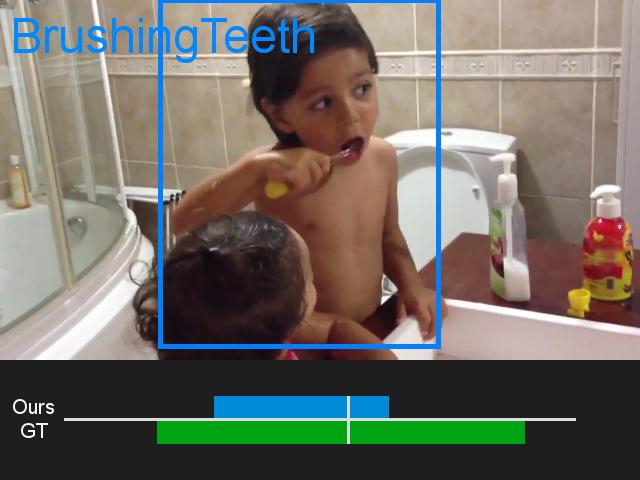} \hfill
 \includegraphics[width=0.24\textwidth]{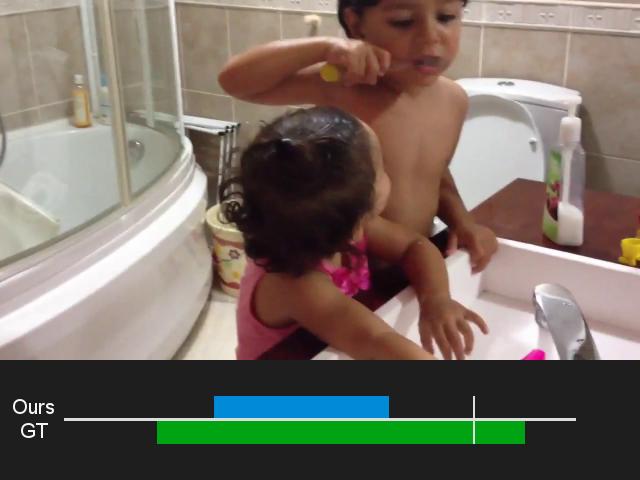} \\
 
 \includegraphics[width=0.24\textwidth]{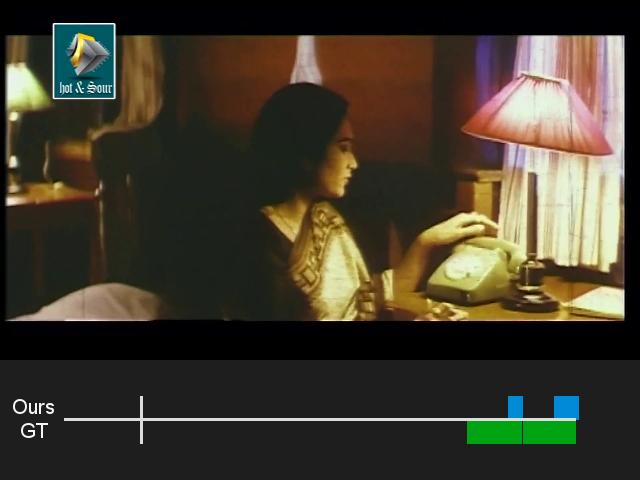} \hfill
 \includegraphics[width=0.24\textwidth]{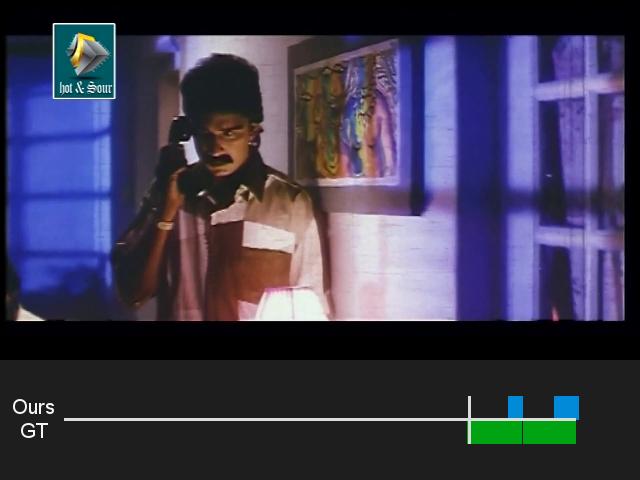} \hfill
 \includegraphics[width=0.24\textwidth]{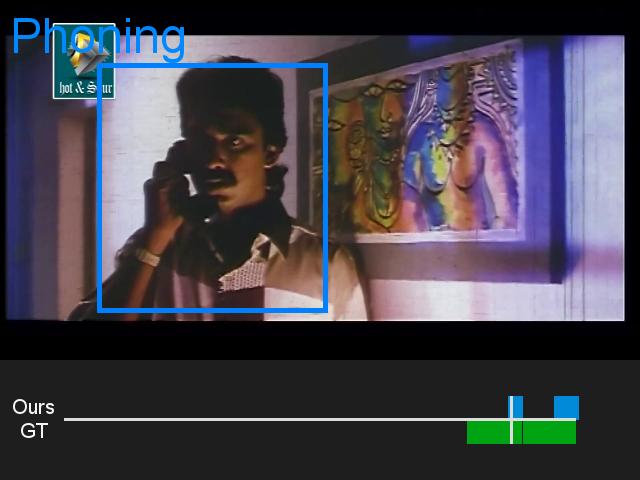} \hfill
 \includegraphics[width=0.24\textwidth]{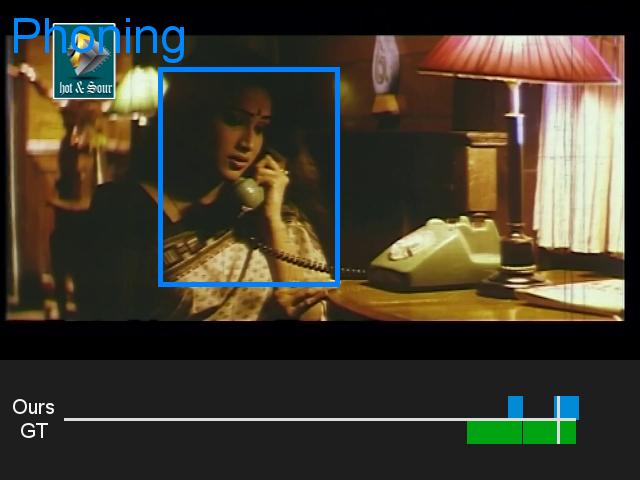} \\
 
 \includegraphics[width=0.24\textwidth]{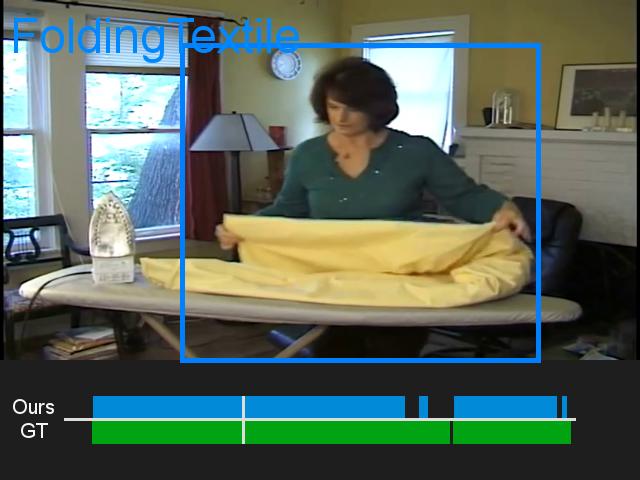} \hfill
 \includegraphics[width=0.24\textwidth]{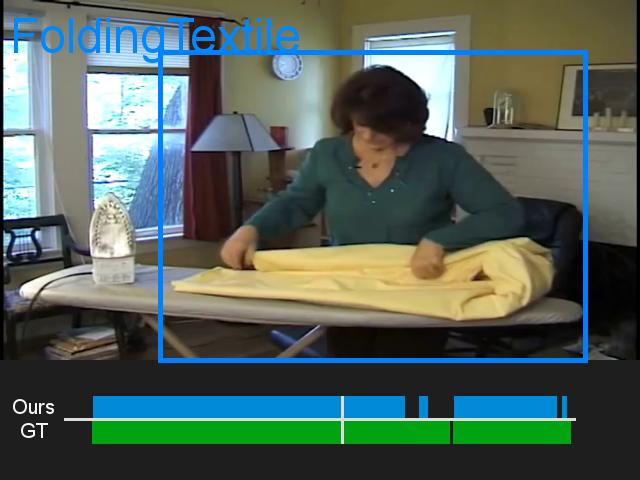} \hfill
 \includegraphics[width=0.24\textwidth]{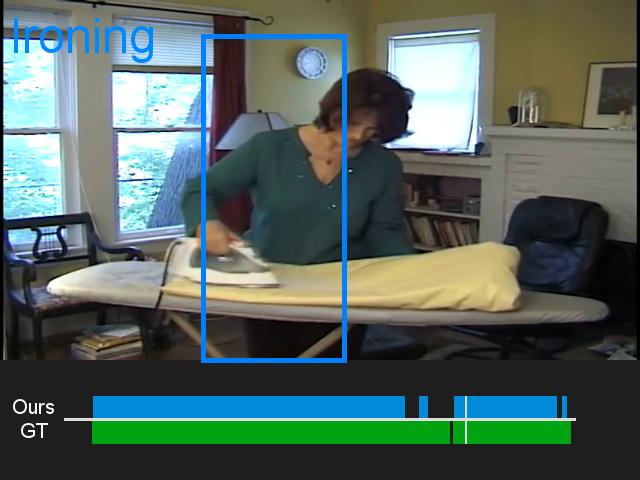} \hfill
 \includegraphics[width=0.24\textwidth]{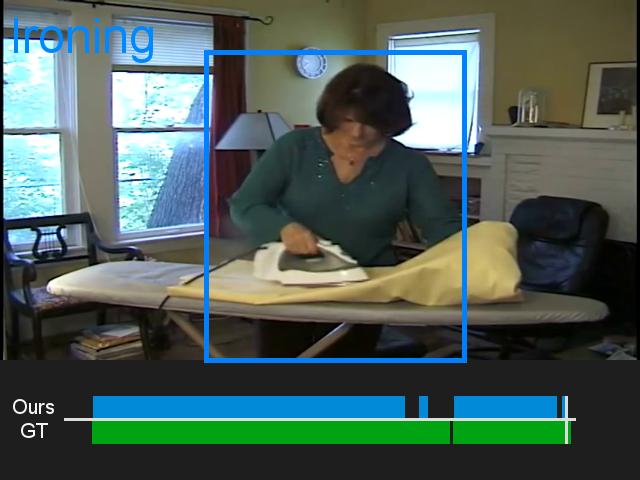} \\
 
 \includegraphics[width=0.24\textwidth]{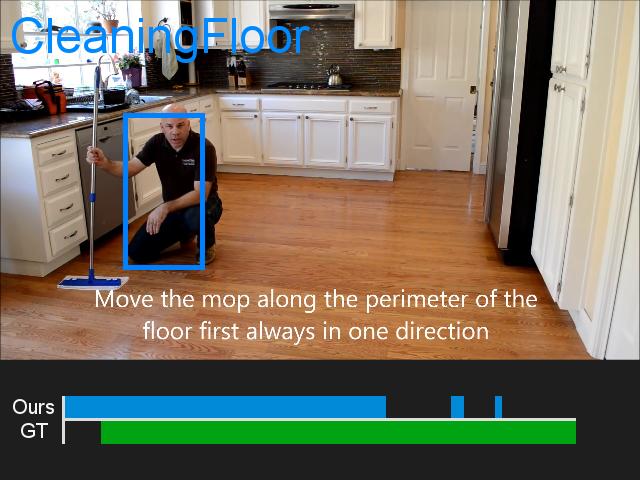} \hfill
 \includegraphics[width=0.24\textwidth]{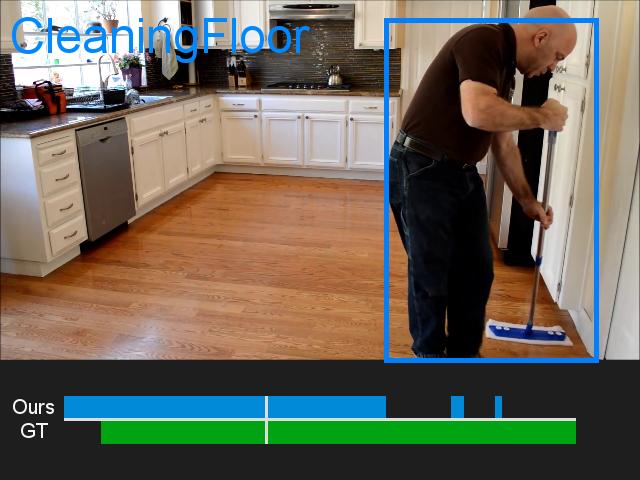} \hfill
 \includegraphics[width=0.24\textwidth]{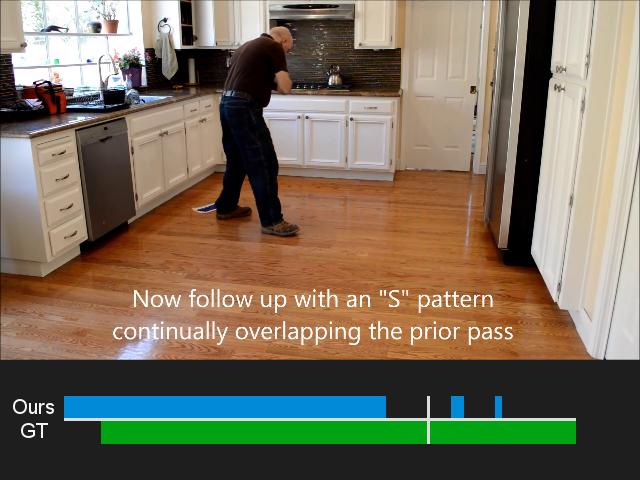} \hfill
 \includegraphics[width=0.24\textwidth]{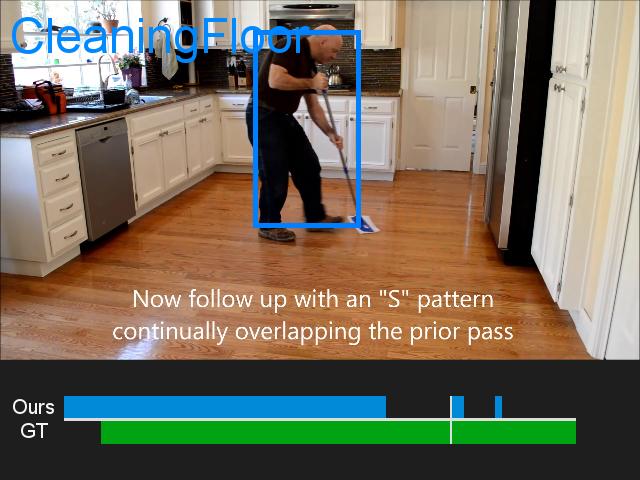} \hfill
 
 \includegraphics[width=0.24\textwidth]{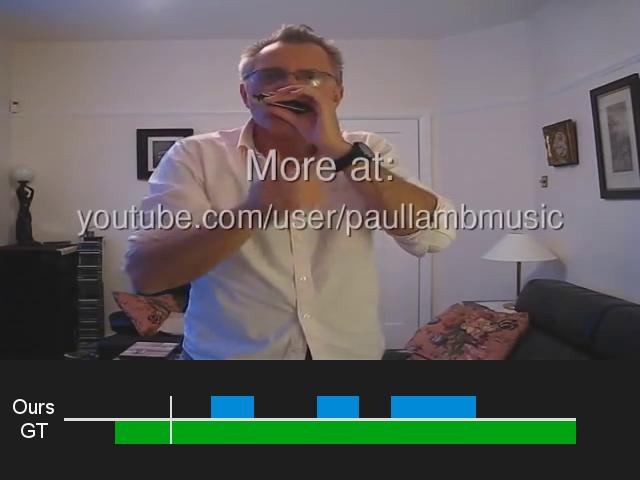} \hfill
 \includegraphics[width=0.24\textwidth]{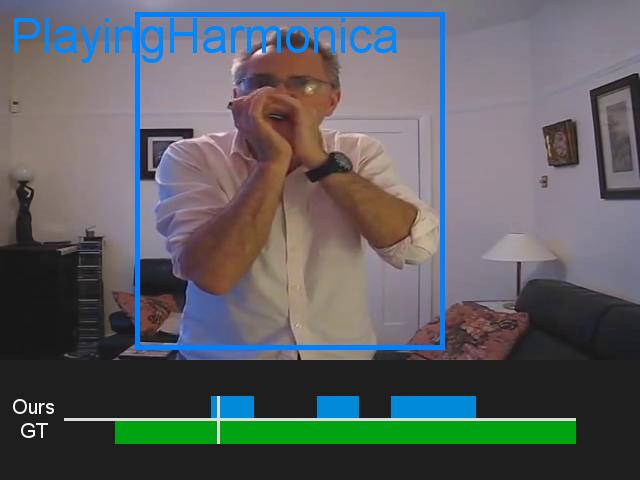} \hfill
 \includegraphics[width=0.24\textwidth]{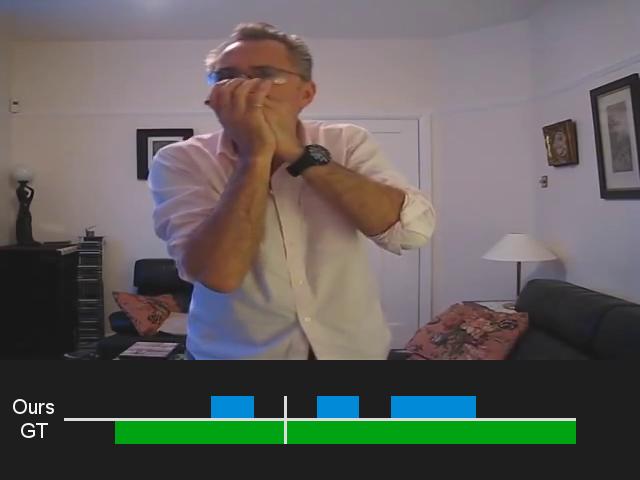} \hfill
 \includegraphics[width=0.24\textwidth]{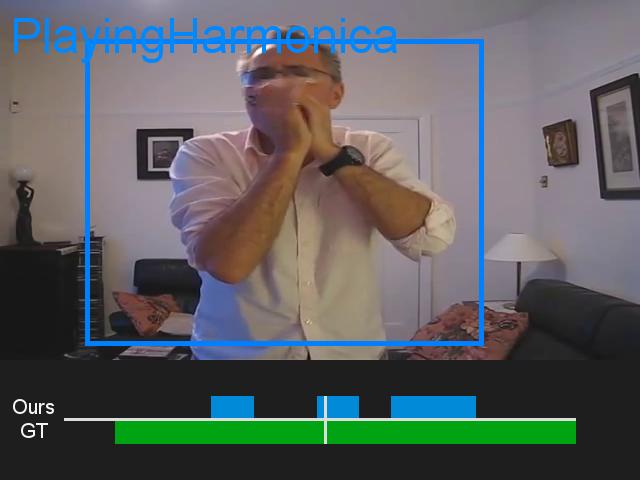} \\
 
 \caption{Spatio-temporal detection examples on the DALY dataset. The timeline shows the ground-truth and our detection for a clip. The video cursor indicates the displayed frame, for which we show the action detection.}
 \label{fig:examples}
\end{figure*}

We show in the attached video a few detection examples in clips from the DALY dataset. 
For each clip in the video, we show a few frames with localization
results in Figure~\ref{fig:examples}
(some of the images are also present in the last figure of the main paper, we duplicate them to match the order of the attached video and for better readability of this section).
The detected bounding boxes are displayed in blue, with the class label written on the top left of the video.
At the bottom of the video is a timeline showing the temporal extent of the ground-truth instances (green) and of our detections (blue).
We now discuss these examples.

The example in the first row contains two people \textit{drinking} multiple times.
Our detections are accurate, with the exception of a false ``drinking" positive on the left person (third column). At this time, the person is holding a cup.
In the second and third rows we detect the correct action with the
correct spatial localization. Our temporal detection, however, is
shorter than the real action. 
The fourth row is a video containing both \textit{folding textile} and \textit{ironing}.
We detect both actions accurately. 
The fifth row shows an example of \textit{cleaning floor}. Our
detection starts a bit earlier and is split at the end into several detections.
However, the person is successfully tracked. 
The last row is a failure case where the action is detected as many
short instances instead of a long one, but the label is correct. All these detections are
considered as wrong since the temporal IoU is below 0.2,   
and thus the spatio-temporal IoU as well.

{\small
\bibliographystyle{ieee}
\bibliography{biblio}
}